\documentclass[letterpaper,12pt,titlepage,oneside,final]{book}

\usepackage{bm}
\usepackage{multirow}
\usepackage{booktabs}
\usepackage{lscape}
\usepackage{amsmath,amsbsy,amsthm,amssymb}

\newcommand{\href}[1]{#1} 

\usepackage{ifthen}
\newboolean{PrintVersion}
\setboolean{PrintVersion}{false} 
\usepackage{amstext} 
\usepackage[pdftex]{graphicx} 
\usepackage[round]{natbib}

\usepackage[pdftex,pagebackref=false]{hyperref} 
\hypersetup{
    plainpages=false,       
    unicode=false,          
    pdftoolbar=true,        
    pdfmenubar=true,        
    pdffitwindow=false,     
    pdfstartview={FitH},    
    pdftitle={Variational Attention for Sequence-to-Sequence Models},    
    pdfauthor={Hareesh Pallikara Bahuleyan},    
    pdfsubject={MASc Thesis Report},  
    pdfkeywords={uwaterloo} {seq2seq} {machine learning} {nlp}, 
    pdfnewwindow=true,      
    colorlinks=true,        
    linkcolor=blue,         
    citecolor=blue,        
    filecolor=magenta,      
    urlcolor=cyan           
}
\ifthenelse{\boolean{PrintVersion}}{   
\hypersetup{	
    citecolor=black,%
    filecolor=black,%
    linkcolor=black,%
    urlcolor=black}
}{} 

\usepackage[automake,toc,abbreviations]{glossaries-extra} 

\let\origdoublepage\cleardoublepage
\newcommand{\clearemptydoublepage}{%
  \clearpage{\pagestyle{empty}\origdoublepage}}
\let\cleardoublepage\clearemptydoublepage

\newglossaryentry{computer}
{
name=computer,
description={A programmable machine that receives input data,
               stores and manipulates the data, and provides
               formatted output}
}

\newglossary*{nomenclature}{Nomenclature}
\newglossaryentry{dingledorf}
{
type=nomenclature,
name=dingledorf,
description={A person of supposed average intelligence who makes incredibly brainless misjudgments}
}

\newabbreviation{ai}{AI}{Artificial Intelligence}

\newabbreviation{nlp}{NLP}{Natural Language Processing}

\newabbreviation{Seq2Seq}{Seq2Seq}{sequence-to-sequence}

\newabbreviation{vae}{VAE}{Variational Autoencoders}

\newabbreviation{dae}{DAE}{Deterministic Autoencoders}

\newabbreviation{ved}{VED}{Variational Encoder-Decoder}

\newabbreviation{ded}{DED}{Deterministic Encoder-Decoder}

\newabbreviation{kl}{KL}{Kullback-Leibler}

\newabbreviation{ml}{ML}{Machine Learning}

\newabbreviation{ann}{ANN}{Artificial Neural Networks}

\newabbreviation{cnn}{CNN}{Convolutional Neural Network}

\newabbreviation{rnn}{RNN}{Recurrent Neural Network}

\newabbreviation{lstm}{LSTM}{Long Short Term Memory}

\newabbreviation{relu}{ReLU}{Rectified Linear Unit}

\newabbreviation{gru}{GRU}{Gated Recurrent Unit}

\newabbreviation{elbo}{ELBO}{Evidence Lower Bound}

\newabbreviation{sgd}{SGD}{stochastic gradient descent}

\newglossary*{symbols}{List of Symbols}
\newglossaryentry{rvec}
{
name={$\mathbf{v}$},
sort={label},
type=symbols,
description={Random vector: a location in n-dimensional Cartesian space, where each dimensional component is determined by a random process}
}
 
\makeglossaries

\newcommand{\btheta}{{\bm \theta}}
\newcommand{\bphi}{{\bm \phi}}

\newcommand{\KL}{{\operatorname{KL}}}

\newcommand{\LSTM}{{\operatorname{LSTM}}}
\newcommand{\ELBO}{{\operatorname{ELBO}}}

\newcommand{\n}{^{(n)}}
\newcommand{\src}{^\text{(src)}}
\newcommand{\tar}{^\text{(tar)}}
\newcommand{\barh}{\bar{\bm h}\src}
\DeclareMathOperator*{\argmin}{argmin}

\begin{document}

\pagestyle{empty}
\pagenumbering{roman}

\begin{titlepage}
        \begin{center}
        \vspace*{1.0cm}

        \Huge
        {\bf Natural Language Generation with Neural Variational Models }

        \vspace*{1.0cm}

        \normalsize
        by \\

        \vspace*{1.0cm}

        \Large
        Hareesh Bahuleyan \\

        \vspace*{3.0cm}

        \normalsize
        A thesis \\
        presented to the University of Waterloo \\ 
        in fulfillment of the \\
        thesis requirement for the degree of \\
        Master of Applied Science \\
        in \\
        Management Science \\

        \vspace*{2.0cm}

        Waterloo, Ontario, Canada, 2018 \\

        \vspace*{1.0cm}

        \copyright\ Hareesh Bahuleyan 2018 \\
        \end{center}
\end{titlepage}

\pagestyle{plain}
\setcounter{page}{2}

\cleardoublepage 
\begin{center}
    {\bf \large Author Declaration }    
\end{center}
  \noindent
This thesis consists of material all of which I authored or co-authored: see \textit{Statement
of Contributions} included in the thesis. This is a true copy of the thesis, including any
required final revisions, as accepted by my examiners.

  \bigskip
  
  \noindent
I understand that my thesis may be made electronically available to the public.

\cleardoublepage

\begin{center}
    {\bf \large Statement of Contributions }    
\end{center}
Chapters~\ref{chap:vae}, \ref{chap:bypass} and \ref{chap:ved} of this thesis are based on the following papers:
\begin{enumerate}
    \item \textbf{Hareesh Bahuleyan}*, Lili Mou*, Olga Vechtomova, and Pascal Poupart. Variational Attention for  Sequence-to-Sequence  Models.   In \textit{27th  International  Conference  on  Computational Linguistics (COLING)}, 2018.
    \item \textbf{Hareesh Bahuleyan}, Lili Mou, Kartik Vamaraju, Hao Zhou, and Olga Vechtomova.  Probabilistic  Natural  Language  Generation  with  Wasserstein  Autoencoders. \textit{arXiv  preprint arXiv:1806.08462}, 2018
\end{enumerate}
\noindent I have contributed to implementation, experimentation, and preparation of the manuscript of the above mentioned papers. 

\cleardoublepage


\begin{center}\textbf{\large Abstract}\end{center}

Automatic generation of text is an important topic in natural language processing with applications in tasks such as machine translation and text summarization. In this thesis, we explore the use of deep neural networks for generation of natural language. Specifically, we implement two sequence-to-sequence neural variational models - variational autoencoders (VAE) and variational encoder-decoders (VED).

VAEs for text generation are difficult to train due to issues associated with the Kullback-Leibler (KL) divergence term of the loss function vanishing to zero. We successfully train VAEs by implementing optimization heuristics such as KL weight annealing and word dropout. In addition, this work also proposes new and improved annealing schedules that facilitates the learning of a meaningful latent space. We also demonstrate the effectiveness of this continuous latent space through experiments such as random sampling, linear interpolation and sampling from the neighborhood of the input. We argue that if VAEs are not designed appropriately, it may lead to bypassing connections which results in the latent space being ignored during training. We show experimentally with the example of decoder hidden state initialization that such bypassing connections degrade the VAE into a deterministic model, thereby reducing the diversity of generated sentences. 

We discover that the traditional attention mechanism used in sequence-to-sequence VED models serves as a bypassing connection, thereby deteriorating the model's latent space. In order to circumvent this issue, we propose the variational attention mechanism where the attention context vector is modeled as a random variable that can be sampled from a distribution. We show empirically using automatic evaluation metrics, namely entropy and distinct measures, that our variational attention model generates more diverse output sentences than the deterministic attention model.  A qualitative analysis with human evaluation study proves that our model simultaneously produces sentences that are of high quality and  equally fluent as the ones generated by the deterministic attention counterpart. 

\cleardoublepage


\begin{center}\textbf{ \large Acknowledgements}\end{center}

This thesis would not have been possible without the constant support that I received from a number of people. 

First and foremost, I take this opportunity to express my heartfelt gratitude to my supervisor Professor Olga Vechtomova for her guidance throughout this research. I am thankful to her for believing in me and being patient with me since the start of my program. The flexibility that she provided in research has allowed me to learn new topics and explore new ideas. I am deeply indebted to her for having taken the time and effort for reading, reviewing and providing valuable inputs for the reports that I had made.

I am grateful to Dr. Lili Mou for being a mentor, sharing knowledge and providing valuable technical advice. His subject expertise and guidance has played a crucial role in structuring this project. Interactions with him have helped me develop my skills in this field and motivated me to deal with research challenges. 

I would like to acknowledge Professor Pascal Poupart for sharing his ideas and providing valuable suggestions. 

I thank Professor Jesse Hoey and Professor Stan Dimitrov for taking the time to review my thesis and provide feedback.

I express my gratitude to Vineet John and Ankit Vadehra for the informative discussions and enlightening me on their respective research topics. 

Life in Canada would not have have been such an enjoyable journey without the friends that I made here. I thank all my friends for making my stay here at Waterloo, a memorable one.

I am extremely grateful to my parents and my sister for supporting me and being with me even during the toughest times. 

\cleardoublepage


\begin{center}\textbf{\large Dedication}\end{center}

\textit{I dedicate this thesis to my beloved parents and sister for their unconditional love, support, and care.} 
\cleardoublepage

\renewcommand\contentsname{Table of Contents}
\tableofcontents
\cleardoublepage
\phantomsection    

\addcontentsline{toc}{chapter}{List of Tables}
\listoftables
\cleardoublepage
\phantomsection		

\addcontentsline{toc}{chapter}{List of Figures}
\listoffigures
\cleardoublepage
\phantomsection		

\printglossaries
\cleardoublepage
\phantomsection		

\pagenumbering{arabic}


\chapter{Introduction}
\section{Background}
The term \gls{ai} was introduced by Professor John McCarthy in 1956, who defined it as ``science and engineering of making intelligent machines''. Although the field of \gls{ai} covers a broad range of topics, it is generally perceived as the task of making machines achieve \textit{human-level} intelligence \citep{mccarthy1989artificial}.  

Fast-forward a few decades, we have achieved great advancements in the field of \gls{ai}. For example, the neural network model developed by \cite{he2016deep} was able to classify images with an accuracy of $96.5\%$, surpassing human-level performance. A new milestone was achieved in the area of speech recognition when the system developed by \cite{xiong2017microsoft} was able to carry out the task with a record minimum word error rate of $5.1\%$. There are numerous other commendable accomplishments made in the last decade such as AI beating the human grandmaster in the game of Go \citep{gibney2016google} and the transformations in the transportation industry with the introduction of self-driving vehicles \citep{bojarski2016end}.

This recent success can be attributed to the research developments made in a multitude of fields such as computer science, mathematics, neuroscience, psychology, linguistics and so on. This work focuses on the field of computer science, specifically the areas of machine learning and natural language processing. 

Machine learning is a sub-field of artificial intelligence in which computers are \textit{taught} to acquire knowledge or \textit{learn} from data, without being explicitly programmed. In the past, machine learning algorithms have been used to recognize patterns in the data and make informed decisions. End-to-end automation with machine learning has helped improve the efficiencies of processes and workflows in sectors such as manufacturing. The \gls{ai} explosion in the recent years is primarily due to the success of a sub-class of machine learning models known as deep neural networks. This field, known as deep learning, along with the availability of massive amounts of data and powerful hardware for computation has made possible the latest advancements in \gls{ai}. 

The focus of this work lies in the development and application of deep learning models to \gls{nlp}. The study of \gls{nlp} is concerned with how computers can effectively interact with humans using natural language. Broadly speaking, it deals with manipulation, understanding, interpretation and generation of textual and speech data. A few examples of \gls{nlp} tasks include question answering, sentiment analysis, named-entity recognition and machine translation. 

Natural language generation is an \gls{nlp} task that deals with generation of text in human language. This is challenging because the text generated by a good system has to be \textit{syntactically} (follow the rules of the language) and \textit{semantically} (meaningful) correct. In this work, \gls{Seq2Seq} models for natural language generation are explored. In \gls{Seq2Seq} models \citep{sutskever2014sequence}, a sequence (of words) is given as input in order to generate another sequence as output. This has applications in tasks such as machine translation where a sentence in English can be fed to the model which generates its corresponding translated sentence in French. Another application is text summarization, where we may attempt to generate a shorter version of a larger body of text, such as a paragraph. 

The aim of this work is to integrate attention mechanism into the \gls{ved} framework. Attention mechanism has been shown to be particularly useful in improving the performance of sequence-to-sequence tasks such as machine translation \citep{bahdanau2014neural,luong2015effective} and dialog generation \citep{yao2016attentional,mei2017coherent}. A class of models that combine deep learning and variational inference, namely \gls{vae} \citep{kingma2013auto} have been successfully applied to the task of text generation \citep{bowman2015generating}. Attention mechanism enables the model to generate fluent sentences, relevant to the input. Simultaneously, we would be able to generate diverse outputs (for the given input) by sampling from a latent space. This way, the proposed model combines the strengths of the attention mechanism and variational models. 

\section{Motivation and Problem Definition}

In this research, we work with deep neural network models known as sequence-to-sequence (Seq2Seq) models that take a sentence as input and generate another sentence as output. 
Let $\bm x = (x_1, x_2, \cdots, x_{|\bm x|})$ be the input sequence of words and $\bm y = (y_1, y_2, \cdots, y_{|\bm y|})$ be the output sequence generated by the model, where $|\bm x|$ and $|\bm y|$ correspond to the number of tokens (words) in the input and output sequences respectively.

Consider a conversational system such as a chatbot, where $\bm x$ is the line input by the user and $\bm y$ is the line generated by the machine. \gls{Seq2Seq} neural network models can be designed to function as such conversational agents. Traditional conversational systems tend to output safe and commonplace responses such as \textit{``I don’t know''} \citep{li2015diversity}. This is because the line \textit{``I don’t know''}  tends to appear in the training dataset with a high frequency. One cannot label such a generic response as incorrect, since it tends to be a valid response (refer Table~\ref{tab:genericresponses}). However, such responses make the conversational agent uninteresting and less engaging. 

\begin{table}[!t]
	\centering
	\resizebox{0.4\linewidth}{!}{
    \begin{tabular}{|c|c|}
    \toprule
    \multicolumn{2}{|c|}{\textbf{Input: What are you doing?}} \\
    \midrule
    \midrule
    \textit{I don't know.} & \textit{Get out of here.} \\
    \textit{I don't know!} & \textit{I'm going home.} \\
    \textit{Nothing.} & \textit{Oh, my god!} \\
    \textit{Get out of the way.} & \textit{I'm talking to you.} \\
    \midrule
    \multicolumn{2}{|c|}{\textbf{Input: What is your name?}} \\
    \midrule
    \midrule
    \textit{I don't know.} & \textit{My name is Robert.} \\
    \textit{I don't know!} & \textit{My name is John.} \\
    \textit{I don't know, sir.} & \textit{My name's John.} \\
    \textit{Oh, my god!} & \textit{My name is Alice.} \\
    \midrule
    \multicolumn{2}{|c|}{\textbf{Input: How old are you?}} \\
    \midrule
    \midrule
    \textit{I don't know.} & \textit{Twenty-five.} \\
    \textit{I'm fine.} & \textit{Five.} \\
    \textit{I'm all right.} & \textit{Eight.} \\
    \textit{I'm not sure.} & \textit{Ten years old.} \\
    \bottomrule
    \end{tabular}%
    }
    \caption{Output dialog generated by a conversational system as shown in \cite{li2015diversity}. For a given input, the first column shows the list of top-ranked (most probable) responses, most of which tend to be generic. The second column shows the lower-ranked but less generic/more engaging responses.}
  \label{tab:genericresponses}%
\end{table}%

The motivation for this work is derived from the above example. We would like to be able to generate a diverse set of responses ($\bm y$) for a given input line ($\bm x$). Neural variational models can be used to encode input data into latent variables. It is further possible to sample multiple points from the latent space in order to generate diverse outputs. Attention mechanisms such as those proposed in \citep{bahdanau2014neural,luong2015effective} have significantly improved the performance of sequence-to-sequence text generation tasks. Attention mechanisms help in dynamically aligning the source $\bm x$ and target $\bm y$ during generation.

In variational Seq2Seq, however, the attention mechanism unfortunately serves as a ``bypassing'' mechanism. In other words, the variational latent space does not need to learn much, as long as the attention mechanism itself is powerful enough to capture source information.

In this work, we study how attention mechanisms can be integrated into variational neural models, while avoiding the issue of ``bypassing''. In this work, we propose a variational attention mechanism to address this problem. This is done by modeling the attention context vector as random variables by imposing a probabilistic distribution. By doing this we would be able to combine the stochasticity introduced by variational models with the alignment capabilities achieved through attention mechanism. 

\section{Contributions}
The contributions of this thesis are multi-fold and listed below:

\begin{enumerate}

\item A variational auto-encoder (VAE) is first designed following the work of \cite{bowman2015generating}. We overcome the difficulties associated with training VAE models for natural language generation by employing strategies such as (1) annealing coefficient of the \gls{kl} loss term and (2) word-dropout. We also propose new and improved annealing schedules. The effectiveness of the model is demonstrated by random sampling and linear interpolation of sentences in the latent space.  

\item We discover a ``bypassing'' phenomenon in VAEs that causes the latent or variational space to be ignored during training. This results in the model becoming more \textit{deterministic} in nature; the evidence for which is revealed through the lower diversity of generated sentences. 

\item We realize that traditional attention mechanism in variational encoder-decoder (VED) models serves as a ``bypassing'' connection. To this end and in contrast to previous models that utilize attention in a deterministic manner, we propose a variational attention mechanism that can be applied in the context of VED models. In the proposed framework, the attention context vector is modeled as a random variable that can be sampled from a distribution. 


\item We propose two plausible priors for modeling the prior distribution of the attention context vector in the variational attention VED framework. Both the priors work equally well in alleviating the problem of “bypassing”, which is observed in the VED baselines with deterministic attention. 

\item Experiments are carried out on two tasks –-- question generation and conversational systems. Quantitative evaluation metrics show that the proposed variational attention yields a higher diversity than variational Seq2Seq with deterministic attention, while retaining high quality of generated sentences. A qualitative analysis with human evaluation study also supports our claim regarding the fluency of sentences generated by the proposed model.

\end{enumerate}

\section{Chapter Outline}
The rest of this thesis report is organized as follows:
\begin{itemize}
\item Chapter 2 provides a background and brief overview of the deep learning techniques used in this work, along with related work in the area of \gls{Seq2Seq} models. 
\item Chapter 3 describes \gls{Seq2Seq} neural network architecture in depth. 
\item Chapter 4 introduces variational inference and a detailed description of variational autoencoders. The heuristics involved in training VAEs for natural language generation are presented followed by the results. 
\item Chapter 5 outlines the bypassing phenomenon that we discover, if the VAE architecture is not designed properly and its implications. 
\item Chapter 6 provides details about the proposed variational attention model, which is compared to the traditional deterministic attention. We demonstrate the benefits of our model through qualitative and quantitative evalutation.
\item Chapter 7 gives a summary of the work, followed by conclusions and scope for future work.
\end{itemize}

\chapter{Background and Related Work}
In this chapter, different classes of machine learning models are first outlined. Then, we introduce deep learning and the structure of artificial neural networks. Following this, we describe recurrent neural networks, specifically long short term memory networks, which are widely used in the NLP literature. Since this work deals with text generation, we review the recent advances in this area, specifically sequence-to sequence models, attention mechanism and audoencoding. The chapter concludes with the related work in the tasks of question generation and dialog systems, which are the two experiments conducted in this study, to evaluate the proposed model. 

\section{Machine Learning}
\label{sec:machinelearning}

The science of \gls{ml} involves enabling computers to \textit{learn} from data, without being explicitly programmed. Data is used to train the system to perform a specific task. The model, which uses some form of mathematical optimization and statistical methods, recognizes the patterns and intricacies within the data. This can be then used to automate tasks or guide decision making, simply based on data and the mathematical model.

Machine learning is being increasingly used in our day-to-day lives. For example, all email service providers today use \gls{ml} to filter out spam emails. Similarly, the online shopping recommendations provided to us by ecommerce websites is based on \gls{ml}. The field of machine learning is developing at a fast pace. Researchers have been developing algorithms and new methodologies and also simultaneously applying these techniques to new application areas (such as medical diagnosis \citep{kourou2015machine,foster2014machine} and climate change \citep{lakshmanan2015machine}). The evolution of intelligent systems are definitely beneficial because it makes processes more efficient, and at the same time, requiring minimal human intervention.

Broadly speaking, machine learning methodologies can be classified into two categories: supervised learning and unsupervised learning. At a high level, this categorization is based on how the learning process is carried out. The following subsections describe each one with examples. 

\subsection{Supervised Learning}
\label{sec:supervised}
In supervised machine learning, we provide the model with sample inputs and their corresponding ground truth labels. Here, the task of the machine learning algorithm is usually to modify the model parameters, such that it obtains the desired output for the given input. The important point to note here is that we have labelled outputs corresponding to each input used to \textit{train} the model. After sufficient training, if the model is provided with a new unseen input data point, it should be able to predict the target, based on what it has previously \textit{learnt}.

To understand supervised learning with the help of an example consider the task of image classification. We feed the machine learning model with images of huskies, retrievers, dachshunds, etc. and label them as \textit{dogs}. Similarly, we can provide images of pigeons, eagles, sparrows, etc., all labelled as \textit{birds}. The model tries to \textit{learn} from the image pixel values and their corresponding class labels, as to what would be the characteristics that differentiate \textit{dogs} from \textit{birds}, which is then used to classify new images. 

\subsection{Unsupervised Learning}
\label{sec:unsupervised}
In contrast to the approach discussed in  Section~\ref{sec:supervised}, the data provided to an unsupervised machine learning model will not contain labels or corresponding target values. The task of such models would be to identify patterns of similarity or differences within the input data points. It can also be used for detecting anomalies, wherein some parts of the data may not fit well with the rest of the data. 

An example of unsupervised learning task would be clustering of documents by topic. Assume that we provide an unsupervised machine learning model with \textit{unlabelled} documents pertaining to different topics such as sports, politics and entertainment. A good model should be able to automatically cluster \textit{similar} documents (belonging to the same topic) together, using information such as the word usage and writing style. 

Other areas in machine learning include topics such as semi-supervised learning and reinforcement learning, which are not covered in this text. 

\section{Deep Learning}
\label{sec:deeplearning}
\gls{ann} are an important class of machine learning models, used for both supervised and unsupervised tasks. The structure and functioning of \gls{ann}s are loosely inspired by biological neural networks. The brain consists of a large number of interconnected neurons, which \gls{ann}s try to mimic. \gls{ann}s consist of multiple layers of simple processing units known as \textit{nodes}, which are connected by \textit{edges} with \textit{weights} \citep{gurney2014introduction} (refer to Section~\ref{sec:dl-intro} for details). 

Over the last decade, there has been an increasing interest in neural network architectures consisting of many layers. Along with the availability of massive amounts of data and powerful hardware for computation, such model architectures were able to outperform humans in a number of cognitive tasks \citep{schmidhuber2015deep,najafabadi2015deep}. This led to the creation of a sub-field of machine learning known as deep learning (referring to deep neural networks) \citep{lecun2015deep}. 

The most basic version of an \gls{ann} model is a feed-forward neural network. However, there exist other architectures such as \gls{rnn} \citep{williams1989learning, elman1990finding} and \gls{cnn} \citep{lecun1995convolutional}. \gls{rnn}s perform particularly well on sequential data such as in natural language processing (where sentences are considered as sequences of words). Hence, the focus of this work will be on \gls{rnn}s, which are described in detail in Section~\ref{sec:rnns}

\subsection{Introduction to Neural Networks}
\label{sec:dl-intro}
\begin{figure}[!ht]
\centering
  \includegraphics[width=0.5\linewidth]{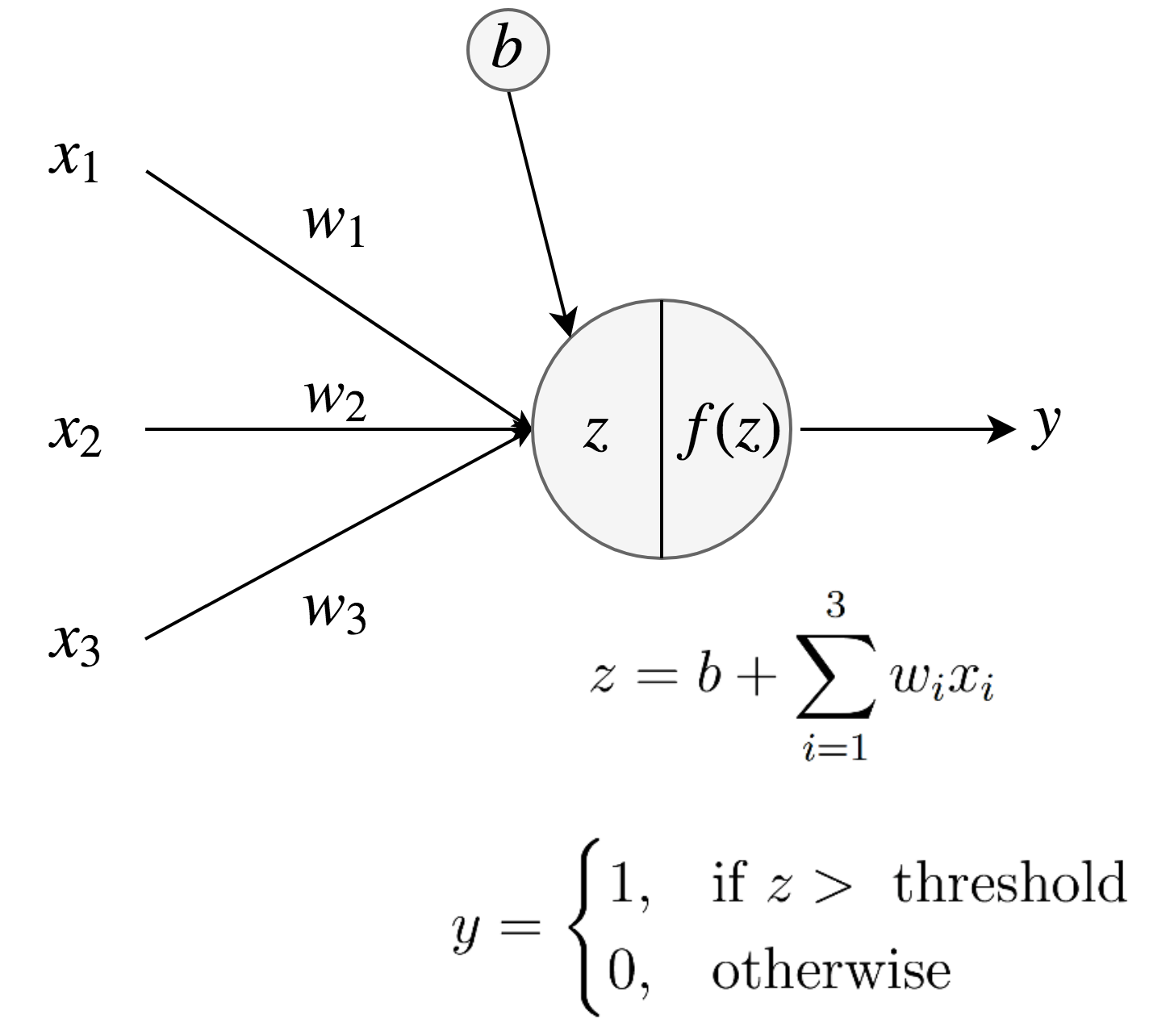}
  \caption{Perceptron Model}
  \label{fig:perceptron}
\end{figure}

In order to understand the computational model of artificial neural networks, one needs to begin from its building block, known as the \textit{perceptron} \citep{rosenblatt1958perceptron}. Inspired from the brain's neurons, a perceptron is a simple computational model that takes in one or more inputs and provides a single value as output. This is illustrated in Figure~\ref{fig:perceptron}. Based on this output and a pre-defined threshold, the perceptron acts as a binary classifier, i.e., if the output value is greater than the threshold, the input is assigned to class 1, else it is assigned to class 0.



Let $x_1, x_2, x_3$ be the inputs to the perceptron model. $w_1, w_2, w_3$ are the series of model weights corresponding to each input variable. This simple model consists of two operations:
\begin{itemize}
\item The first step is to multiply each input with its weight, followed by a summation. To this result, we also add the bias term $b$ so that the model has a flexibility for location shift.
\item Next, we assign a class label (either 0 or 1), based on a binary activation function which requires a pre-defined threshold (refer Figure~\ref{fig:perceptron}). 
\end{itemize} 

The result is the predicted value of output $y$ corresponding to the given set of inputs. In order for the predicted output to be close to the desired output (ground truth), we would need to make adjustments to the weights $w_1, w_2, w_3$ and bias term $b$. 

\begin{figure}[!ht]
\centering
  \includegraphics[width=0.7\linewidth]{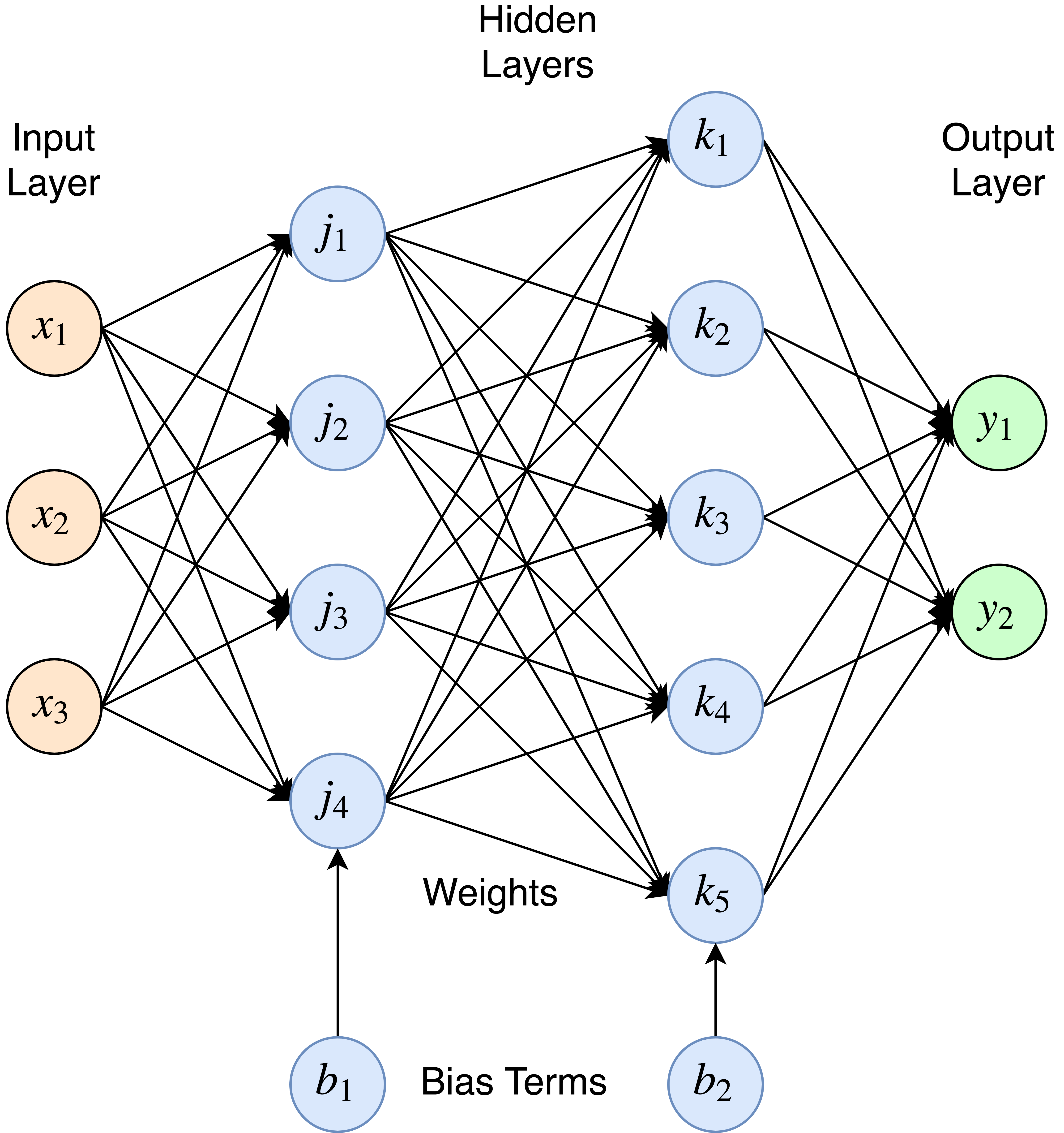}
  \caption{Feed-Forward Neural Network}
  \label{fig:ffnn}
\end{figure}

However, modern neural networks do not use the simple perceptron anymore. Instead, they consist of computational units known as \textit{neurons} (or nodes), which replace the simple binary activation function with non-linear functions such as $\operatorname{sigmoid}$, $\tanh$ or \gls{relu}. It is possible to combine multiple layers of neurons to form a more powerful model known as the feed-forward neural network. Each neuron is connected to every other neuron in the previous and next layer. However, there are no connections between neurons within the same layer. As illustrated in Figure~\ref{fig:ffnn}, there can be multiple inputs and multiple outputs which are connected via a number of \textit{hidden} layers. The input at each neuron gets transformed by weighted summation followed by non-linear activation. The computation happens starting from the input layer, all the way till the output layer and is known as \textbf{forward propagation}. Feed-forward neural network are capable of learning non-linear representations of the data and have been successfully applied to many classification and regression tasks. 

Although artificial neural networks have been around since the 1960s, it was not until the late 1980s that an efficient training procedure for \gls{ann}s was discovered. There existed no structured methodology to adjust the model weights, other than by trial and error. Researchers like \cite{rumelhart1986learning} and \cite{werbos1990backpropagation} contributed to the development of the method known as \textbf{backpropagation of errors}, which made it possible to estimate the weights in an ANN model. Backpropagation makes use of the \textit{chain rule of differentiation}, and computes the gradients in an iterative manner.  

In order to develop an intuition of backpropagation, it is necessary to understand how an optimization method known as \textbf{Gradient Descent} works. In neural networks, we compare the predicted output to the actual output based on a pre-defined loss function. Common examples of loss functions are mean squared error (MSE) and negative log-likelihood (NLL). Our objective is to adjust the model weights in a way that minimizes the loss. It is mathematically guaranteed that moving in the direction of the \textit{gradient} of the loss function (derivative with respect to the model weights), results in loss minimization. 

Assume $\mathcal{L}(\bm w)$ to be the loss function, with $\bm w$ being the model weights. We start with a random initialization of the weights, followed by an iterative update rule as shown in Equation~\ref{eqn:graddescent},

\begin{align}
\label{eqn:graddescent}
\bm w \leftarrow \bm w - \eta \cdot \nabla_{\bm w} \mathcal{L}(\bm w)
\end{align}

\noindent where, $\eta$ is a hyperparameter (set by the user) known as the \textit{learning rate}, which corresponds to the step size towards the local minima in each iteration and $\nabla$ refers to the gradient operator. While a low learning rate results in the training process to progress slowly, a high learning rate may cause the training to diverge from the minima.  Because of this trade-off, the learning rate needs to be set carefully. We stop the iteration process either when we reach the pre-defined maximum number of iterations (known as epochs) or when the change in model weights between iterations is smaller than a specified threshold $\epsilon$. Readers are referred to \cite{bishop2006prml} for further details on backpropagation and Gradient Descent. 

\subsection{Recurrent Neural Networks}
\label{sec:rnns}
One of shortcomings of feed forward neural networks such as the one illustrated in Figure~\ref{fig:ffnn} is that it assumes that all input data are independent of each other. As a result, it fails to capture the notion of sequential order which is present in some types of data.  Consider the task of predicting the next character in a word. If we are given an incomplete word such as `\textit{neura}', one can guess that the next character in the sequence would be `\textit{l}' and the word is `\textit{neural}'. However, if the order of the previous characters was jumbled (such as `\textit{renau}') and provided independently, it would be very difficult to identify the final character. This is where \gls{rnn}s are found to be extremely useful. One of the earliest versions of the recurrent neural network was proposed by \cite{elman1990finding}. The input to an \gls{rnn} is provided in a sequential manner, and the network makes use of the inputs in the previous timesteps in order to make a decision at the current timestep. 

\begin{figure}[!ht]
\centering
  \includegraphics[width=0.7\linewidth]{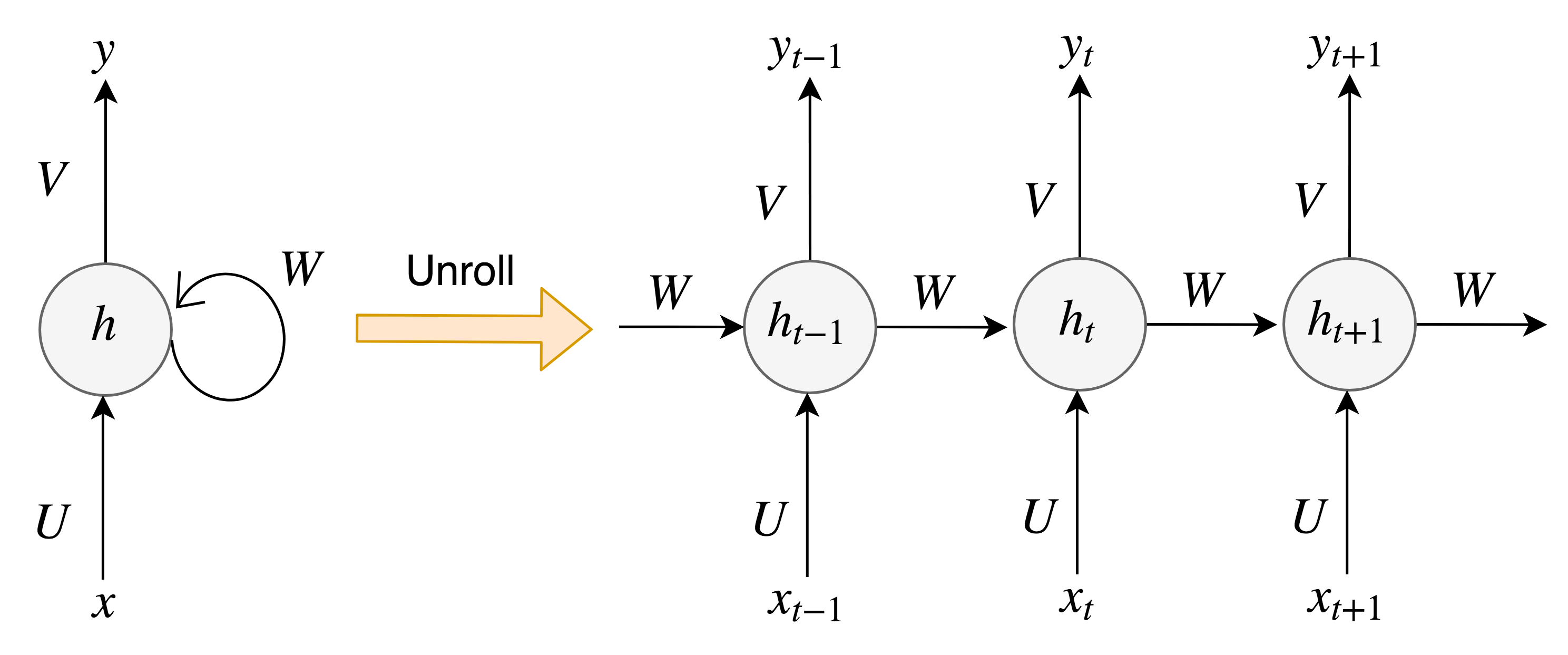}
  \caption{Unrolling of a recurrent neural network (RNN) \citep{britzrnn}}
  \label{fig:rnnfig}
\end{figure}

A recurrent neural network can be depicted as a network with loops (see Figure~\ref{fig:rnnfig}), through which information is transferred between timesteps of the network. By unrolling the network, we realize that the information at each timestep passes through multiple copies of the same network \citep{olahlstm}. 

The notation in Figure~\ref{fig:rnnfig}, adapted from \citep{britzrnn} is described below:
\begin{itemize}
\item $x_t$ corresponds to the input at each timestep $t$
\item $y_t$ refers to the output at each timestep $t$
\item $h_t$ is called the hidden state at each timestep $t$, and is calculated using the input at the current timestep $x_t$ and the hidden state from the previous timestep $h_{t-1}$, i.e., 
\begin{align}
\label{eqn:rnn}
h_t = f(Ux_t + Wh_{t-1})
\end{align}
where $f$ corresponds to some non-linear activation function such as $\tanh$ or \gls{relu}. In RNN literature, $h_t$ is also referred to as the \textit{memory} because in theory, it is assumed to capture information from all previous timesteps. However, this does not hold true in practice since the RNN memory fails to \textit{remember} information beyond few previous timesteps. 
\item $U$, $V$ and $W$ are weight matrices. From the unrolled RNN figure, one can note that these weights are shared across all timesteps of the RNN. Doing this reduces the model complexity by reducing the number of parameters that need to be optimized. Moreover, we aim to perform the same operation across timesteps, just with different inputs. 
\end{itemize}

Training of \gls{rnn}s is done via an extension of the backpropagation algorithm, known as \textit{backpropagation through time }(BPTT). As discussed earlier, RNNs perform well on sequential data and have been extensively used for tasks such as language modelling \citep{mikolov2010recurrent}, text generation \citep{graves2013generating} and speech recognition \citep{graves2013speech}.  

\subsection{Long Short Term Memory}
\label{sec:lstms}
In practice, vanilla RNNs suffer from the inability to capture long term dependencies. In other words, when the length of input sequence becomes large, RNNs are unable to remember the dependencies between inputs which are far apart in the sequence. The reason for this is attributed to the vanishing/exploding gradient problem \citep{pascanu2013difficulty}. This happens due to numeric underflow or overflow, i.e., when the multiplication of derivative terms during backpropagration become extremely small or very large. Exploding gradients can be an easier problem to solve - by truncating gradients when their absolute value crosses a pre-specified threshold \citep{pascanu2012understanding}.

In order to circumvent the issue of vanishing gradients, extensions to the vanilla RNN architecture, \gls{lstm} Units \citep{hochreiter1997long} and \gls{gru} \citep{chung2014empirical} were proposed. This thesis work makes use of RNNs with LSTM units, which will be described in this section. 

\begin{figure}[!ht]
\centering
  \includegraphics[width=0.6\linewidth]{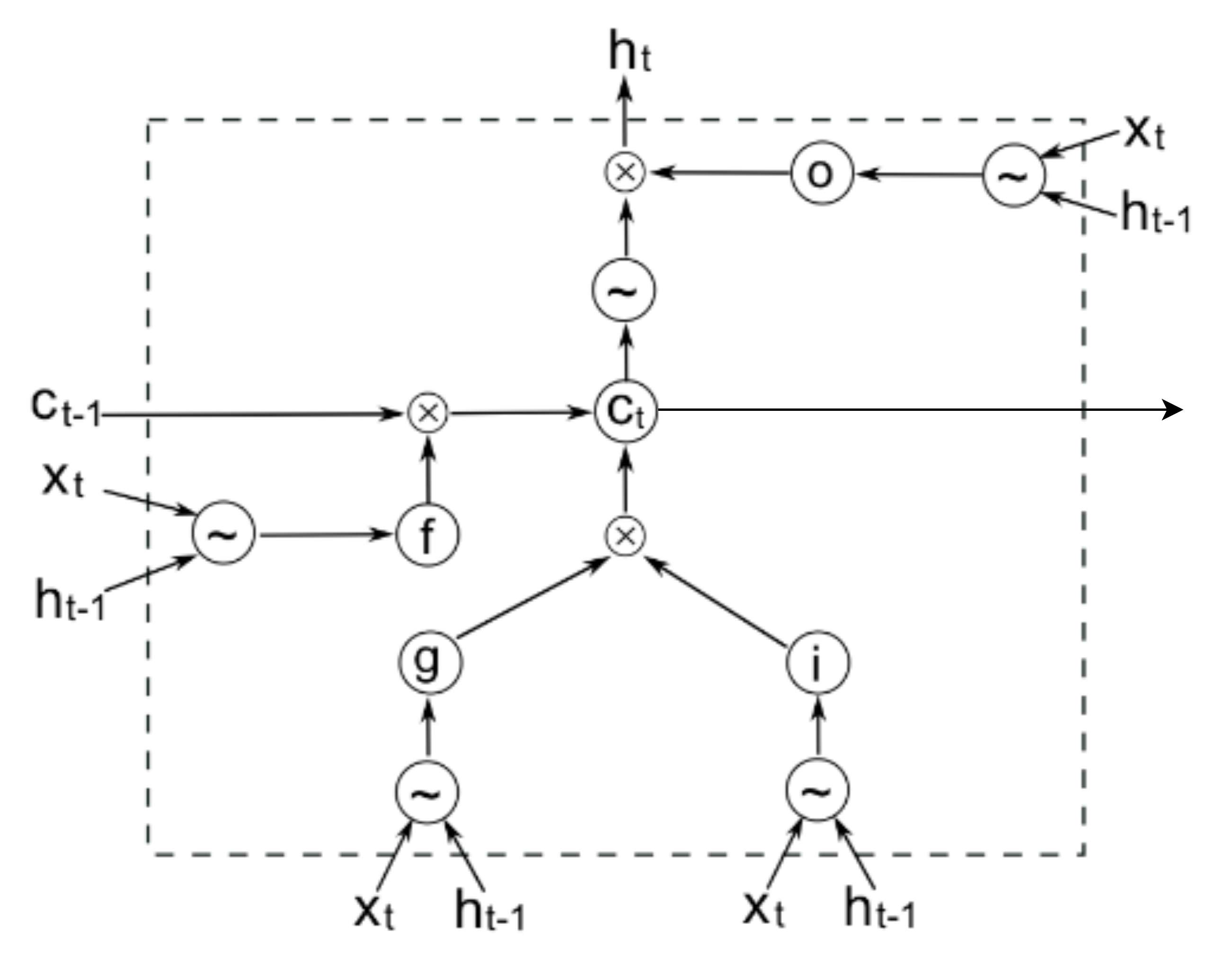}
  \caption{LSTM Unit Representation \citep{xu2015classifying}}
  \label{fig:lstmfig}
\end{figure}

In \gls{lstm}s, we replace the simple activation function $f$ of Equation~\ref{eqn:rnn} with an entire module, also known as \textit{cell}. The input to each repeating module consists of $x_t$ and $h_{t-1}$ along with a new term $c_{t-1}$, known as the cell state. The output at timestep $t$ now includes both $h_{t}$ and $c_{t}$. This is depicted in Figure~\ref{fig:lstmfig}. 

The LSTM unit consists of gated operations and element-wise multiplications. Gates, represented by $sigmoid$ activations (which output values between 0 and 1) essentially decide how much of the incoming information should flow through. The equations of the three adaptive gates namely, the input gate ($\bm i_t$), the forget gate ($\bm f_t$) and the output gate ($\bm o_t$) are given below:

\begin{align}
    \bm i_t = \sigma(W_i \cdot x_t + U_i \cdot h_{t-1} + b_i) \label{eqn:lstminputgate} \\
    \bm f_t = \sigma(W_f \cdot x_t + U_f \cdot h_{t-1} + b_f) \label{eqn:lstmforgetgate} \\
    \bm o_t = \sigma(W_o \cdot x_t + U_o \cdot h_{t-1} + b_o) \label{eqn:lstmoutputgate}
\end{align}

\noindent where $\sigma$ denotes the $\operatorname{sigmoid}$ function.
Each of the gates has its own weights ($U$, $W$ and $b$) and require as input the previous hidden state $h_{t-1}$ and the current input $x_t$. In addition, we also compute a candidate cell state vector ($\bm g_t$) as follows:

\begin{align}
    \bm g_t = \tanh(W_g \cdot x_t + U_g \cdot h_{t-1} + b_g)\label{eqn:lstmcandidatecell}
\end{align}

We now have all the vectors required to determine the cell state ($c_t$) and hidden state ($h_t$) of the current timestep:

\begin{align}
    c_t = \bm i_t \otimes \bm g_t + \bm f_t \otimes c_{t-1} \label{eqn:lstmcellstate} \\
    h_t = \bm o_t \otimes \tanh (c_t) \label{eqn:lstmhiddenstate}
\end{align}

\noindent where $\otimes$ refers to the element-wise multiplication operator. 

\noindent To summarize the procedure, we start by setting the cell state and hidden state of the initial timestep, namely $c_1$ and $h_1$ as zero vectors. Then, the values of $c_t$ and $h_t$ are computed sequentially for each timestep using Equations \ref{eqn:lstminputgate} to \ref{eqn:lstmhiddenstate}, till the end of the sequence is reached.

\gls{lstm}s excel at capturing long term dependencies and have been applied to a number of \gls{nlp} tasks: sequence tagging \citep{huang2015bidirectional}, relationship classification \citep{xu2015classifying}, textual entailment \citep{rocktaschel2015reasoning}, machine comprehension \citep{cheng2016long}, sentiment analysis \citep{tai2015improved} and many more.

\subsubsection{Bi-directional LSTMs}
\label{sec:bilstms}
LSTMs work based on the principle that the output at the current timestep is dependent on the inputs (and outputs) at previous timesteps. However, in some tasks such as part-of-speech (POS) tagging \citep{huang2015bidirectional} and text-to-speech (TTS) synthesis \citep{fan2014tts}, it is useful to be aware of the elements in the future timesteps. Bidirectional LSTMs are known to perform better than regular LSTMs in such scenarios \citep{schuster1997bidirectional}. Essentially, a bidirectional LSTM consists of two LSTMs stacked one on top of the other - the first LSTM reads the sequence in the forward manner (e.g., a regular sentence), while the sequence is fed in the backward direction to the second LSTM (e.g., the same sentence with its word order reversed). The output vectors of the two LSTMs are then concatenated and fed to the next layer in the neural network. 

\subsection{Sequence-to-Sequence Models}
\label{sec:seq2seqmodels}
In natural language processing, sequence-to-sequence tasks usually refer to the ones in which the model takes as input one sequence and generates another sequence as output (instead of a single value). The sequence can range from whole documents to individual words. In the case of documents or sentences, the individual \textit{tokens} that form the sequence are words. In contrast, when words themselves are treated as sequences, their characters become the individual \textit{tokens}. 

\gls{Seq2Seq} models are typically implemented with the help of two recurrent neural networks (usually an \gls{lstm} or \gls{gru}). In the most basic version, the input/source sequence is fed token-by-token to the first \gls{rnn} (encoder), which computes a vector representation for the whole sequence. This vector representation becomes the starting point for the second \gls{rnn} (decoder), which generates the output/target sequence, again in a token-by-token manner. 

\cite{sutskever2014sequence} first introduced \gls{Seq2Seq} models for the task of machine translation. Their LSTM based approach was able to achieve a translation performance close to the state-of-the-art. The advantage of their method is that it could be trained completely end to end (assuming the availability of a parallel corpora), without the need for any manual feature engineering. In order to generate vector representations for sentences, similar to the idea of \texttt{word2vec} \citep{mikolov2013distributed} for words, \cite{kiros2015skip} proposed a \gls{Seq2Seq} learning approach. Essentially the network would be trained to generate a given sentence, based on two of its neighbouring sentences (previous and next). 

In NLP, Seq2Seq models are extensively used for the text generation. \cite{yin2015neural} developed a Seq2Seq model trained on question-answer pairs and knowledge-base triples for the task of answering short factoid questions. Seq2Seq models have opened up the possibility to train dialog sytems in an end-to-end manner, without the need for any hand-crafted features \citep{vinyals2015neural}. Modifications to the original negative log likelihood objective function \citep{li2015diversity} and beam search optimization \citep{wiseman2016sequence} have made dialog systems to be more \textit{human-like}.

\gls{Seq2Seq} models have also been successfully applied to multimodal data. For instance, \cite{venugopalan2015sequence} demonstrate how \gls{Seq2Seq} LSTMs could be used to generate textual descriptions when they are trained with video clips as input. In \citep{yao2015sequence}, the authors develop a model to synthesize speech from textual data. 

Since generation of text is the focus of this thesis work, LSTM based sequence-to-sequence models are used following previous research in this area. 

\subsection{Auto-encoding}
\label{sec:autoencoding}
Autoencoding is an unsupervised learning technique in which we provide an input (such as image or text) to a model, learn an intermediate representation and then try to reconstruct the original input from this representation. The model is usually an artificial neural network, and the intermediate representation typically has lower dimensionality than the original input \citep{hinton2006reducing}. Hence, the goal becomes to learn an efficient encoding that stores just the necessary information required for reconstruction. 

The intermediate representation can later be used for other supervised tasks in the machine learning pipeline. For example, consider the task of recognizing hand written digits. We train an autoencoder with images of digits from 0 to 9. Next, instead of using the original image, we could use their intermediate representations to train a feed forward neural network to classify these digits. Autoencoders have been successfully applied to solve problems such as image super-resolution \citep{zeng2017coupled} and speech enhancement \citep{lu2013speech}. Since the lower dimension representations generated by autoencoders are useful in identifying patterns pertaining to the original data, they have also been applied to anomaly detection \citep{malhotra2016lstm}. 

In the domain of \gls{nlp}, autoencoders are usually \gls{Seq2Seq} models. \gls{rnn}s encode the input sentence into its latent representation. The decoder then uses this representation to reconstruct and generate the original sentence. In \citep{andrew2015semi}, the authors demonstrate how a sequence autoencoder can serve as a `pretraining' method for enhancing the performance of downstream supervised tasks. Autoencoders can be used not just to represent sentences, but also paragraphs and documents \citep{li2015hierarchical}. 

\subsection{Attention Mechanism}
\label{sec:attnmech}
Broadly speaking, attention mechanism in neural networks is a way to guide the training process, by informing the model as to what parts of inputs or features it needs to focus on, in order to accomplish the task at hand. In this section, we will review the attention mechanisms used in NLP. This is different from visual attention in computer vision tasks such as image captioning \citep{xu2015show} and object detection \citep{borji2014look}. 

The two popular attention mechanisms used in \gls{Seq2Seq} models are Luong Attention \citep{luong2015effective} and Bahdanau Attention \citep{bahdanau2014neural}. Both of these models were introduced for machine translation, where they were shown to perform better than vanilla Seq2Seq models. Attention mechanisms achieve this by aligning tokens on the target side to the tokens on the source side. While the core idea behind both attention mechanisms remain the same, the method in which the attention context vector is computed is different - it takes a multiplicative form in Luong Attention whereas in Bahdanau Attention it has an additive form. 

To explain this Seq2Seq attention mechansim intuitively, consider the task of translating the following sentence from French to English: \textit{c'est un chien} $\longrightarrow$ \textit{that is a dog}. During the decoding phase, when we arrive at the timestep that decodes the word \textit{dog}, the model looks at each word on the source side and has to identify that the word \textit{chien}, is where it has to focus on, thereby giving that particular source token the highest weight when computing the attention context vector. The mathematical details pertaining to how attention mechanism works will be detailed in Chapter~\ref{chap:ved}.

Apart from machine translation, attention mechanism has been found to improve the performance of text summarization \citep{rush2015neural}, dialog generation \citep{li2017adversarial}, textual entailment \citep{rocktaschel2015reasoning}, question generation \cite{du2017learning} and so on.  

\subsection{Variational Inference}
\label{sec:varinf}
In variational inference, we use machine learning and optimization to approximate probability distributions which are otherwise difficult to estimate \citep{blei2017variational}. In Bayesian Inference, it is often of interest to compute posterior distributions. This usually involves solving for intractable integrals which becomes cumbersome. In general terms, we try to find an approximate distribution from a family of distributions that is similar to the posterior which we wish to estimate. In other words, we minimize the Kullback-Leibler divergence between the two distributions. 

Although there are methods such as mean field approximation \citep{VI} for variational inference, we will focus on variational auto-encoders (VAE) in this work. In comparison to such traditional methods, \gls{vae}s leverage modern neural networks which are universal function approximators and are a more powerful density estimator. VAEs were first introduced by \cite{kingma2013auto} in the image domain, to learn latent representations for images of handwritten digits. What makes VAEs powerful is that these learnt latent representation (approximately) belong to a pre-defined distribution, such as Gaussian with a known mean and variance. This makes it possible to simply sample a vector from this known distribution and to generate the desired image. It is also possible to manipulate the latent representation to change certain characteristics of the input image. For instance, \cite{deep2015tejas} showed that VAEs could be used as a 3D graphics rendering engine. Specifically, they could manipulate the latent representation of an input image in order to change the pose and orientation of objects within that image. \cite{pu2016variational} train VAEs jointly with images and captions. They demonstrate that the same learnt intermediate representations could be used for a number of downstream supervised tasks including image classification and image captioning. 

The VAEs discussed so far either use MLPs or CNNs as encoders and decoders. In NLP, the straightforward alternative choice is to use RNNs. However, VAEs that use RNNs have been found to be more difficult to train, due to issues relating to the KL divergence between the posterior and prior vanishing to zero. \cite{bowman2015generating} were able to successfully train LSTM-VAEs after implementing optimization strategies such as KL cost annealing and word dropout. In \cite{yang2017improved}, the authors retain an LSTM encoder, but use a CNN decoder for generation of text. In addition, they use dilated convolutions along with residual connections in order to prevent the collapse of the KL term during training. Similar to the image domain VAEs, it is possible to sample from the latent space and generate text. The latent space also exhibits properties such as homotopy \citep{bowman2015generating}, i.e., it is possible to smoothly interpolate between points in the latent space and generate meaningful sentences.

\subsection{Question Generation}
\label{sec:ngenwork}
The task of question generation is as follows: given an input sentence or paragraph, the model is required to generate a question relevant to the input. Such a task would have applications in the field of education to prepare questions relevant to a given passage within a piece of text \citep{heilman2009question}. Question generation could also be used to automatically generate frequently asked questions (FAQs), given product descriptions. 

Question generation is a relatively new research topic. One of the first studies in this area was conducted by \cite{heilman2011automatic}, who defined a set of rules to transform sentences into factoid questions. Another rule-based approach proposed by \cite{chali2015towards} makes use of named entities and semantic role labeling for automatic question generation. Neural networks were implemented for this task only recently by \cite{du2017learning} and \cite{zhou2017neural}. The Seq2Seq \gls{gru} model by \cite{zhou2017neural} generates questions for sentences from the Stanford Question Answering Dataet (SQuAD) \citep{rajpurkar2016squad}. The model requires as input the word embeddings along with lexical and answer position features. The LSTM encoder-decoder model developed by \cite{du2017learning} was evaluated for fluency for both sentence level and paragraph level inputs. This model was trainable completely end-to-end, without the need for any feature engineering. Question generation can also be carried out with knowledge base triples as input \citep{song2016question}. 

\subsection{Dialog Systems}
\label{sec:dialogwork}
Dialog systems (or chatbots) that can converse like humans can be viewed as one of the characteristics of intelligent machines. One of the earlist chatbots, ELIZA was developed by \cite{weizenbaum1966eliza}. ELIZA would provide responses based on a set of pre-defined rules, most of which try to paraphrase the user questions or sentences. ALICE bot \citep{wallace2009anatomy} was an extension to ELIZA, which incorporated more rules and was provided with template responses from more domains \citep{kerly2007bringing}.

However, with the introduction of Seq2Seq models, conversational agents could now be trained end-to-end without the need for any rules \citep{vinyals2015neural}. Neural dialog systems were enhanced by making them context sensitive \citep{serban2016building, sordoni2015neural}, i.e., the conversation history would be provided as input to the model to generate a response. \cite{bordes2016learning} show how deep neural networks can be used to design goal-oriented domain specific dialog systems. The persona-based conversational system developed by \cite{li2016persona} encodes speaker information through a learnt embedding, so that the system is more personalized and provides responses that are consistent. Recent studies in neural dialog generation also focus on making agent responses more diverse and less generic \citep{li2015diversity}. Generative Adversarial Networks (GANs) \citep{li2017adversarial} and deep reinforcement learning techniques \citep{li2016deep} have also been implemented for dialog systems. 

\chapter{Sequence to Sequence Models}
\label{chap:seq2seq}
We first introduce the concept of word embeddings, which are a convenient vector representation for text data. In continuation to the \gls{lstm} concepts introduced in the previous chapter, we provide a mathematical description of \gls{Seq2Seq} models. To conclude this chapter, the tools used to build the word embeddings and neural network models are briefly discussed. 

\section{Word Embeddings}
\label{sec:wordemb}
In order to feed textual data into machine learning models, we need to have corresponding numeric representations. There has been multiple methods proposed in the literature to address this problem \citep{mitra2017neural}, such as bag-of-words (BoW) and one-hot representation. However, word embeddings such as GloVe \citep{pennington2014glove} and word2vec \citep{mikolov2013distributed} are the most common way of representing text for deep neural network models used in \gls{nlp}.  

The notion of distributional similarity was introduced by \cite{harris1954distributional}, who stated that ``words that occur in similar contexts would have similar meaning''. For example, the words \textit{sport} and \textit{game} occur in similar contexts in documents and hence, share similarities in their meanings. This means that the numeric or vector representations of these two words should be similar. In NLP, the cosine distance metric is typically used to measure similarity between two vectors.
\begin{align}
    \cos{\theta} = \frac{\Vec{a}\cdot\Vec{b}}{|{\Vec{a}}|\cdot|{\Vec{b}}|} \label{eqn:cosdist}
\end{align}
where $|\cdot|$ refers to the L2-norm of the corresponding vector. A higher value of $\cos{\theta}$ implies that the angle between the two vectors is small, and hence they are more similar and \textit{vice-versa}. 

A word embedding maps words to real valued vectors, W: $words \rightarrow \mathbb{R}^n$, where $n$ is the dimension of each word vector. \cite{mikolov2013distributed} show that word2vec embeddings can be obtained by training a neural network with a single hidden layer, which is provided with one-hot vectors as input. Consider a context window of $m+1$ words where the centre word is called the focus word and the remaining $m$ words are known as the context words. They propose two methods: 1) given the context words, predict the focus word (continuous bag-of-words or CBOW approach); 2) given the focus word, predict the context words (skipgram approach). In both cases, the weight matrix in the hidden layer of this shallow neural network, at the end of training, will be our word embeddings.  

\begin{figure}[!ht]
\centering
  \includegraphics[width=0.6\linewidth]{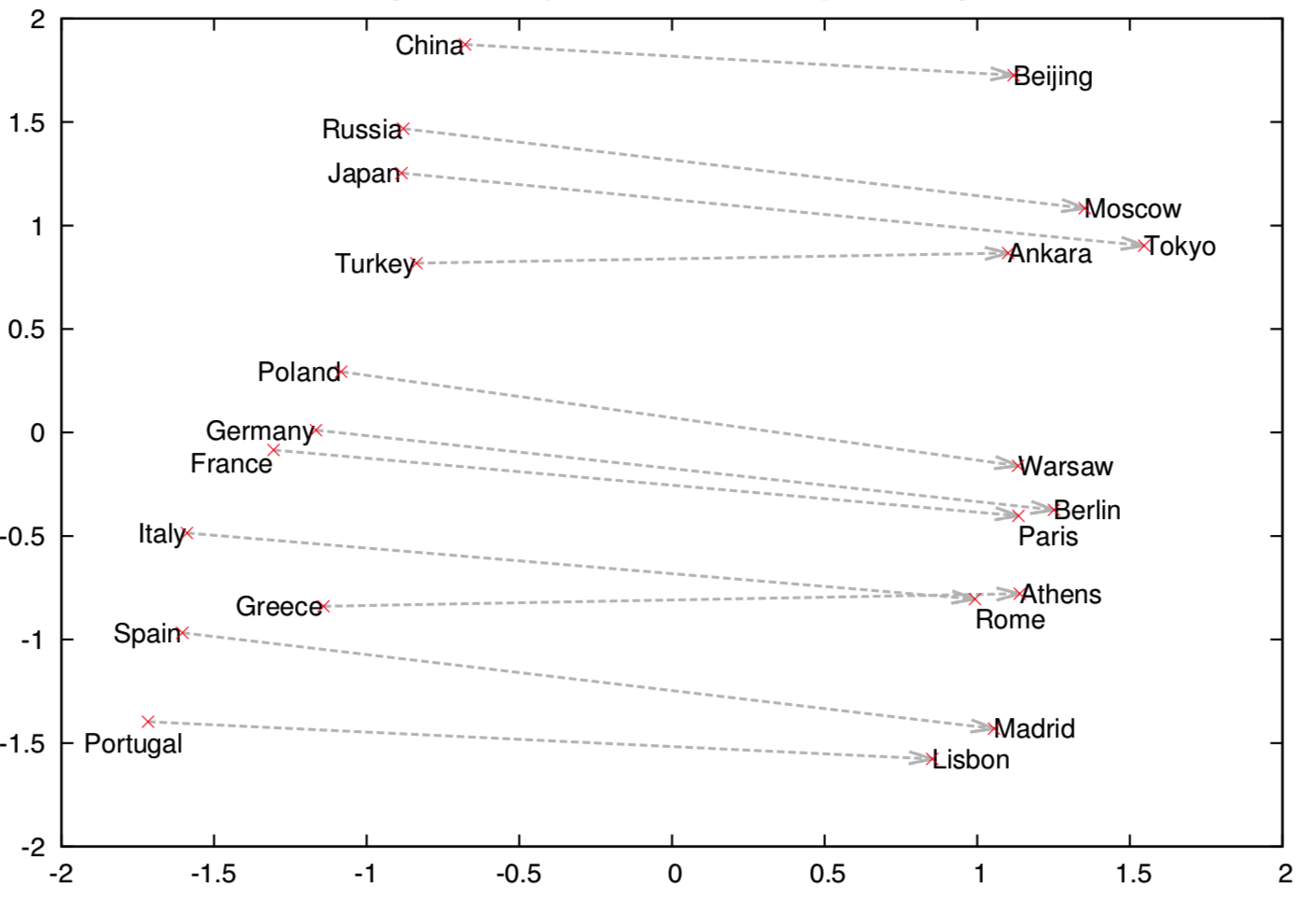}
  \caption{2d PCA projection of 1000d word embeddings of countries and their capitals from \cite{mikolov2013distributed}}
  \label{fig:w2vpca}
\end{figure}

Although this model seems simple, it generates surprisingly meaningful word vectors. The embeddings represent inherent concepts and relationships between words. An example is illustrated in Figure~\ref{fig:w2vpca} taken from \cite{mikolov2013distributed}. It shows how the word vectors for countries and their capitals are organized, when projected onto a 2D surface using Principle Component Analysis (PCA). Another classic example is the following: \textit{king - man + woman $\approx$ queen}. This means that if we have the vectors corresponding to \textit{king}, \textit{man} and \textit{woman}, and if we carry out the arithmetic operation on the LHS, we get a vector which is approximately equal to the vector representation of \textit{queen}. In plain English, this makes sense because a `king' who is not a `man' but a `woman' corresponds to a `queen'. 

Through this example of vector arithmetic, we realize how word embeddings represent semantic information. Word embeddings have been successfully applied to a wide range of NLP tasks due to their compactness (in comparison to one-hot representation) and ability to capture semantic concepts (without any explicit supervision). This study uses only word2vec embeddings and hence the details regarding other embedding models are skipped.

\section{Sequence to Sequence LSTMs}
\label{sec:seqlstms}
Text generation is an important area in NLP. Introduced by \cite{sutskever2014sequence}, sequence-to-sequence models have greatly benefited text generation tasks such as question answering, dialog systems and machine translation. Essentially, these models take one sequence as input and generate another sequence as output. This is in contrast to regular classification models, which output only a single class label, and not an entire sequence. \gls{Seq2Seq} models are typically implemented using two recurrent neural networks, one which is referred to as the encoder and the other is called the decoder. The encoder creates a vector representation of the input sequence that is then fed into the decoder, which then generates tokens in a sequential manner. This study uses the LSTM recurrent neural networks for encoding and decoding sentences. 

\begin{figure}[!ht]
\centering
  \includegraphics[width=0.6\linewidth]{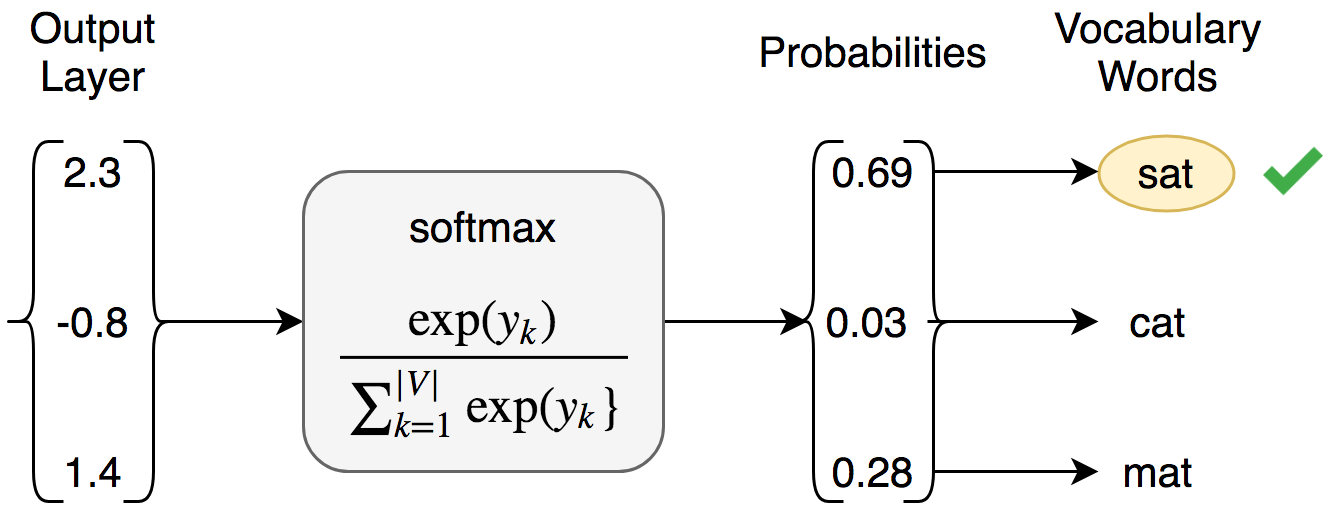}
  \caption{Illustration of the $\operatorname{softmax}$ output layer}
  \label{fig:softmax}
\end{figure}

More concretely, let $\bm x = (x_1, x_2, \cdots, x_{|\bm x|})$ be the tokens (i.e., the corresponding word embeddings) from the source sequence and $\bm y = (y_1, y_2, \cdots, y_{|\bm y|})$ be the tokens from the target sequence. Note that $|\bm x|$ and $|\bm y|$ correspond to the number of tokens (words) in the input and output sequences, respectively. At the end of the encoding process, we will have $\bm h\src_{|\bm x|}$ and $\bm c\src_{|\bm x|}$, the final hidden and cell states respectively from the source sequence (see Section~\ref{sec:lstms}). We then set the initial states ($\bm h\tar_{\bm 1}$ and $\bm c\tar_{\bm 1}$) of the decoder LSTM to $\bm h\src_{|\bm x|}$ and $\bm c\src_{|\bm x|}$. This is a method to transfer information from the source side to the target side, and is known as hidden state initialization. Then, at each further timestep of the decoding process, we compute $\bm h\tar_j$ using an input word embedding $\bm y_{j-1}$ (typically the groundtruth during training and the prediction from the previous timestep during testing). This is given by 
\begin{align}
    \bm h\tar_j = \LSTM_\btheta(\bm h\tar_{j-1}, \bm y_{j-1}) \label{eqn:lstm-htar}
\end{align}
where $\btheta$ refers to the weights of the LSTM network. The predicted word at timestep $j$ is then given by a softmax layer as follows: 
\begin{align}
    p(y_j)=\operatorname{softmax}(W_\text{out}\bm h\tar_j) \label{eqn:decoderout}
\end{align}
where $W_\text{out}$ is a weight matrix and the $\operatorname{softmax}$ function is defined by Equation~\ref{eqn:softmax}
\begin{align}
    \operatorname{softmax}(y_{jk})=\frac{\exp{y_{jk}}}{\sum_{k=1}^{|V|} \exp{y_{jk}}} \label{eqn:softmax}
\end{align}
where $y_{jk}$ refers to the value of the $k$th dimension of the output vector at timestep $j$. In total the output vector at each timestep has $|V|$ dimensions, where $|V|$ corresponds to the vocabulary size. The $\operatorname{softmax}$ layer essentially normalizes the output layer and computes a vector of probabilities. From among the $|V|$ dimensions of the output layer, we pick the dimension with the highest calculated probability and generate the word corresponding to that index. The $\operatorname{softmax}$ operator is demonstrated with an examples in Figure~\ref{fig:softmax}. 

\begin{figure}[!ht]
\centering
  \includegraphics[width=0.8\linewidth]{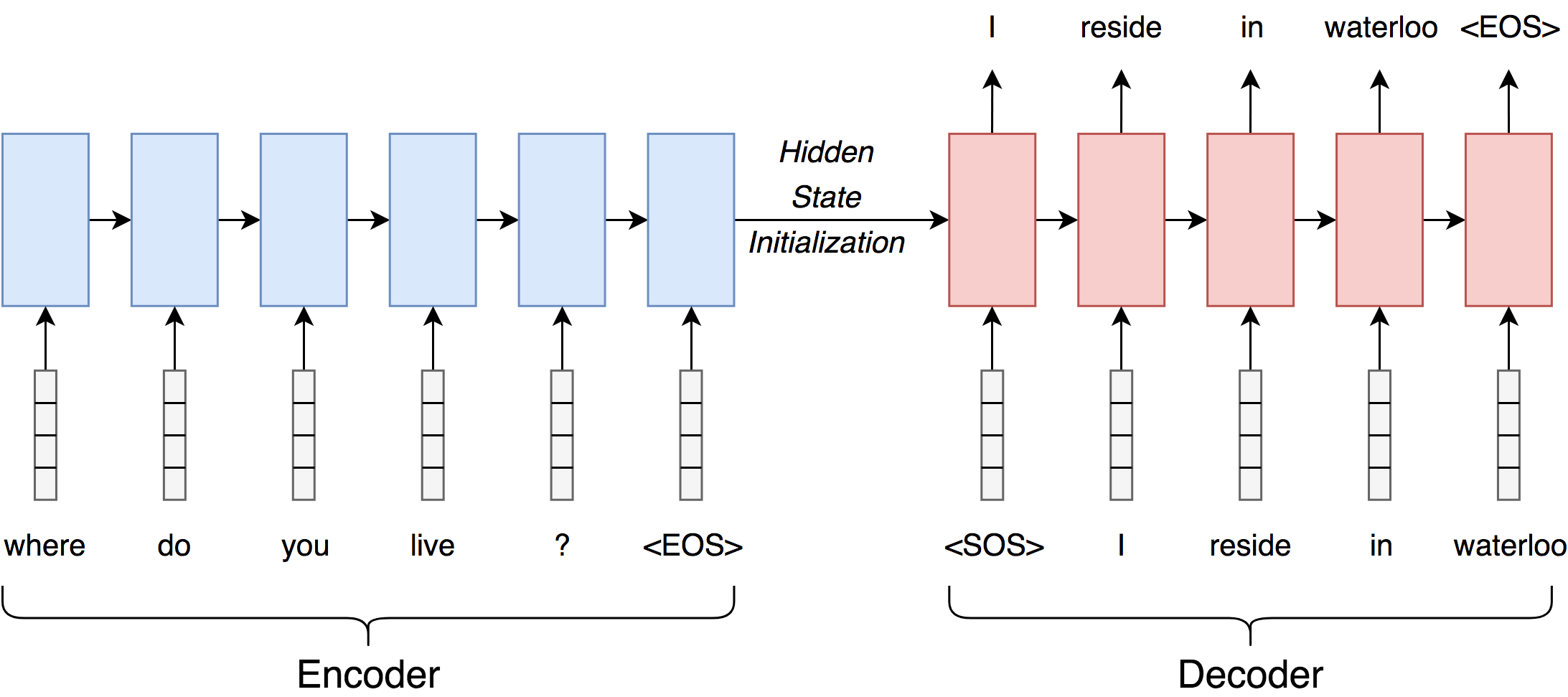}
  \caption{Sequence-to-sequence encoder-decoder model framework}
  \label{fig:seq2seqlstms}
\end{figure}

Figure~\ref{fig:seq2seqlstms} is an illustration of the working of a sequence-to-sequence LSTM model. It has a tokenized input sequence [\textit{where}, \textit{do}, \textit{you}, \textit{live}, \textit{?}] provided to the encoder LSTM. Note that the input at each timestep is the word embedding corresponding to the respective word in the vocabulary. The decoder LSTM generates the outputs [\textit{i}, \textit{reside}, \textit{in}, \textit{waterloo}]. Note the use of special tokens: (1) the start-of-sequence token \textless SOS\textgreater, which signals the decoder LSTM to start the decoding process; and (2) the end-of-sequence token \textless EOS\textgreater, which indicates when to stop decoding. The word embeddings for these can also be learnt either by pretraining the word2vec model or while training the Seq2Seq model.

\section{Environment and Libraries}
\label{sec:envlibs}
Python 3 was used as the programming environment for this project\footnote{Python Software Foundation \url{http://www.python.org}}. For building word vectors, the \texttt{gensim} package \citep{rehurek2010software} in Python was utilized. The deep learning models were implemented using \texttt{keras} \citep{chollet2015keras} and \texttt{tensorflow} \cite{abadi2016tensorflow}. In particular, the \texttt{tf.contrib.seq2seq} module was used for building encoders and decoders and calculating the loss. 

\chapter{Variational Autoencoders}
\label{chap:vae}

\section{Introduction}
\label{sec:intro-vae}
In deep learning, autoencoders \citep{vincent2008extracting, bengio2014deep} can be used to encode high dimensional input data into lower dimensional latent code. Variational autoencoders are a very useful class of models that combine neural networks and variational inference. In Bayesian statistics, it is common to compute posterior probability distributions. Variational inference provides a method to approximate these difficult-to-compute probability distributions through optimization. A known probability distribution of the latent code makes it possible to do generative modelling, i.e., to synthesize new samples (e.g., images) similar to the original data. 
In this chapter, we first provide a short introduction to variational inference. Then we describe the working of variational autoencoders including the reparameterization trick. Finally, we discuss the training difficulties associated with VAEs and empirical results obtained on a natural language dataset.

\section{Variational Inference}
\label{sec:vi}
Consider a latent variable model (a probability distribution over two sets of variables) given by Equation~\ref{eqn:lv-model}
\begin{align}
    p(X,Z) = p(Z)p(X|Z)
    \label{eqn:lv-model}
\end{align}
where, $X$ refers to the observed data and the is $Z$ is the latent variable. In Bayesian modeling, $p(Z)$ is known as the prior distribution of the latent variable and $p(X|Z)$ is the likelihood of the observation $X$ given the latent code $Z$. 

The inference problem in Bayesian statistics refers to computing the posterior distribution, which refers to the conditional density of the latent variables given the data, and is given by $p(Z|X)$ \citep{blei2017variational}. Mathematically, this can be written as follows:

\begin{align}
    p(Z|X) = \frac{p(X,Z)}{p(X)}
    \label{eqn:inf-prob}
\end{align}

The denominator in Equation~\ref{eqn:inf-prob} is the marginal distribution of the data, also called the \textit{evidence}. It is computed by marginalizing out the latent variables from the joint distribution $p(X,Z)$. 

\begin{align}
    p(X) = \int{p(X,Z) dz}
    \label{eqn:marginalizing}
\end{align}

In many cases, the evidence integral is intractable and cannot be computed in closed form \citep{daveVI}.  

In Variational Inference, we obtain an approximate inference solution by trying to find an approximate distribution from a family of distributions that is similar to the posterior which we wish to estimate. In other words, we minimize the Kullback-Leibler divergence between the approximation and the true posterior distribution. The \gls{kl} divergence \citep{kullback1951information} is a statistical method to measure how different two probability distributions are. A lower value of divergence indicates that the distributions are more similar. It is also referred to as relative entropy. Assuming P and Q are two probability distributions, the equation for KL divergence in the discrete case is given by,

\begin{align}
    \KL(P||Q) = \sum_i P(i) \log{\frac{P(i)}{Q(i)}}
\end{align}

\noindent In the continuous case, this can be written as:

\begin{align}
    \KL(P||Q) = \int_{- \infty}^{\infty} p(x) \log{\frac{p(x)}{q(x)}}dx
\end{align}

Returning to the discussion of variational inference, we need to find a distribution over the latent variables, namely $q(Z)$,  from a family of distributions $\tau$ that minimize the KL divergence with respect to the exact posterior $p(Z|X)$. In equation form, this corresponds to

\begin{align}
    q^*(Z) = \argmin_{q(Z) \in \tau} \quad \KL(q(Z)||p(Z|X))
    \label{eqn:kl-optim}
\end{align}

It is to be noted that we may get a better approximation if we choose a more complex family of distributions, at the cost of a more complex optimization process. Also, it is not possible to directly optimize Equation~\ref{eqn:kl-optim} since it contains the term $p(Z|X)$, which was difficult to compute in the first place. Using rules of probability and logarithm, we can rewrite Equation~\ref{eqn:kl-optim} as follows:
\begin{align}
    \KL(q(Z)||p(Z|X)) 
    &= \mathbb{E}_{q(Z)}\left[ \log{\frac{q(Z)}{p(Z|X)}} \right] \nonumber \\
    &= \mathbb{E}_{q(Z)}\left[ \log{q(Z)} \right] - \mathbb{E}_{q(Z)}\left[ \log{p(Z|X)} \right] \nonumber \\
    &= \mathbb{E}_{q(Z)}\left[ \log{q(Z)} \right] - \mathbb{E}_{q(Z)}\left[ \log{\frac{p(Z,X)}{p(X)}} \right] \nonumber \\
    &= \mathbb{E}_{q(Z)}\left[ \log{q(Z)} \right] - \mathbb{E}_{q(Z)}\left[ \log{p(Z,X)} \right] +\mathbb{E}_{q(Z)}\left[ \log{p(X)} \right]
    \label{eqn:kl-plus-elbo-1}
\end{align}
Rearranging the terms, we obtain
\begin{align}
    \left\{ \mathbb{E}_{q(Z)}\left[ \log{p(Z,X)} \right] - \mathbb{E}_{q(Z)}\left[ \log{q(Z)} \right] \right\} + \KL(q(Z)||p(Z|X))  = \log{p(X)}
    \label{eqn:kl-plus-elbo-2}
\end{align}
\begin{align}
    \ELBO(q) + \KL(q(Z)||p(Z|X))  = \log{p(X)}
    \label{eqn:kl-plus-elbo-3}
\end{align}
The first term on the LHS of Equation~\ref{eqn:kl-plus-elbo-2} is known as the variational lower bound \citep{kingmathesis}
The term on the RHS, $\log{p(X)}$, known as log \textit{evidence} is constant with respect to $q(Z)$. Since the variational lower bound is also a lower bound to the log \textit{evidence}, it is also known as \gls{elbo}\citep{yang2017understanding}. That is, $\log{p(X)}\geq\ELBO(q)$ for any $q(Z)$.

\begin{figure}[!ht]
\centering
  \includegraphics[width=0.5\linewidth]{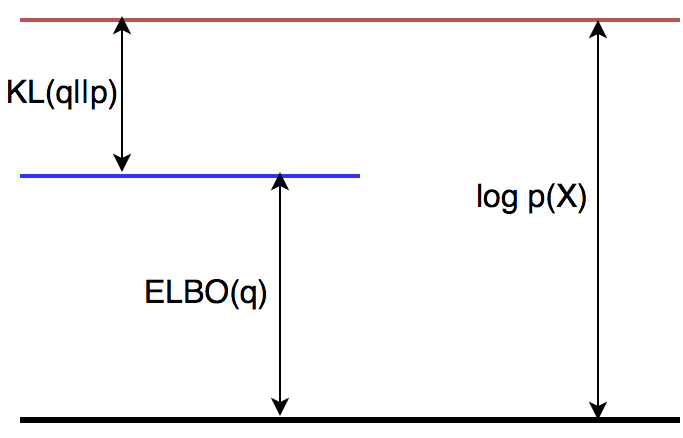}
  \caption{Relationship between $\KL(q(Z)||p(Z|X))$, $\ELBO(q)$ and $\log{p(X)}$ }
  \label{fig:elbo-kl}
\end{figure}

The LHS, which is the sum of the variational lower bound $\ELBO(q)$ and the KL divergence $\KL(q(Z)||p(Z|X))$ adds up to a constant term on the RHS (the log probability of the observations) in Equation~\ref{eqn:kl-plus-elbo-3}. Hence, the objective of minimizing KL divergence is equivalent to maximizing the variational lower bound. This is illustrated in Figure~\ref{fig:elbo-kl}.

We can rewrite the \gls{elbo} into a simpler and more interpretable form as follows
\begin{align}
    \ELBO(q)
    &= \mathbb{E}_{q(Z)}\left[ \log{p(Z,X)} \right] - \mathbb{E}_{q(Z)}\left[ \log{q(Z)} \right] \nonumber \\
    &= \mathbb{E}_{q(Z)}\left[ \log{p(X|Z)p(Z)} \right] - \mathbb{E}_{q(Z)}\left[ \log{q(Z)} \right] \nonumber \\
    &= \mathbb{E}_{q(Z)}\left[ \log{p(X|Z)} \right] + \mathbb{E}_{q(Z)}\left[ \log{p(Z)} \right] - \mathbb{E}_{q(Z)}\left[ \log{q(Z)} \right] \nonumber \\
    &= \mathbb{E}_{q(Z)}\left[ \log{p(X|Z)} \right]  + \mathbb{E}_{q(Z)}\left[ \log{\frac{p(Z)}{q(Z)}} \right] \nonumber \\
    &= \mathbb{E}_{q(Z)}\left[ \log{p(X|Z)} \right]  - \KL(q(Z)||p(Z))
    \label{eqn:elbo}
\end{align}
The first term in Equation~\ref{eqn:elbo} corresponds to the expected log-likelihood of the data. The second term refers to the negative KL divergence between approximate posterior $q(Z)$ and the prior $p(Z)$. With the overall objective to maximize $\ELBO(q)$, we maximize the log-likelihood while encouraging a posterior with a density function that is close to the prior. 

One way to compute the approximate posterior $q^*(Z)$ in Equation~\ref{eqn:kl-optim} is using mean field inference \citep{VI}. The main assumption in the mean field variational family of distributions is that each dimension in the latent code $\mathbf{z}$ is mutually independent and is modelled by its own density function $q_i(z_i)$. The optimization is carried out using Coordinate ascent mean-field variational inference (CAVI) algorithm and readers are referred to \cite{daveVI} and \cite{stephanPGM} for details. This thesis explores how deep learning models can be used to compute approximate posteriors, namely with the help of variational autoencoders. In comparison to traditional methods, the \gls{vae} leverage modern neural networks and is a more powerful density estimator.

\section{Variational Autoencoders}
\label{sec:vae-theory}
Introduced by \cite{kingma2013auto}, variational autoencoders use neural networks to parametrize the density distributions $p$ and $q$, discussed in Section~\ref{sec:vi}. In theory, with sufficient layers used, neural networks can work as universal function approximators, i.e., they can be used to represent any function. 

\begin{figure}[!ht]
\centering
  \includegraphics[width=0.5\linewidth]{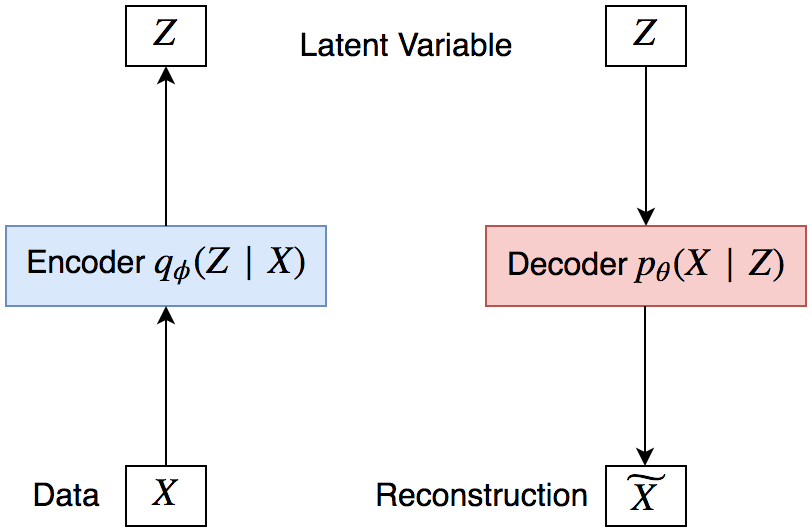}
  \caption{Encoder network for encoding the original data into the latent space and decoder network for reconstruction using the latent representation}
  \label{fig:parametrized-enc-dec}
\end{figure}

In the case of VAEs, neural networks can be used to represent the inference network (the encoder) and the generative network (the decoder) \citep{jaanVAE, stephanPGM}. To compute the approximate posterior $q$, we can design a neural network with parameters $\phi$, called the encoder $q_\phi(Z|X)$. In order to reconstruct the data, we can use another neural network with parameters $\theta$, which is represented as $p_\theta(X|Z)$, referred to as the decoder. This is illustrated with the help of Figure~\ref{fig:parametrized-enc-dec}. As described in Section~\ref{sec:dl-intro}, these parameters correspond to the model weights and biases.

Using the notation described above, we can replace the approximate posterior $q(Z)$ in Section~\ref{sec:vi} to $q_\phi(Z|X)$, the one parametrized by the encoder neural network. Consider a dataset $\mathcal{D}=\{x\n\}_{n=1}^N$, the likelihood of a data point (log \textit{evidence}) and the $\ELBO(q)$ are related as follows: 
\begin{align}
  \log p_\btheta(x\n) &\ge \begin{aligned}[t]
      &\mathbb{E}_{ z\sim q_\bphi( z|x\n)}\left[\log\left\{\dfrac
{p_\btheta(x\n,  z)}
{q_\bphi( z|x\n)}
\right\}\right]\\\nonumber
       \end{aligned}\\
  &= \begin{aligned}[t]
      &\mathbb{E}_{ z\sim q_\bphi( z|x\n)}\left[
\log p_\btheta(x\n| z)
\right]-\KL\left(q_\bphi( z|x\n)\|p( z)\right)
       \end{aligned}\label{eqn:VAElb}
\end{align} 

The above equations essentially rewrite Equations~\ref{eqn:kl-plus-elbo-3} and~\ref{eqn:elbo} in the notation discussed for neural network parameterization of probability density functions. We are required to maximize the $\ELBO(q)$ shown in Equation~\ref{eqn:VAElb}, which is equivalent to minimizing $-\ELBO(q)$. The loss function for the neural network can be written as 
\begin{align}
J\n &= -\mathbb{E}_{ z\sim q_\bphi( z|x\n)}\left[
\log p_\btheta(x\n| z)
\right]+\KL\left(q_\bphi( z|x\n)\|p( z)\right) \nonumber \\ &= J_\text{rec}(\btheta,\bphi,x\n) + \KL\!\left(\!q_\bphi( z|x\n)\|p( z)\!\right)\label{eqn:vae-loss}
\end{align}

The first term, called \textit{reconstruction loss}, is the (expected) negative log-likelihood of data, similar to traditional deterministic autoencoders. For sequence-to-sequence models, this is calculated as a summation of the categorical cross entropy of the prediction across all timesteps of the decoder. The second term refers to the \gls{kl} divergence between the approximate posterior distribution $q_\phi(Z|X)$ that the encoder network maps the original data space into, and the pre-specified prior. In case of continuous latent variables, the prior is typically assumed to be Gaussian $\mathcal{N}(\textbf{0},\textbf{I})$. In this case, the KL divergence denoted as $\KL(\mathcal{N}({\bm \mu_z\n},{\bm \sigma_z\n})\|\mathcal{N}(\textbf{0},\textbf{I}))$ for input $x\n$ is given by the Equation~\ref{eqn:vae-normal-prior} (assuming that only one sample is drawn for each input data point).
\begin{align}
    \frac{1}{2}(1 + \log((\bm \sigma_z\n)^2) - (\bm \mu_z\n)^2 - (\bm \sigma_z\n)^2)
    \label{eqn:vae-normal-prior}
\end{align}

\begin{figure}[!ht]
\centering
  \includegraphics[width=0.45\linewidth]{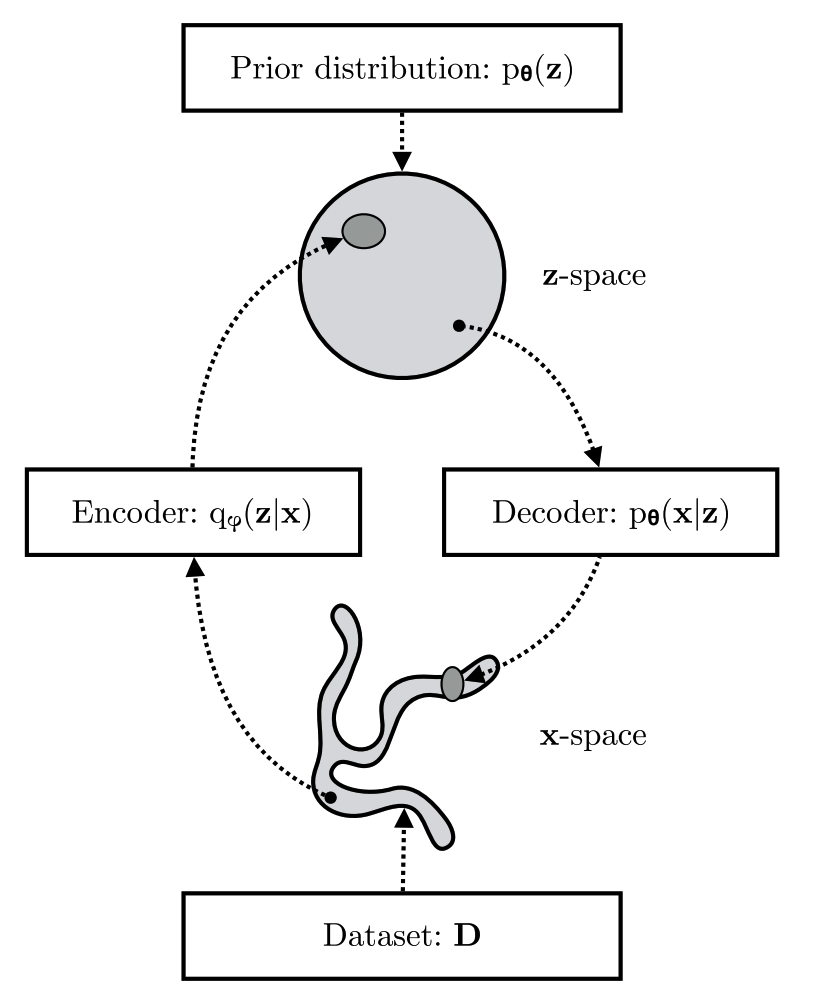}
  \caption{Latent space mappings learnt by a VAE. New data can be synthesized by sampling from the known prior distribution \citep{kingmathesis}}
  \label{fig:vae-latent-space}
\end{figure}

As described in \cite{doersch2016tutorial}, the $\KL$ divergence term can be viewed as a regularization technique (similar to L2-regularization that is used to avoid overfitting). Without the $\KL$ term, the model boils down to a regular \gls{dae}, that encodes the data into a latent space which can be then used for reconstruction. With the $\KL$ term regularization, the latent space is forced into a pre-specified distribution. As a result, the latent space now follows a known distribution, from which one can sample and synthesize new data, such as images \citep{deep2015tejas} and sentences \citep{bowman2015generating}. This is in contrast to traditional DAEs, whose latent space can only be used for reconstruction and does not typically possess any such interesting properties. In other words, DAEs map input data onto arbitrary points on a high dimensional manifold. Whereas, VAEs project data onto continuous ellipsoidal regions that fill the latent space, rather than simply memorizing the arbitrary mappings for the input data \citep{bowman2015generating}. This is depicted in Figure~\ref{fig:vae-latent-space}, where a VAE is used to project the observed data $X$, which has an unknown distribution, into a latent code $Z$ with a known distribution (trained to be approximately similar to that of the prior by using the $\KL$ term regularization). 

\begin{figure}[!ht]
\centering
  \includegraphics[width=0.6\linewidth]{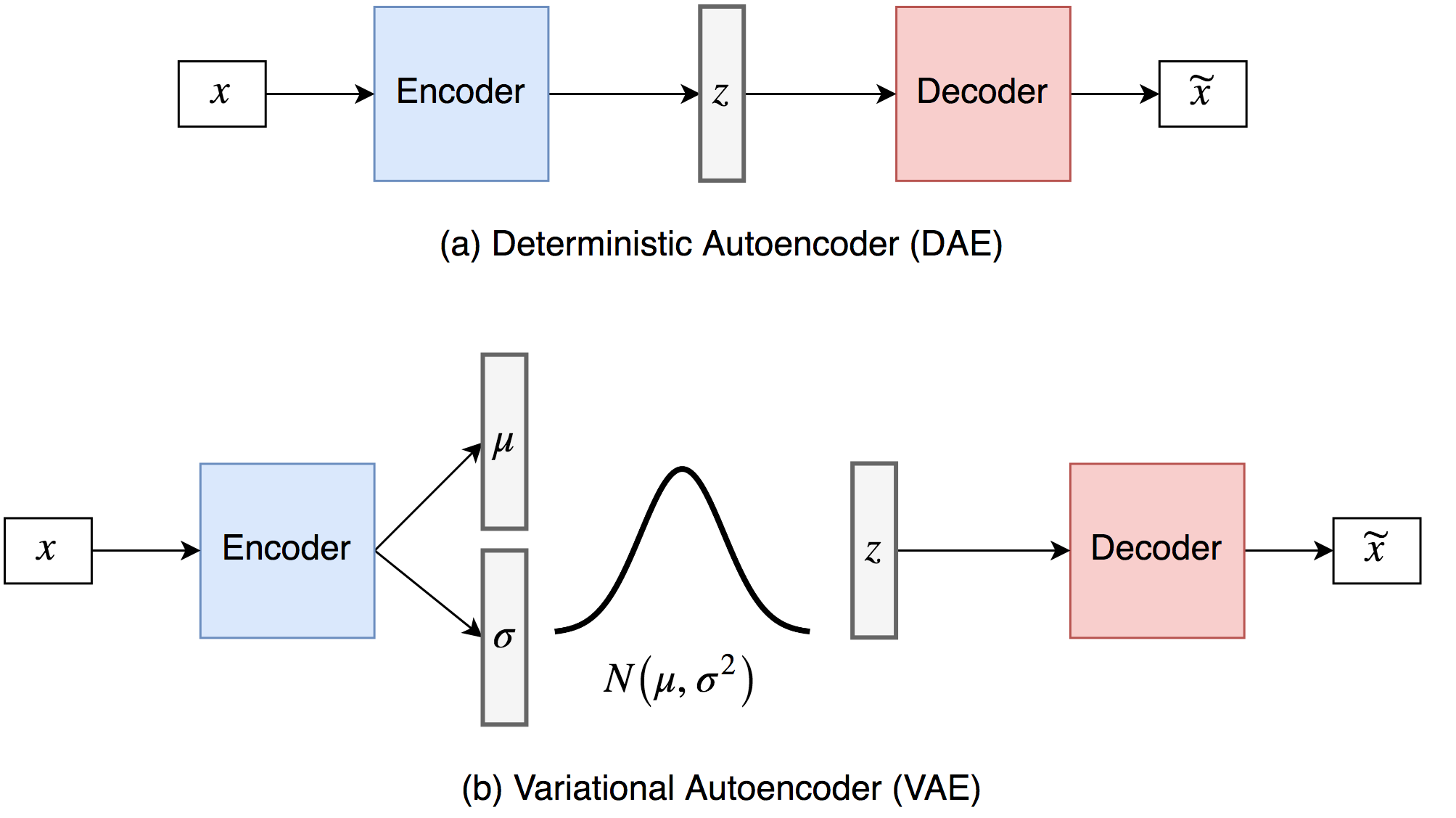}
  \caption{Difference in DAE and VAE architectures}
  \label{fig:daevae-forwad-pass}
\end{figure}

Figure~\ref{fig:daevae-forwad-pass} provides a comparison of the architectures for a DAE versus a VAE. In the traditional DAE, after we learn a latent code $\bm z$, this is directly fed to the decoder. In contrast, for VAEs, we use the encoder outputs to learn the parameters of the underlying posterior distribution. For example, if we assume Gaussian, we would want to learn its mean $\bm \mu$ and variance $\bm \sigma$ (for simplicity we can assume a diagonal covariance matrix $\bm \Sigma$, which implies that the dimensions of the latent code are mutually independent). Next, we can generate a random sample $\bm z$ from this known Gaussian distribution $\mathcal{N}(\bm \mu, \bm \sigma^2)$ and feed it to the decoder. This difference in the procedure during the forward pass is demonstrated in Figure~\ref{fig:daevae-forwad-pass}.

\section{Reparameterization Trick}
\label{sec:reptrick}
\begin{figure}[!ht]
\centering
  \includegraphics[width=0.8\linewidth]{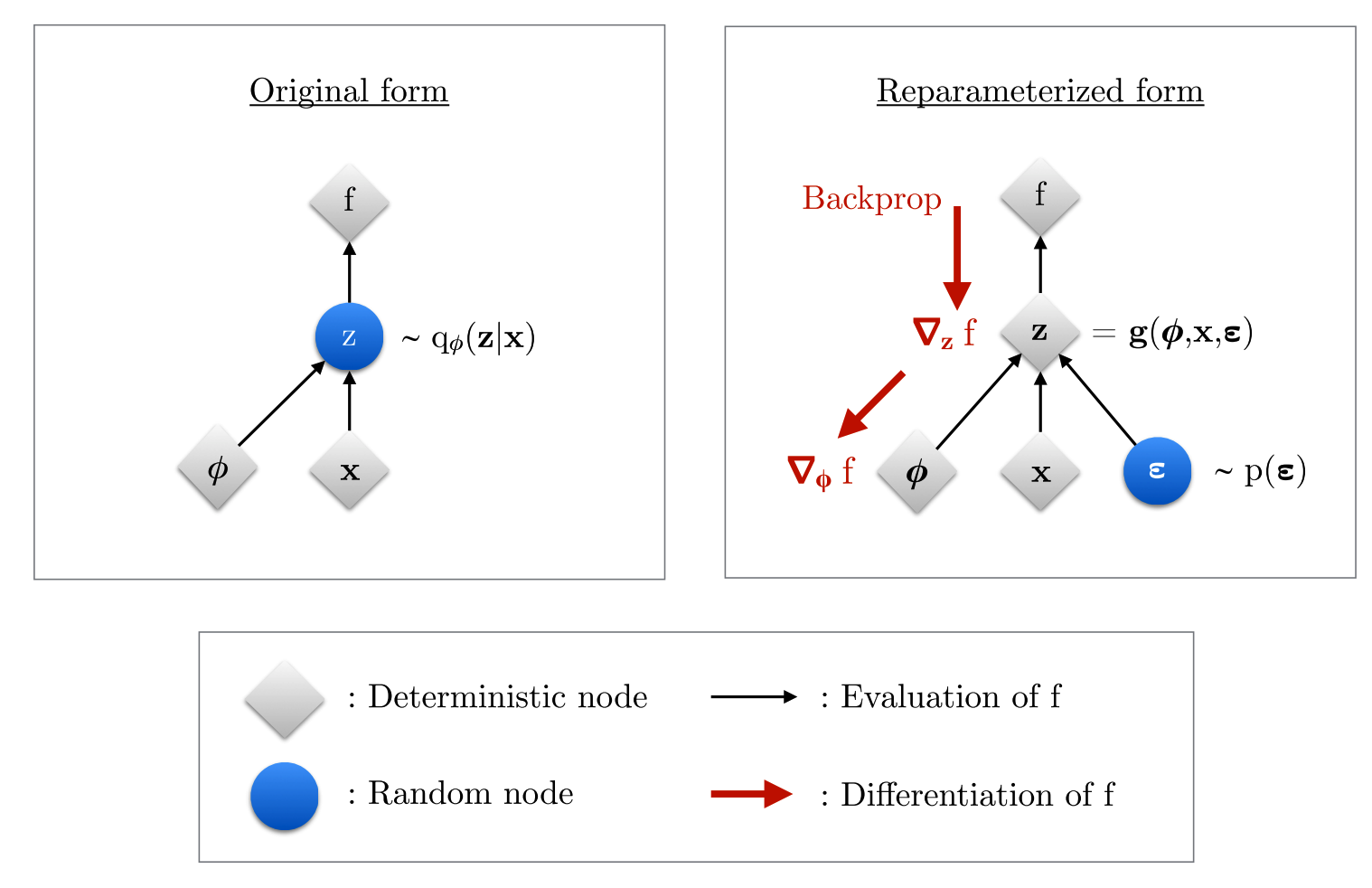}
  \caption{Demonstation of the Reparameterization Trick in VAEs \citep{kingmathesis}}
  \label{fig:reparam}
\end{figure}

Once we have the \gls{vae} network architecture defined, we next focus on training the model using \gls{sgd}. However, this leads to an issue because the model in its original form has a probabilistic node in the computational graph as shown in Figure~\ref{fig:reparam}. Sampling of the latent variable $\bm z$ from the approximate posterior is carried out at the node indicated in blue on the LHS. This stochastic node results in a disconnect in the computational graph and we cannot propagate gradients back to the encoder network. 

In order to circumvent this issue, \cite{kingma2013auto} proposed the \textit{reparameterization} trick. This simple solution involves sampling from a fixed distribution, followed by a variable transformation to the original latent space.

We can consider the example of a Gaussian distribution to illustrate the reparameterization trick. Instead of directly sampling $\bm z$ from its posterior distribution given by $\mathcal{N}(\bm \mu, \bm \sigma^2)$, we can instead sample $\epsilon$ from $\mathcal{N}(\mathbf{0}, \mathbf{I})$ and do a simple variable transformation as shown below:
\begin{align}
    \bm z = \bm \mu + \bm \sigma \otimes \epsilon
    \label{eqn:z-sampled}
\end{align}
where $\bm \mu$ and $\bm \sigma$ have already been obtained by transforming the encoder output (see Figure~\ref{fig:daevae-forwad-pass}). The effect of this reparameterization is that the gradient passes back to the encoder network (through $\bm \mu$ and $\bm \sigma$) and the model is trained end to end. There is no gradient update at the node where we do the sampling of $\epsilon$ from the fixed distribution $\mathcal{N}(\mathbf{0}, \mathbf{I})$. This is exactly what we need as we do not want the fixed sampling to be influenced during gradient propagation. 

\section{Experiments}
\label{sec:vae-experiments}
In this section, the autoencoding experiments with VAEs are detailed. We start with a description of the data and the pre-processing steps. This is followed by the training details including the optimization challenges and the strategies adopted to train the network in a stable manner. 

\subsection{Dataset}
\label{sec:vae-dataset}
For VAE training and hyperparameter tuning, we use the Stanford Natural Language Inference (SNLI) Dataset \citep{bowman2015large}. In order to create the SNLI dataset, the authors adopted the following annotation procedure --- human annotators were shown an image along with its original one-line caption; they were then asked to provide three separate sentences about the scene in the image, corresponding to the following three class labels --— \textit{entailment}, \textit{neutral}, and \textit{contradiction}, for the task of recognizing textual entailment \citep{androutsopoulos2010survey}. The original corpus consists of ~570k pairs of sentences and in this experiment, we use a randomly sampled subset of 80k sentences for training the VAE to carry out reconstruction. 

\subsection{Data Pre-processing}
\label{sec:vae-preprocess}
The following data pre-processing steps are adopted:
\begin{enumerate}
    \item All sentences are converted to lowercase.
    \item All punctuations except \textit{comma} (,) and \textit{full stop} (.) are dropped.
    \item We generate \texttt{word2vec} embeddings for all the words in the subset corpus using the CBOW model (see Section~\ref{sec:wordemb}). The size of the context window was set to 5 and word embedding dimension was chosen to be 300d.
    \item The sentences are then shuffled and a train/validation/test split of 78k/1k/1k is created. 
    \item Each sentence is appended with an end of sequence token $<$EOS$>$.
    \item We then decide on the vocabulary size of the model $|V|$. Only the top $|V|$ most frequently occurring words are retained in the corpus, while the rest are replaced with a special token $<$UNK$>$, referring to words \textit{unknown} in the vocabulary. 
    \item Next, we tokenize sentences into a list of words using the NLTK tokenizer \citep{bird2004nltk}. Each word is then mapped into an integer index. 
    \item Finally, we set the maximum sequence length $m$ (chosen as 10 for this experiment). Sentences with fewer than $m$ words are resized to be of size $m$ by appending a special token named $<$PAD$>$. Sentences with more than $m$ words are trimmed off at $m$ words. 
\end{enumerate}

\section{VAE Optimization Challenges}
\label{sec:vae-optim}
As described in \cite{bowman2015generating}, training VAEs for text generation using RNN encoder-decoder is not straightforward. Optimization challenges associated with the Kullback Leibler (KL) divergence term (between the approximate posterior and the prior) of the loss function, vanishing to zero makes the task of training VAEs notoriously difficult \citep{bowman2015generating, yang2017improved}. When the KL loss is zero, this means that the approximate posterior is exactly the same as the prior. As a consequence of this, the model fails to encode any useful information into the latent space. This causes the model to have a poor reconstruction for any given input. Hence, there is a need to balance the reconstruction term and the KL term of the loss function described in Equation~\ref{eqn:vae-loss}. \cite{bowman2015generating} suggest two strategies to overcome this issue of KL term collapse, which are adopted in this work and are described in the following subsections.

\subsection{KL Cost Annealing}
\label{sec:vae-klanneal}
In this approach, we introduce a coefficient to the KL term of the loss function. This coefficient, referred to as the KL weight is gradually increased (annealed) from zero to a threshold value, as the training progresses. We can rewrite Equation~\ref{eqn:vae-loss} as follows
\begin{align}
J\n &= J_\text{rec}(\btheta,\bphi,x\n) + \lambda \cdot \KL\!\left(\!q_\bphi( z|x\n)\|p( z)\!\right)\label{eqn:vae-anneal}
\end{align}
where $\lambda$ refers to the KL weight, whose value is set to be a function of the iteration number during training. The key idea behind this technique is that we first allow the model to learn to reconstruct the input sentences well, and then we gradually focus on mapping the sentence encodings onto a continuous latent space by making the approximate posterior to be close to the prior. Another way to think of this annealing of the KL weight is that we  gradually transform the model from a completely deterministic autoencoder into a variational one. In this study, we experimentally identify new and improved KL weight annealing schedules, which are discussed in Section~\ref{sec:vae-variants}. We find that the model is sensitive to the rate at which the KL weight is increased and the details are provided in Section~\ref{sec:vae-results}.

\subsection{Word Dropout}
\label{sec:vae-worddrop}
\begin{figure}[!ht]
\centering
  \includegraphics[width=0.5\linewidth]{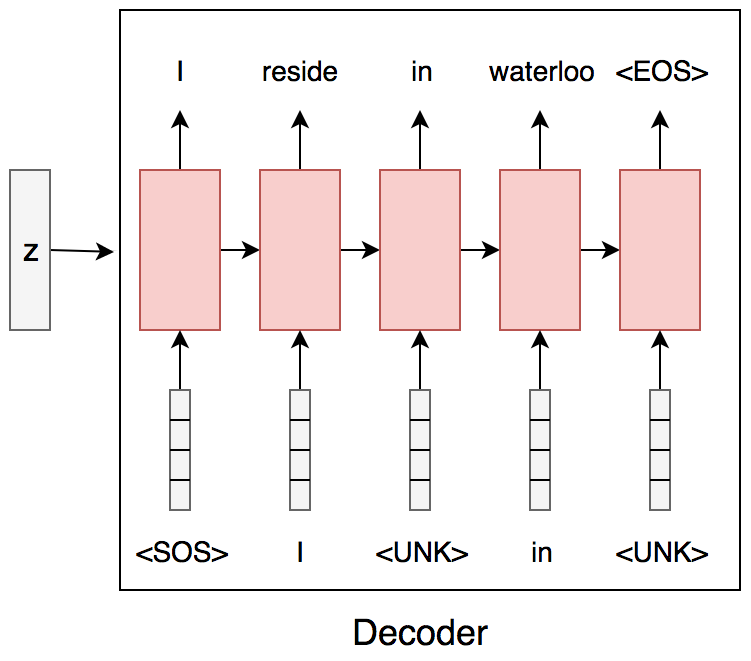}
  \caption{Demonstration of Word Dropout}
  \label{fig:word-dropout}
\end{figure}

The other method to ensure that we learn a useful latent representation is called word dropout. As mentioned in Section~\ref{sec:seqlstms}, we feed the ground truth tokens delayed by one timestep to the decoder during training (see Figure~\ref{fig:seq2seqlstms}). In the word dropout strategy, we replace any given input token to the decoder RNN with the $<$UNK$>$ token with certain probability $p \in \left[0,1\right]$. If $p=0$, the it means that no words are dropped out during decoding. On the other extreme, if $p=1$ we replace each word fed to the decoder with $<$UNK$>$. The $<$UNK$>$ token essentially conveys no information about the source sentence (that is to be reconstructed). Referring to Figure~\ref{fig:word-dropout}, an example input with $p=0.5$ dropout can be [\textit{i}, \textit{$<$UNK$>$}, \textit{in}, \textit{$<$UNK$>$}], i.e., half of the words are replaced with the $<$UNK$>$ token. Doing this weakens the decoder and encourages the model to encode more information in the latent variable $\bm z$ to make accurate reconstructions.

\subsection{Training Details}
\label{sec:vae-training}
For training this sequence-to-sequence variational autoencoder model, we used LSTM units of dimension 100d for both the encoder and the decoder. The dimension of the latent vector $\bm z$ was also chosen to be 100d. We adopted 300d word embeddings \citep{mikolov2013distributed}, pretrained on the 80k subset of the SNLI dataset described in Section~\ref{sec:vae-dataset}. For both the source and target sides, the maximum sequence length was set to be 10. The vocabulary was limited to the most frequent 20k tokens (i.e., $|V|=20000$). The batch size was set to be 32. Following \cite{kingma2013auto}, we choose the standard normal distribution, $\mathcal{N}(\mathbf{0}, \mathbf{I})$ to be the prior.

To learn the model weights, we experiment with both the stochastic gradient descent (SGD) algorithm \citep{sgd} and the Adam optimizer \citep{adam}. For both the optimizers, a constant learning rate of 0.001 was used throughout the training process.

The model is trained for 10 epochs. We observe that the validation set converges at around 10 epochs and hence stop training further. We also compute BLEU (Bilingual Evaluation Understudy) scores on the reconstructed sentences. Originally introduced for automatic evaluation of machine translation systems, BLEU~\citep{papineni2002bleu} scores can be used to assess the sentence reconstruction capability of autoencoders. BLEU-1 measures the unigram overlap between the generated sentence and a set of reference sentences, while penalizing generated sentences that are short. In the same manner, we can determine the bigram, trigram and 4-gram overlap and report BLEU-2, BLEU-3 and BLEU-4 respectively. The automatic evaluation using BLEU scores has been reported to be correlated with human judgment \citep{papineni2002bleu}. For computing, BLEU-\textit{j} based on the \textit{j}-gram overlap (i.e., $\text{precision}_j$), we use the following equation.

\begin{align}
    \text{BLEU-}j = \min\left(1, \frac{\text{generated-length}}{\text{reference-length}}\right)*(\text{precision}_j)
\end{align}

In addition to the ability of a VAE to fluently reconstruct the original input, we also need to assess the quality of the latent space created (refer Figure~\ref{fig:vae-latent-space}). This is done in a qualitative manner by randomly sampling points from the latent space and generating sentences. The exact details will be discussed in Section~\ref{sec:vae-results}. A well trained VAE model should be able to generate new sentences (unseen in the training set) that are both syntactically and semantically correct. 

\subsection{VAE Variants}
\label{sec:vae-variants}
As mentioned in Section~\ref{sec:vae-optim}, training variational autoencoders for probabilistic sequence generation is not very straightforward. We try out multiple settings to train the VAE. Here, we describe and compare five VAE variants which are summarized in Table~\ref{tab:vae-variants}. 

\begin{table}[htbp]
  \centering
    \begin{tabular}{|c|p{27.165em}|}
    \toprule
    \multicolumn{2}{|c|}{\textbf{VAE Variants}} \\
    \midrule\midrule
    ADAM-NoAnneal-$1.0\lambda$ & VAE trained with no KL cost annealing. The KL coefficient ($\lambda$) is set to a constant value of 1.0 throughout training. Optimizer used is ADAM. \\
    \midrule
    ADAM-NoAnneal-$0.001\lambda$ & Same setting as above, except that the $\lambda$ is set to a constant value of 0.001 \\
    \midrule
    ADAM-$\tanh$-$3000$ & KL cost annealing from 0 to 3000 iterations using a rescaled tanh function (refer Eqn~\ref{eqn:vae-tanh}). Optimizer used is ADAM. \\
    \midrule
    SGD-$\tanh$-$3000$ & Same setting as above, but with Stochastic Gradient Descent Optimizer (\gls{sgd}) \\
    \midrule
    ADAM-linear-$10000$ & KL cost annealing from 0 to 10000 iterations in a linear manner (refer Eqn~\ref{eqn:vae-linear}). Optimizer used is ADAM. \\
    \bottomrule
    \end{tabular}%
    \caption{Training VAE with different settings}
  \label{tab:vae-variants}%
\end{table}%

The models ADAM-NoAnneal-$1.0\lambda$ and ADAM-NoAnneal-$0.001\lambda$ are trained with no annealing and no word dropout. ADAM-$\tanh$-$3000$ and ADAM-linear-$10000$ have different annealing schedules, i.e., the rate and function based on which annealing is done. In ADAM-$\tanh$-$3000$, we anneal till 3000 iterations based on Equation~\ref{eqn:vae-tanh}. At this point, the value of $\lambda$ reaches $0.047$, and we continue training with this constant $\lambda$ till model convergence. We have a similar setting in ADAM-linear-$10000$, where the value of $\lambda$ reaches $0.05$ after 10000 iterations (based on Equation~\ref{eqn:vae-linear}) and is then kept constant for the rest of the training. For the final 3 variants in Table~\ref{tab:vae-variants}, word dropout was implemented as follows - at the start of training no words are dropped out ($p=0.0$) and at the end of every epoch, we increase the dropout rate by $0.05$ until it reaches a maximum value of $p=0.5$. The model SGD-$\tanh$-$3000$ is trained with the same settings as ADAM-$\tanh$-$3000$, except that we use an SGD optimizer instead of ADAM.  

\begin{align}
    \lambda_i =  \frac{\tanh{(\frac{i - 4500}{1000})} + 1}{2} 
    \label{eqn:vae-tanh}
\end{align}
    
\begin{align}
    \lambda_i =  \frac{i}{200000}
    \label{eqn:vae-linear}
\end{align}
where $i$ corresponds to the iteration number. 

\begin{figure}
\centering
   \includegraphics[width=1\linewidth]{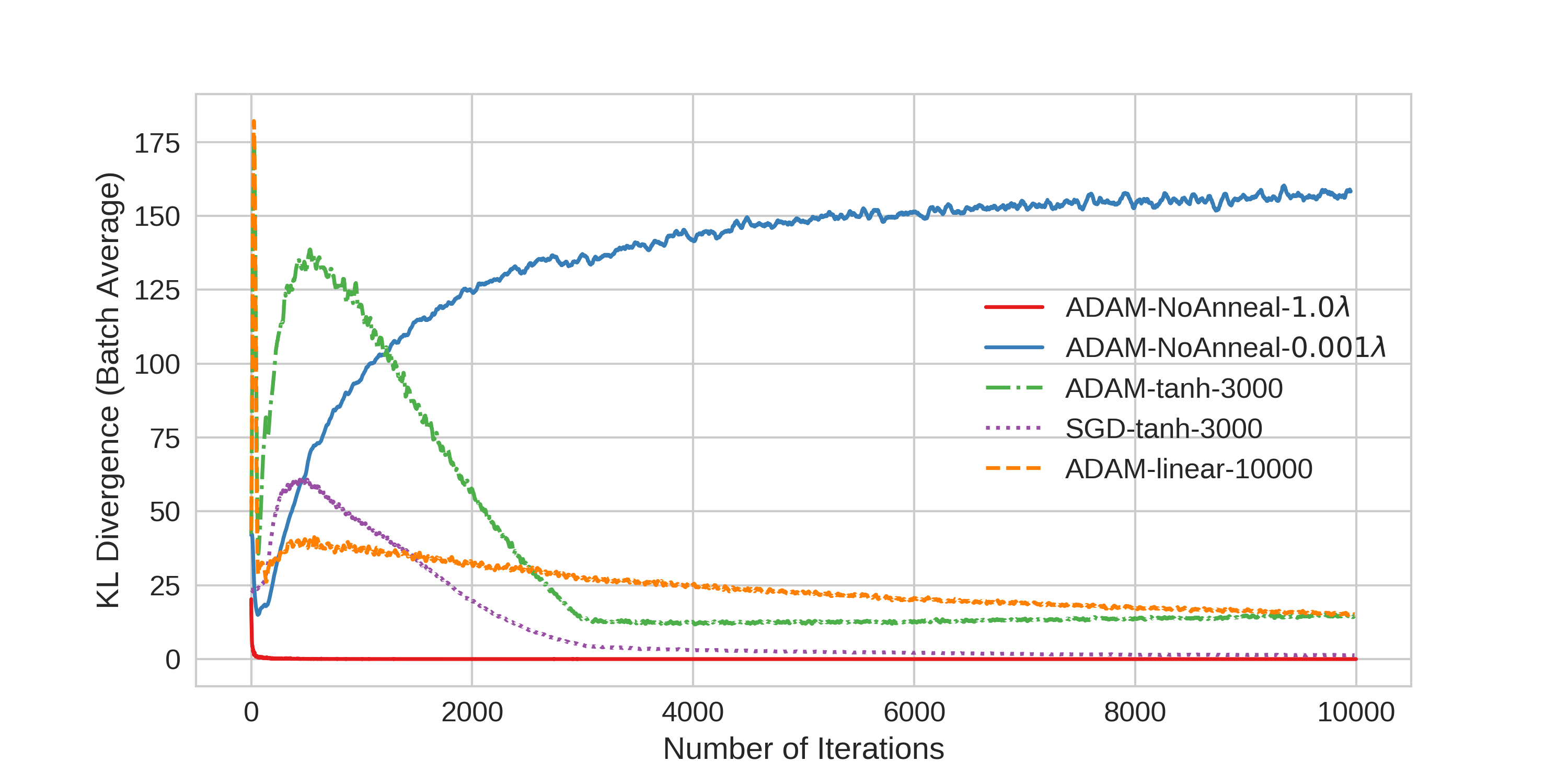}
   \includegraphics[width=1\linewidth]{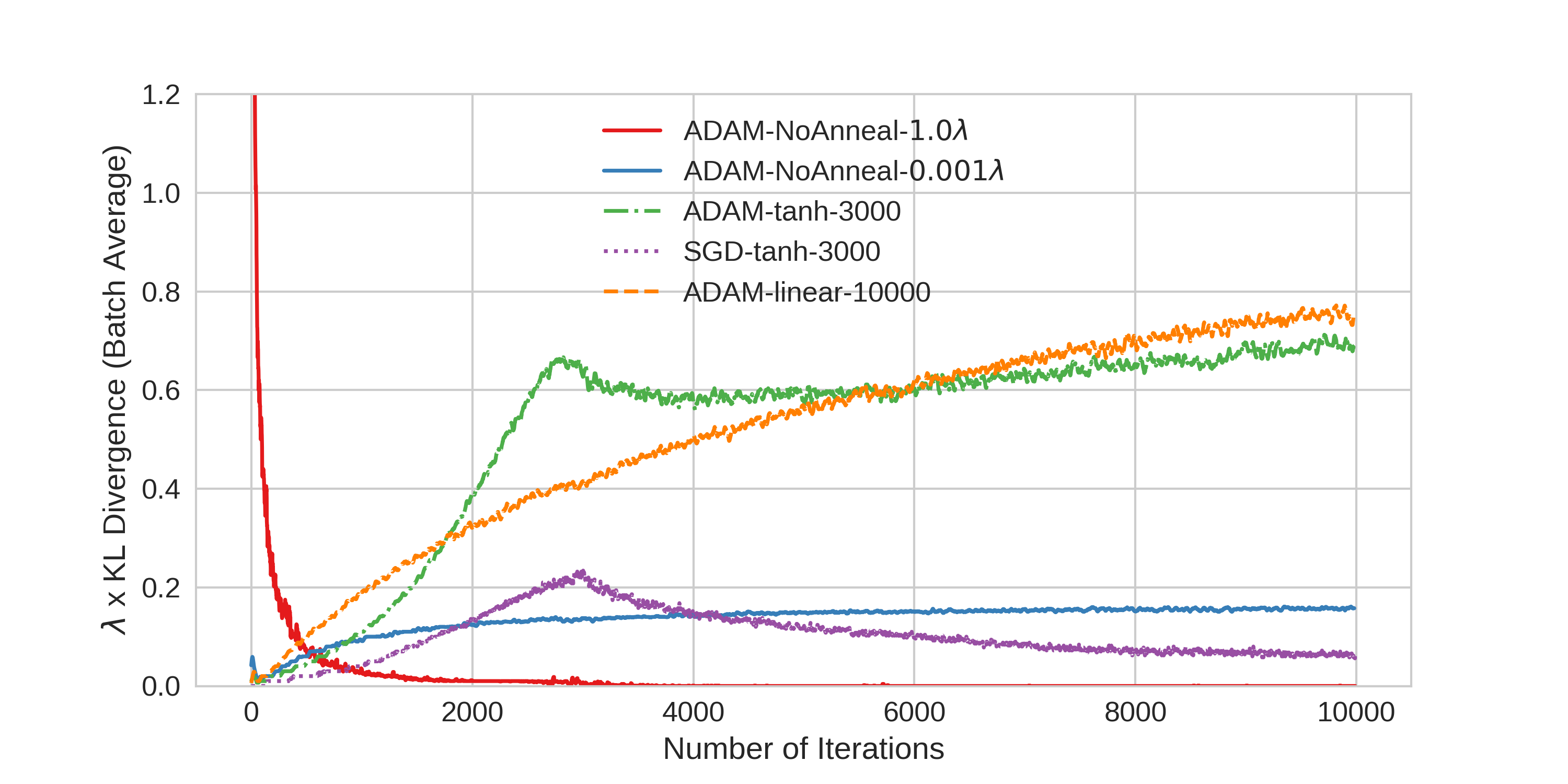}
\caption{Learning curves of the VAE variants. \textbf{Top}: KL divergence, \textbf{Bottom}: $\lambda \times \KL$ divergence. }
\label{fig:vae-kl}
\end{figure}

The learning curves for the different variants described earlier are illustrated in  Figure~\ref{fig:vae-kl}. It can be seen that when there is no annealing procedure in place, the KL loss instantly vanishes to a near zero value within the first few iterations. In contrast, when $\lambda=0.001$, the value of $\lambda \times \KL$ is low and the model has a very small effect of the KL regularizer term. As a result, such a model tends to be more \textit{deterministic} in nature. 

For the models that have annealing in place, the iteration till which annealing is done is an important hyperparameter. The method by which we decide the threshold value till which annealing is carried out is based on the $\lambda \times \KL$ graph. The green line in Figure~\ref{fig:vae-kl} shows that beyond 3000 iterations (approximately), the value of $\lambda \times \KL$ starts to decrease after reaching a maximum value. We realize that if we do not stop annealing at this point, the KL loss steadily decreases further and collapses to zero. We determine that it is ideal to stop annealing once the value of $\lambda \times \KL$ has reached its maximum value. Beyond this point, we can continue with this constant KL coefficient for the rest of the training. For ADAM-$\tanh$-$3000$ and ADAM-linear-$10000$, these values are 0.047 and 0.05 respectively. It can be observed that during later stages of training, the graphs for these two models tend to converge.  When the optimizer is changed to SGD, a completely different pattern is observed. The reason for this unusual trend needs to be investigated further.

\section{Results}
\label{sec:vae-results}

\subsection{Sentence Reconstruction and Random Sampling}
\label{sec:vae-sampling}

The reconstruction performance measured in terms of BLEU scores for the different model variants are listed in Table~\ref{tab:vae-bleu}. In this case, to generate the reconstructed sentence, we feed the mean vector ($\bm \mu$) to the decoder, rather than the sampled $\bm z$ (refer Equation~\ref{eqn:z-sampled}). This is done so that we do not consider any variance in the latent space and pick the most probable value, the mean $\bm \mu$ (referred to as the max \textit{a posteriori}  or MAP estimate).

\begin{figure}[!ht]
\centering
  \includegraphics[width=0.2\linewidth]{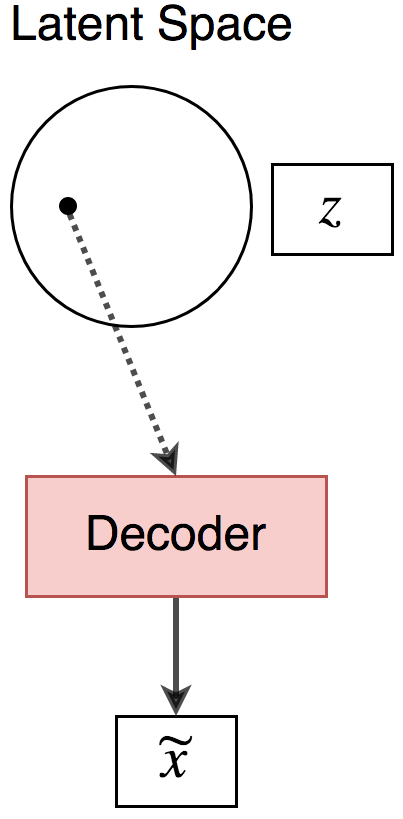}
  \caption{Demonstration of Random Sampling from Latent Space}
  \label{fig:random-sampling}
\end{figure}

In order to assess the quality of the latent space, we randomly sample points from the prior distribution $\mathcal{N}(\mathbf{0}, \mathbf{I})$ and feed the sampled $\bm z$ to the decoder to generate new sentences. This is illustrated in Figure \ref{fig:random-sampling}. If the learnt latent space is continuous, we can expect to generate a meaningful sentence by sampling from anywhere within the latent space. Note that in this setting, the encoder network can be discarded after completion of training. The randomly generated sentences for each VAE variant are shown in Table~\ref{tab:vae-randomsampling}.

For comparison purposes, the BLEU scores and random sentence generations obtained using a deterministic autoencoder (DAE) are also presented. It is to be noted that while \gls{dae}s can give better reconstructions, there is no useful latent space learnt. 

The regular \gls{vae} model trained without any optimization strategies, ADAM-NoAnneal-$1.0\lambda$, produces the same output sentence irrespective of the input. This is due to the KL divergence part of the loss function collapsing to zero which causes the model to have both (1) poor reconstruction capability and (2) poor latent space. 

The observation on training the model with a small constant KL coefficient of 0.001 (ADAM-NoAnneal-$0.001\lambda$) is that the model tends to function like a deterministic autoencoder. This is expected since $\lim \lambda \rightarrow 0$, the model ignores the KL term and becomes a deterministic autoencoder. Although the model has good reconstruction performance indicated by the high BLEU scores, its latent space is not very desirable. The sentences generated are not very meaningful and also not grammatically correct in most cases. 

KL weight annealing and word dropout are indeed very useful training heuristics. This can be seen from the models ADAM-$\tanh$-$3000$ and ADAM-linear-$10000$, both of which have sentences of similar quality being generated from the latent space. The sentences are usually syntactically and semantically correct. They are also diverse, i.e., truly random in the sense that they talk about different topics. In terms of BLEU scores, ADAM-$\tanh$-$3000$ performs relatively better than the linear KL annealing model. However, it is to be noted that when the optimizer was changed to SGD instead of ADAM, the same model turns out to be extremely poor. The sentences generated are more or less the same, i.e., not diverse. The reconstruction capability is only slightly better than the model with the worst performance, ADAM-NoAnneal-$1.0\lambda$. 

By comparing ADAM-NoAnneal-$0.001\lambda$ and ADAM-$\tanh$-$3000$, we realize that reconstruction performance and quality of the latent space are conflicting objectives. The model with better BLEU scores typically results in a non-continuous latent space, generating sentences of lower quality and vice-versa. If our primary objective was just sentence reconstruction, we could simply use a deterministic autoencoder, that can reconstruct sentences in a near perfect manner. From the perspective of probabilistic natural language generation by sampling from a known distribution, ADAM-$\tanh$-$3000$ is a more desirable model. 

\begin{table}[htbp]
  \centering
    \begin{tabular}{|l|c|c|c|c|}
    \toprule
    \multicolumn{1}{|c|}{\textbf{Model}} & \textbf{BLEU-1} & \textbf{BLEU-2} & \textbf{BLEU-3} & \textbf{BLEU-4} \\
    \midrule
    \midrule
    Deterministic AE & 89.56 & 83.22 & 78.29 & 73.73 \\
    \midrule
    ADAM-NoAnneal-1.0 & 26.57 & 10.32 & 4.72  & 2.05 \\
    \midrule
    ADAM-NoAnneal-0.001 & 88.59 & 81.90 & 76.77 & 72.05 \\
    \midrule
    ADAM-tanh-3000 & 66.97 & 53.55 & 44.36 & 36.50 \\
    \midrule
    SGD-tanh-3000 & 32.58 & 13.74 & 6.79  & 2.70 \\
    \midrule
    ADAM-linear-10000 & 65.55 & 52.03 & 42.98 & 35.29 \\
    \bottomrule
    \end{tabular}%
  \label{tab:vae-bleu}%
  \caption{Sentence reconstruction performance for the Deterministic AE and different VAE variants}
\end{table}%

\begin{table}[!th]
  \centering
  \resizebox{\linewidth}{!}{
    \begin{tabular}{|c|c|}
    \toprule
    \textbf{Deterministic AE} & \textbf{ADAM-NoAnneal-1.0} \\
    \midrule
    \midrule
    \textit{a men wears an umbrella waits to} & \textit{a man is sitting on a bench .} \\
    \textit{a couple cows a monument} & \textit{a man is sitting on a bench .} \\
    \textit{some a play mat on the gym} & \textit{a man is sitting on a bench .} \\
    \textit{falling bricks is checking to a tree .} & \textit{a man is sitting on a bench .} \\
    \textit{skate other women .} & \textit{a man is sitting on a bench .} \\
    \textit{a seagull is brown to a sandbox} & \textit{a man is sitting on a bench .} \\
    \textit{a training underwater with the jog down the mountains} & \textit{a man is sitting on a bench .} \\
    \textit{there is sleeping and two rug .} & \textit{a man is sitting on a bench .} \\
    \textit{a man in a pick photos} & \textit{a man is sitting on a bench .} \\
    \textit{a boy are people at a lake escape .} & \textit{a man is sitting on a bench .} \\
    \midrule
    \textbf{ADAM-NoAnneal-0.001} & \textbf{ADAM-tanh-3000} \\
    \midrule
    \midrule
    \textit{two men sit past where the government entering a scene} & \textit{the dog is sleeping in the grass .} \\
    \textit{they are excited formation to ride a castle of a} & \textit{the girls are being detained .} \\
    \textit{their janitor is leaving the dirt wearing his suits .} & \textit{the group of people are going to begin .} \\
    \textit{two children in it exits a} & \textit{a girl with blond-hair on a bike with a stick} \\
    \textit{six people sitting are sorting at single radio in .} & \textit{a woman and a man are walking on a street} \\
    \textit{the guy gets cancer .} & \textit{a brown dog barking at a small dog .} \\
    \textit{i woman who is on watch a factory} & \textit{the people and a woman are giving a speech .} \\
    \textit{three people are wearing overalls .} & \textit{the boy is competing with his athletes .} \\
    \textit{an artist is giving a women in china .} & \textit{a biker is on the racetrack .} \\
    \textit{the camel is pulling on a bull .} & \textit{a man wearing white racing outside .} \\
    \midrule
    \textbf{SGD-tanh-3000} & \textbf{ADAM-linear-10000} \\
    \midrule
    \midrule
    \textit{people are playing in the street .} & \textit{woman talking to the boy .} \\
    \textit{a man is playing a blue shirt and a blue} & \textit{there are boys playing on a trampoline .} \\
    \textit{a man is playing in the street .} & \textit{a group of people are cooking} \\
    \textit{a man is playing in a blue shirt .} & \textit{two girls are cleaning up their beds .} \\
    \textit{the man is in a blue shirt and a blue} & \textit{a child with bare eyes closed and being pulled by} \\
    \textit{the man is playing in the street .} & \textit{the little girl is brushing her teeth .} \\
    \textit{a man is wearing a blue shirt and a blue} & \textit{a few school students riding in a meadow .} \\
    \textit{two men are playing in the street .} & \textit{the woman in the bathtub .} \\
    \textit{a man is wearing a blue shirt and a blue} & \textit{two cyclists ride horses .} \\
    \textit{a man is playing in a blue shirt .} & \textit{some kids are sitting .} \\
    \bottomrule
    \end{tabular}%
    }
    \caption{Generation of random sentences by sampling from the latent space}
  \label{tab:vae-randomsampling}%
\end{table}%

The ADAM-$\tanh$-$3000$ variant has a good balance between reconstruction capability and smoothness of the latent space. Hence we adopt only ADAM-$\tanh$-$3000$ for further discussions and evaluations. We however compare it with the deterministic autoencoder to demonstrate other interesting properties exhibited by VAEs.

\subsection{Linear Interpolation}
\label{sec:vae-linint}
The \gls{vae} objective function simultaneously minimizes the negative log likelihood of the data and the \gls{kl} divergence between the approximate posterior distribution and the pre-specified prior. At the end of training, we can assume that the posterior and prior distributions are approximately close. This allows us to sample from the prior, which is $\mathcal{N}(\mathbf{0}, \mathbf{I})$ in our case, and generate new sentences from the latent space (Section~\ref{sec:vae-sampling}). If the learnt latent space is continuous, then we expect the points sampled from any part of the latent space to generate valid sentences. Another method to determine the continuity of the latent space is using linear interpolation or \textit{homotopy} \citep{bowman2015generating}.

\begin{figure}[!ht]
\centering
  \includegraphics[width=0.2\linewidth]{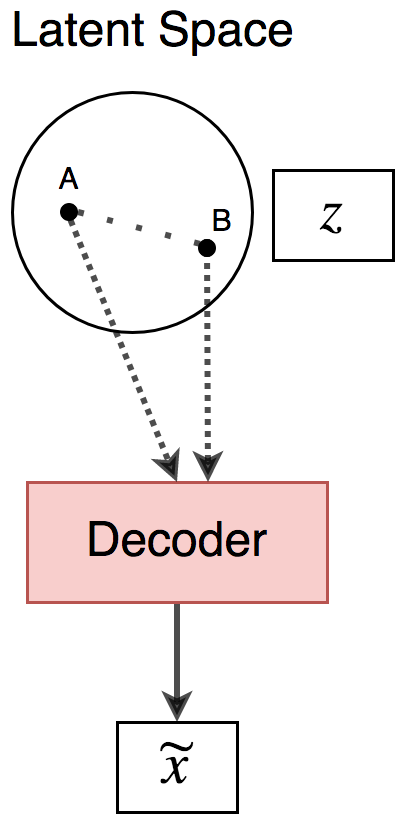}
  \caption{Demonstration of Linear Interpolation between points in the Latent Space}
  \label{fig:linear-int}
\end{figure}

Assume that we are give two input sentences A and B. After mapping them to the latent space, we can obtain their latent representations $\bm z_A$ and $\bm z_B$. For instance, we can linearly interpolate between A and B by manipulating the latent vector as follows:

\begin{align}
    \bm z_{\alpha_i} = \alpha_i \cdot \bm z_A + (1-\alpha_i) \cdot \bm z_B
    \label{eqn:lin-int}
\end{align}
where $\alpha_i \in \left[0, \frac{1}{5}, \frac{2}{5}, \frac{3}{5}, \frac{4}{5}, 1\right]$. This gives us 4 new sentences between A and B. With a good latent space, we expect the transition from A to B to happen in a smooth manner. Each linearly interpolated vector in the latent space should result in a syntactically and semantically correct sentence. The method by which linear interpolation of points in the latent space is carried out is depicted in Figure~\ref{fig:linear-int}.

We carry out the above linear interpolation operation for both the deterministic AE and the variational AE. We can observe from Table~\ref{tab:lin-int} that in the case of the VAE, the transition is smooth. The sentences in between are both fluent and meaningful. Although, one may note that the input sentences A and B have not been perfectly reconstructed in case of the VAE. The deterministic AE on the other hand, has a better reconstruction of the original sentences. However, the intermediate sentences are not grammatically correct or meaningful. The transitions are also observed to be very irregular or non-continuous. Hence, we conclude that inclusion of the KL regularization term in the loss function of the VAE gives rise to a smooth and continuous latent space.

\begin{table}[!th]
  \centering
  \resizebox{\linewidth}{!}{
    \begin{tabular}{|cc|}
    \toprule
    \multicolumn{1}{|c|}{\textbf{Deterministic AE}} & \textbf{VAE ADAM-tanh-3000} \\
    \midrule
    \midrule
    \multicolumn{2}{|c|}{\textbf{Sentence A}: there is a couple eating cake .} \\ \midrule
    \multicolumn{1}{|c|}{} &  \\
    \multicolumn{1}{|l|}{\textit{ there is a couple eating cake .}} & \multicolumn{1}{l|}{\textit{ there is a couple eating cake .}} \\
    \multicolumn{1}{|l|}{\textit{ there is a couple eating cake .}} & \multicolumn{1}{l|}{\textit{ there is a couple eating .}} \\
    \multicolumn{1}{|l|}{\textit{ there is a couple eating cake .}} & \multicolumn{1}{l|}{\textit{ there is a couple eating dinner .}} \\
    \multicolumn{1}{|l|}{\textit{ there is a group of people eating a party .}} & \multicolumn{1}{l|}{\textit{ there is a couple of people eating dinner .}} \\
    \multicolumn{1}{|l|}{\textit{ a group of men are watching a party .}} & \multicolumn{1}{l|}{\textit{ a group of people are having a conversation .}} \\
    \multicolumn{1}{|l|}{\textit{ a group of men are watching a dance party .}} & \multicolumn{1}{l|}{\textit{ a group of men are having a discussion .}} \\
    \multicolumn{1}{|l|}{\textit{ a group of men are watching a dance party .}} & \multicolumn{1}{l|}{\textit{ a group of men are watching a movie .}} \\
    \multicolumn{1}{|l|}{\textit{ a group of men are watching a dance party .}} & \multicolumn{1}{l|}{\textit{ a group of men are watching a movie theater .}} \\
    \multicolumn{1}{|r|}{} &  \\ \midrule
    \multicolumn{2}{|c|}{\textbf{Sentence B}: a group of men are watching a dance party .} \\
    \midrule
    \midrule
    \multicolumn{2}{|c|}{\textbf{Sentence A}: two boys are performing their fencing skills .} \\ \midrule
    \multicolumn{1}{|c|}{} &  \\
    \multicolumn{1}{|l|}{\textit{ two boys are performing their fencing skills .}} & \multicolumn{1}{l|}{\textit{ two boys are performing their martial arts .}} \\
    \multicolumn{1}{|l|}{\textit{ two boys are performing their fencing skills .}} & \multicolumn{1}{l|}{\textit{ two boys are doing their homework .}} \\
    \multicolumn{1}{|l|}{\textit{ two boys are performing their sports skills .}} & \multicolumn{1}{l|}{\textit{ two boys are doing their homework outside .}} \\
    \multicolumn{1}{|l|}{\textit{ two boys are performing chess the last .}} & \multicolumn{1}{l|}{\textit{ two men are playing video games .}} \\
    \multicolumn{1}{|l|}{\textit{ a young man is eating air inside a theater .}} & \multicolumn{1}{l|}{\textit{ a young man is playing golf at a park .}} \\
    \multicolumn{1}{|l|}{\textit{ a white man is eating alone inside a restaurant .}} & \multicolumn{1}{l|}{\textit{ a young man is eating dinner in a restaurant .}} \\
    \multicolumn{1}{|l|}{\textit{ a white man is eating alone inside a restaurant .}} & \multicolumn{1}{l|}{\textit{ a young man is eating inside a restaurant .}} \\
    \multicolumn{1}{|l|}{\textit{ a white man is eating alone inside a restaurant .}} & \multicolumn{1}{l|}{\textit{ a homeless man is eating inside a restaurant .}} \\
    \multicolumn{1}{|r|}{} &  \\ \midrule
    \multicolumn{2}{|c|}{\textbf{Sentence B}: a white man is eating alone inside a restaurant .} \\
    \bottomrule
    \end{tabular}%
    }
    \caption{Linear interpolation between Sentences A and B}
  \label{tab:lin-int}%
\end{table}%

\subsection{Sampling From Neighborhood}
\label{sec:vae-neighbourhood}
In a \gls{vae}, the reparameterization trick uses the mean and standard deviation to compute the latent vector as $\bm z = \bm \mu + \bm \sigma \otimes \epsilon$. In other words, we sample a point in the latent space that is within one standard deviation from the mean. If we sample further away from the mean, we can expect the generated sentence to be more different from the original input which we needed to reconstruct. 

\begin{figure}[!ht]
\centering
  \includegraphics[width=0.2\linewidth]{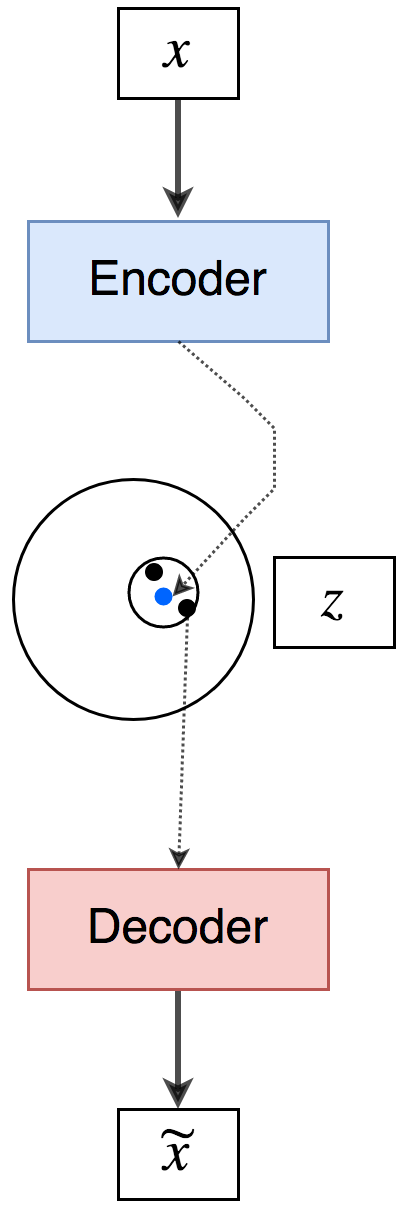}
  \caption{Demonstration of sampling from the neighborhood of a given input $\bm x$}
  \label{fig:neighbourhood-sampling}
\end{figure}

In this experiment, we sample the latent vector as $\bm z = \bm \mu + 3\bm \sigma \otimes \epsilon$, i.e., within 3 standard deviations from the mean. This experiment of sampling from the neighborhood of a given input $\bm x$ is illustrated in Figure~\ref{fig:neighbourhood-sampling}. In a VAE, the latent space is continuous, which means that there are no empty regions. Because of this, the latent vectors sampled from any region will have meaningful representations which can be decoded into sentences. Also, since we sample not too far away from the mean, we can expect the generated sentence to have some topical similarity with the input. This can be observed from the VAE examples in Table~\ref{tab:neighbourhood}, where for a given input, we generate sentences using multiple sampled latent vectors. Each sampled $\bm z$ generates a different sentence, which is however topically similar to the input.

On the other hand, the deterministic autoencoder does not encode input information into a continous space. The latent space in this case is an arbitrary high dimensional manifold. As a result, when we sample points which are 3 standard deviations from the mean, we may end up in empty regions which have no useful latent encodings present. This causes the model to generate the exact same sentence which we would have obtained by feeding the mean vector to the decoder, even when multiple samples are drawn. In other words, the \gls{dae} maps input sentences into points on the high dimensional space that are far apart with empty regions in between. 

\begin{table}[!ht]
  \centering
  \resizebox{\linewidth}{!}{
    \begin{tabular}{|l|l|}
    \toprule
    \multicolumn{1}{|c|}{\textbf{Deterministic AE}} & \multicolumn{1}{c|}{\textbf{VAE ADAM-tanh-3000}} \\
    \midrule
    \midrule
    \multicolumn{2}{|c|}{\textbf{Input Sentence}: a dog with its mouth open is running .} \\
    \midrule
    \textit{ a dog with its mouth is open running .} & \textit{ a dog with long hair is eating .} \\
    \textit{ a dog with its mouth is open running .} & \textit{ a guy and the dogs are holding hands} \\
    \textit{ a dog with its mouth is open running .} & \textit{ a dog with a toy at a rodeo .} \\
    \midrule
    \multicolumn{2}{|c|}{\textbf{Input Sentence}: the man is wearing a black suit .} \\
    \midrule
    \textit{ the man is wearing a black suit .} & \textit{ there are two men in a blue shirt .} \\
    \textit{ the man is wearing a black suit .} & \textit{ the man is drinking from the bar .} \\
    \textit{ the man is wearing a black suit .} & \textit{ the men in a black suit walking through a crosswalk .} \\
    \midrule
    \multicolumn{2}{|c|}{\textbf{Input Sentence}: there are people sitting on the side of the road} \\
    \midrule
    \textit{ there are people sitting on the side of the road} & \textit{ the boy is walking down the street .} \\
    \textit{ there are people sitting on the side of the road} & \textit{ there are people standing on the street outside} \\
    \textit{ there are people sitting on the side of the road} & \textit{ the police are on the street corner .} \\
    \bottomrule
    \end{tabular}%
    }
    \caption{Generating sentences by sampling from the neighbourhood of the mean in the latent space}
  \label{tab:neighbourhood}%
\end{table}%

With the results from the 3 experiments, namely random sampling, linear interpolation and sampling from neighbourhood, we can conclude that the VAEs produce latent spaces that are continuous and characterized by a known distribution. This interesting property allows us to use VAEs for probabilistic generation of text, if trained in a careful manner.

\chapter{Bypassing Phenomenon}
\label{chap:bypass}
In this chapter, we discuss an important design aspect for variational neural network models. Specifically, we introduce a problem which we refer to as \textit{bypassing} connection. We discuss its effects on the performance of variational autoencoders both quantitatively and qualitatively.

\section{Problem Description}
\label{sec:bypass-description}
\begin{figure}[!ht]
\centering
  \includegraphics[width=0.6\linewidth]{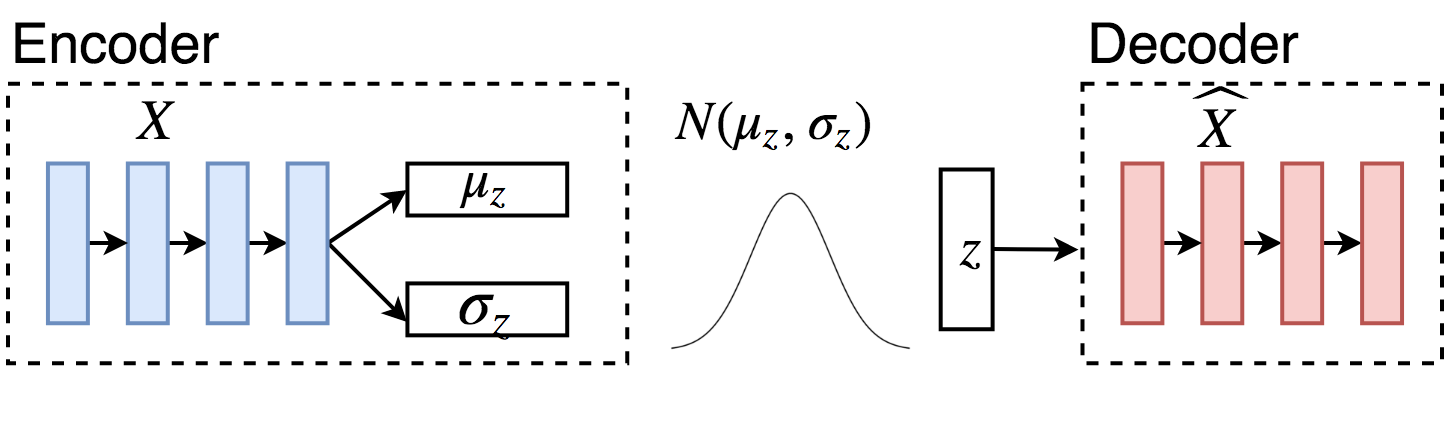}
  \caption{Sequence-to-sequence VAE Architechture}
  \label{fig:vae-original}
\end{figure}

The VAE is essentially an encoder-decoder model. With the help of the learnt latent variable, the decoder can be used as a generative model $p_{\theta}(X|Z)$. The implementation of the VAE neural network architecture consists of sampling latent vectors and feeding them to the decoder network. This has been illustrated with the help of a diagram in Figure~\ref{fig:vae-original}. We use the reparameterization trick to sample from a fixed distribution and carry out a variable transformation using the learnt mean ($\bm \mu$) and standard deviation ($\bm \sigma$). 

\begin{figure}[!ht]
\centering
  \includegraphics[width=0.6\linewidth]{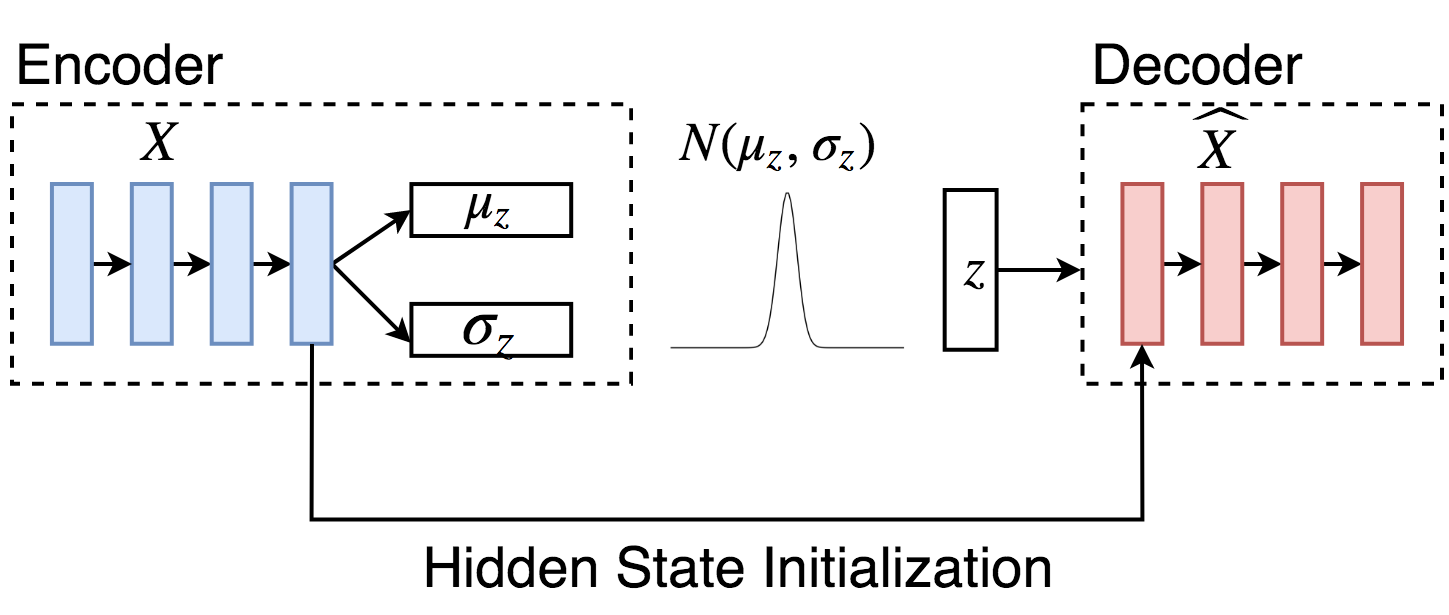}
  \caption{Sequence-to-sequence VAE with bypass}
  \label{fig:vae-hidden-state}
\end{figure}

In regular sequence-to-sequence \citep{sutskever2014sequence} models, it is common to use hidden state initialization to transfer source information to the target side. In LSTM-RNNs, this is done by setting the initial state of the decoder to the encoder's final hidden and cell states. In the \gls{Seq2Seq} literature, the hidden state initialization technique has been extended to train VAEs \citep{serban2017hierarchical, cao2017latent}. This architecture has been illustrated in Figure~\ref{fig:vae-hidden-state}. However, based on our experiments with \gls{vae}s, we argue that such hidden state initializations results in \textit{bypassing} phenomenon. 

We observe that, if the decoder has a direct, deterministic access to the source information, the latent variables $Z$ might not capture much information, which causes the variational space to not play a role in the training process. We call this a \textit{bypassing phenomenon}. This renders the learnt latent space to be inadequate without exhibiting any variational properties such as linear interpolation and sampling from neighbourhood. The distribution of $\bm z$ corresponding to a given input tends to be more peaked with the bypass connection (see Figure~\ref{fig:vae-hidden-state}). In other words, the standard deviation around the mean is very small. As a result, sampling points around the mean, we can generate only the same or very similar sentences. In contrast, VAEs without the bypass connection can generate more diverse sentences that pertain to the same topic as the input. The latent variable distribution is flatter as depicted in Figure~\ref{fig:vae-original}. 

This can also be explained theoretically. Assume that $\hat{X}$ is the reconstruction that needs to be generated using the latent variable $Z$. We can denote the decoder as $p_{\theta}(\hat{X}|Z)$. If the decoder  is provided with a bypass connection to the source $X$, it can now be written as $p_{\theta}(\hat{X}|X,Z)$. Since the latent space is much harder to learn, the decoder in this case can choose to completely ignore $Z$. It can learn to reconstruct the input just using the information from $X$, i.e., $p_{\theta}(\hat{X}|X)$. In this case, the reconstruction loss from Equation~\ref{eqn:vae-loss} can be minimized without the effect of the $\KL$ term. In other words, the KL term fails to act as a regularizer as it can be minimized independently by fitting the posterior to its prior. This will result in the model learning a meaningless latent space that does not encode any useful source information. Hence, a bypass connection degrades a variational \gls{Seq2Seq} model to a deterministic one. We prove this empirically in Section~\ref{sec:bypass-results}. 

\section{Evaluation Metrics}
\label{sec:bypass-evalmetrics}
In a variational model, we learn a continuous latent space. We had qualitatively evaluated the latent space by random sampling and linear interpolation of sentences in Chapter~\ref{chap:vae}. When provided with an input, we expect a VAE to generate sentences that are similar to the input sentence but not necessarily the exact same sentence as output. Because we sample points from around the mean vector corresponding to the input, this can result in some variability in the output. As a result, the generated sentences may be diverse, although they speak about the same topic. Here, we introduce two quantitative metrics to assess the diversity of sentences generated.

\subsection{Entropy}
\label{sec:bypass-entropy}
Assume that for a given input $\bm x$, we generate $k$ outputs $\bm y_1, \bm y_2,..., \bm y_k$ by sampling a new latent variable each time. We can consider this set of $k$ output sentences as our corpus and compute the unigram probability of each token. Note that the unigram probability of token $w$ is calculated by normalizing the count of that particular token by the total number of tokens in the corpus. We can compute the entropy of this unigram probability distribution as
\begin{align}
    H = -\sum_w p(w)\log p(w)
\end{align}
A higher value of entropy corresponds to more randomness in the system. In our case, the \gls{vae} that produces more diverse sentences will have a higher entropy. 

\subsection{Distinct Scores}
\label{sec:bypass-distinctscores}
The \textit{distinct} metrics were introduced by \cite{li2015diversity} for evaluating the diversity of responses provided by neural conversational agents. Similar to the case of entropy calculation, we can generate $k$ sentences for a given input. To compute Distinct-1 and Distinct-2 scores for this set of $k$ sentences, we can use the following equations.
\begin{align}
    \text{Distinct-1} = \frac{\text{Count of distinct unigrams}}{\text{Total unigram count}}\\
    \text{Distinct-2} = \frac{\text{Count of distinct bigrams}}{\text{Total bigram count}}
\end{align}
The higher the distinct score, the more diverse the output sentences will be.

\section{Results}
\label{sec:bypass-results}
\begin{figure}[!t]
\centering
  \includegraphics[width=0.9\linewidth]{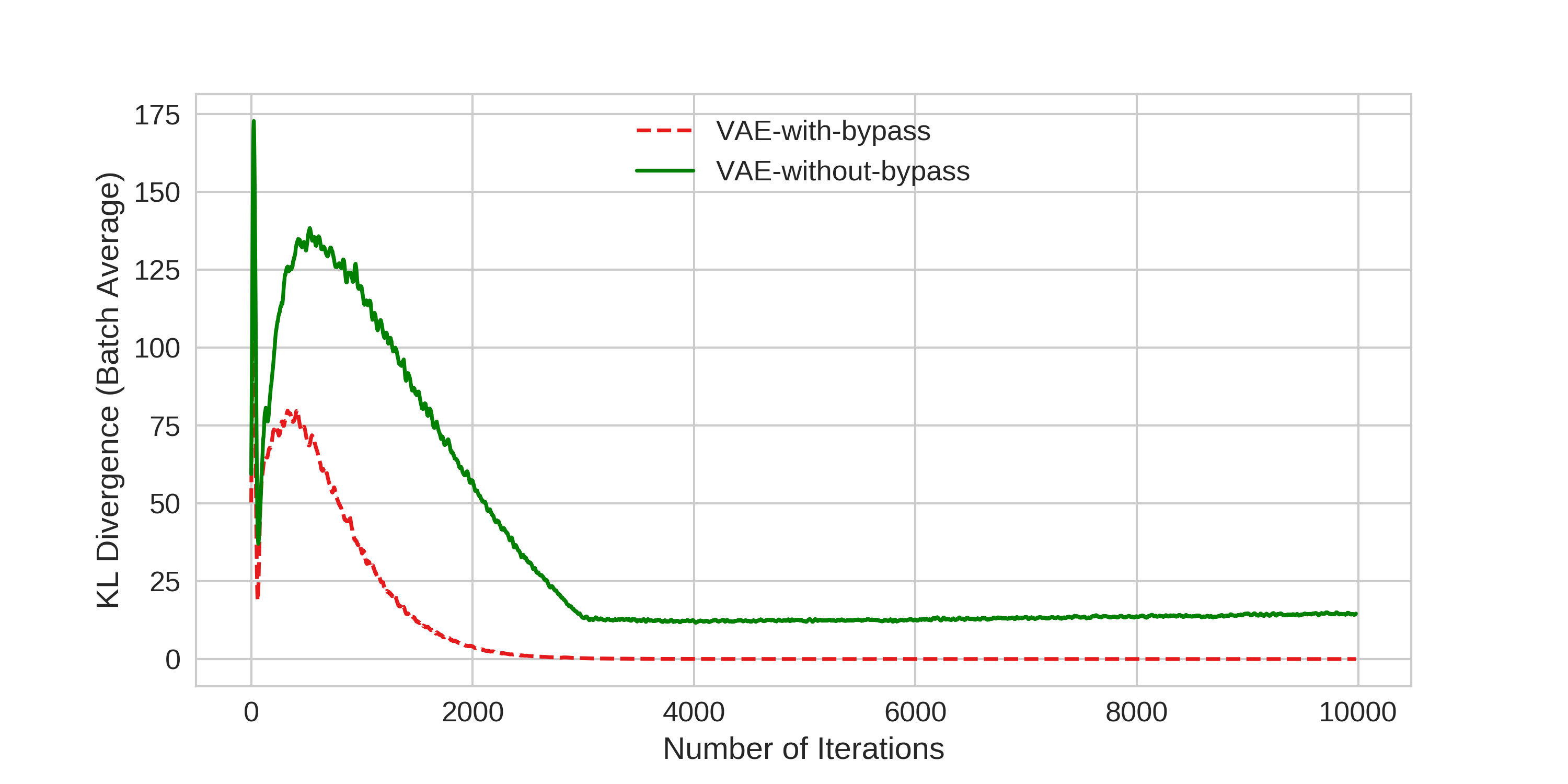}
  \caption{Comparison of learning curves of VAE with and without bypass}
  \label{fig:bypassKL}
\end{figure}

We make use of the best performing \gls{vae} in Section~\ref{sec:vae-results}, namely the ADAM-$\tanh$-$3000$ variant for these experiments. Specifically, we train the exact same model with and without a hidden state initialization of the decoder, which we consider a bypass connection. We observe very different learning curves for the two cases. When the model is trained with a bypass connection, the \gls{kl} term of the loss function gradually vanishes to zero (refer Figure~\ref{fig:bypassKL}). In contrast, when there is no bypass connection, the \gls{vae} can be trained in a more stable manner. 

Due to the differences observed during training, we expect both models to perform differently. We employ the same technique as in Section~\ref{sec:vae-sampling} and sample points within 1 standard deviation around the mean. In this manner, we generate sentences from the latent space conditioned on a given input. We obtain diverse and topically related sentences for the \gls{vae} with no hidden state initialization.  However, the bypass connection in the other model degrades it to a deterministic autoencoder. The latent space is not continuous or smooth. This can be seen from the generated sentences in Table~\ref{tab:bypass-qualitative}, which have little or no diversity. 

We also verify the qualitative findings with the automatic evaluation metrics described in Section~\ref{sec:bypass-evalmetrics}. For each input sentence in the test set, we sample 10 points in the latent space and generate 10 corresponding sentences. We then compute the entropy of each set of 10 sentences using its unigram probability distribution. The average entropy of the outputs generated by the VAE with bypass is much lower than the same model without bypass connection. Note that even a decrease in entropy of 0.68 corresponds to a relatively large difference in diversity because entropy is computed using the log-scale. The same trend is observed for the other diversity metrics, namely Distinct-1 and Distinct-2. The number of distinct unigrams and bigrams produced by the VAE without hidden state initialization is much higher than the VAE with bypass. 

\begin{table}[!t]
  \centering
  \resizebox{\linewidth}{!}{
    \begin{tabular}{|l|l|}
    \toprule
    \multicolumn{1}{|c|}{\textbf{VAE with Bypass}} & \multicolumn{1}{c|}{\textbf{VAE without Bypass}} \\
    \midrule
    \midrule
    \multicolumn{2}{|c|}{\textbf{Input Sentence}: the men are playing musical instruments} \\ 
    \midrule
    \textit{ the men are playing musical instruments} & \textit{ the men are playing video games} \\
    \textit{ the man is playing musical instruments} & \textit{ the men are playing musical instruments} \\
    \textit{ the men are playing musical instruments} & \textit{ the musicians are playing musical instruments} \\
    \midrule
    \multicolumn{2}{|c|}{\textbf{Input Sentence}: a child holds a shovel on the beach .} \\
    \midrule
    \textit{ a child holds a shovel on the beach .} & \textit{ a child playing with the ball on the beach .} \\
    \textit{ a child holds a shovel on the beach .} & \textit{ a child holding a toy on the water .} \\
    \textit{ a child holds a shovel on the beach .} & \textit{ a child holding a toy on the beach .} \\
    \midrule
    \multicolumn{2}{|c|}{\textbf{Input Sentence}: a group of professional football players having a game} \\
    \midrule
    \textit{ a group of professional football players having a game} & \textit{ a group of football players celebrate a game} \\
    \textit{ a group of professional football players having a game} & \textit{ a group of men watching a soccer match} \\
    \textit{ a group of professional football players having a game} & \textit{ a group of football players having a good time} \\
    \bottomrule
    \end{tabular}%
    }
      \caption{Generating sentences conditioned on a given input by sampling from the latent space - comparison of VAE with and without bypass connection}
  \label{tab:bypass-qualitative}%
\end{table}%

\begin{table}[!t]
  \centering
    \begin{tabular}{|r|c|c|}
    \toprule
          & \textbf{VAE with Bypass} & \textbf{VAE without Bypass} \\
    \midrule
    \midrule
    \textbf{Entropy} & 2.004 & 2.686 \\ \hline
    \textbf{Distinct-1} & 0.099 & 0.302 \\ \hline
    \textbf{Distinct-2} & 0.118 & 0.502 \\ \hline
    \bottomrule
    \end{tabular}%
    \caption{Comparison of VAE with and without bypass connection in terms of automatic diversity metrics}
  \label{tab:bypass-quantitative}%
\end{table}%

This observation regarding the bypassing phenomenon sheds light on the design philosophy of variational neural models. This motivates us to further explore possibilities of such bypassing connections that limits the performance of variational encoder-decoder models. We discuss this in the upcoming chapters along with ways to address the issue.

\chapter{Variational Attention for Seq2Seq Models}
\label{chap:ved}

\section{Motivation}
In the previous chapters, the process of autoencoding was discussed. Given an input sequence of words, the task was to learn an intermediate representation, from which it is possible to reconstruct the input. However, in most real life applications we would need to transform a given input sequence into a different output sequence. An example is the task of machine translation, where the input sentence may be in French and we would like to obtain the corresponding translated output in English. Another example is that of a conversational system, where chatbots learn to respond to user inputs. 

In such scenarios, we would need to implement encoder-decoder models (rather than autoencoders). In deep learning, recurrent neural network based sequence-to-sequence models \citep{sutskever2014sequence} are the most popular models used for text generation. They have been extensively used for machine translation \citep{bahdanau2014neural}, dialog systems \citep{vinyals2015neural}, text summarization \citep{rush2015neural} and so on. In most cases, these models tend to be deterministic, i.e., trained to simply maximize the log-likelihood of the data. 

Variational Seq2Seq models have also been applied for encoding-decoding purposes. \cite{serban2017hierarchical} report that when dialog systems are trained with variational neural models, the output responses tend to be longer and more diverse. \cite{zhang2016variational} show that by using variational encoder-decoder (\gls{ved}) models for neural machine translation, they achieve higher BLEU scores than existing deterministic \gls{Seq2Seq} baselines.

The introduction of attention mechanisms \citep{bahdanau2014neural, luong2015effective} resulted in major performance improvements to existing Seq2Seq models. Attention mechanisms essentially align source information on the encoder side to target information on the decoder side. Due to this, the decoding process becomes more accurate by appropriately weighting the source information.  Attention mechanisms in Seq2Seq models are described in detail in Section~\ref{sec:ved-attn-mech}.

However, we argue that traditional attention mechanism cannot be directly applied to variational encoder-decoder models. Doing so unfortunately leads to bypassing phenomenon, similar to the one described in Chapter~\ref{chap:bypass}. In this case, deterministic attention provides direct access to the source information. This may cause the latent space to be ignored during training, resulting in an ineffective variational model. In this chapter we investigate the effect of such a bypassing connection caused by deterministic attention mechanism. We propose an alternative attention mechanism, which we refer to as variational attention that can circumvent this bypassing issue in variational encoder decoder models. 

First, the traditional attention mechanism is introduced and then we discuss how this can be transformed to variational attention. Next, we report empirical results on two experiments - question generation and dialog systems. We prove the advantage of our proposed method by showing that it alleviates the bypassing phenomenon and increases the diversity of generated sentences while maintaining language fluency.

\section{Variational Encoder Decoder}
\label{sec:ved-intro}
\begin{figure}[!t]
\centering
  \includegraphics[width=0.6\linewidth]{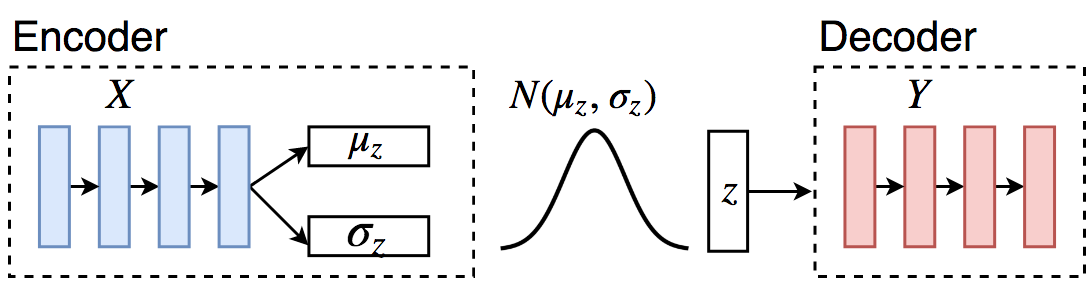}
  \caption{Sequence-to-sequence VED Architecture}
  \label{fig:ved-simple}
\end{figure}

Encoder-decoder models are used when we wish to transform source information $X$ into target information $Y$. To make the model variational, we need to learn a latent variable $Z$ whose distribution (referred to as the posterior) is close to a pre-specified prior. In the literature, different approaches have been proposed to learn the latent space in VEDs. \cite{zhang2016variational} and \cite{cao2017latent} use both $X$ and $Y$ as input variables to encode information into $Z$. In this case, $\bm z$ is sampled from a posterior distribution given by $q_\bphi(Z|X,Y)$, the encoder neural network. However, this causes a discrepancy between training and prediction because $Y$ is not available during the prediction stage. Hence, we follow the approach mentioned in \cite{zhou2017morphological}. Here, the assumption is that $Y$ can be considered as a function of $X$, i.e., $Y = Y(X)$ and as a result, we have $q_\bphi(Z|X,Y)\overset{\Delta}{=}q_\bphi(Z|X,Y(X))\overset{\Delta}{=}q_\bphi(Z|X)$. Hence, we can simply use an encoder that requires only $X$ as input to learn the latent variable $Z$. This is depicted in Figure~\ref{fig:ved-simple}.

\section{Attention Mechanism}
\label{sec:ved-attn-mech}
In deep learning, attention mechanism refers to the techniques by which neural networks are trained to focus on specific aspects of the input in order to accomplish the task at hand. In this section, the attention mechanisms popularly used in \gls{Seq2Seq} models are explained. Note that this is different from visual attention in computer vision tasks such as image captioning \citep{xu2015show} and object detection \citep{borji2014look}, which will not be covered in this thesis.

Consider \gls{rnn}s with \gls{lstm} units as both encoder and decoder. Following the same notation as in Chapter~\ref{chap:seq2seq}, let $\bm x = (x_1, x_2, \cdots, x_{|\bm x|})$ be the tokens from the source sequence and $\bm y = (y_1, y_2, \cdots, y_{|\bm y|})$ be the tokens from the target sequence. At each timestep $j$ of the decoding process, we predict the target token by computing a $\operatorname{softmax}$ across the words in the vocabulary $V$ to obtain probabilities as follows
\begin{align}
    p(y_j)=\operatorname{softmax}(W_\text{out}\bm h\tar_j) \label{eqn:vocabsoftmax}
\end{align}
where, $W_\text{out}$ is a weight matrix, and $\bm h\tar_j$ is the decoder LSTM output at timestep $t$. 
\begin{align}
    \bm h\tar_j = \LSTM_\btheta(\bm h\tar_{j-1}, \bm y_{j-1}, \bm z) \label{eqn:decoder-htar}
\end{align}
where $\btheta$ refers to the weights of the LSTM network, $\bm y_{j-1}$ is the decoder input word embedding (of the output word at the previous timestep) and $\bm z$ is the sampled latent representation of the input sentence $\bm x$

In \gls{ded}, some source information is passed on to the decoder via hidden state initialization (see Section~\ref{sec:bypass-description}). In contrast, it is to be noted that the decoder LSTM in the vanilla Seq2Seq VED does not have access to source information other than through the latent vector $\bm z$. Attention mechanism can serve as a way to learn additional source information in encoder-decoder models. Essentially, during each timestep $j$ of the decoding process, the decoder is provided access to all of the encoded source tokens. More concretely, the decoder at timestep $j$ is provided with the encoder LSTM outputs, i.e., $\bm h\src_{i}$ where $i \in \{1, 2, \cdots, |\bm x|\}$. Now, instead of giving equal weight to every $\bm h\src_{i}$, depending on the target word to be decoded, the decoder LSTM can give higher weight to certain $\bm h\src_{i}$ and lower weight to others. This means that the source outputs can be weighted differently at each decoding timestep. The idea behind this is that all source words need not contribute equally while generating the current target word.

Mathematically, attention mechanisms compute a probabilistic distribution (across the source tokens) at each decoding timestep $j$ given by
\begin{equation}
\alpha_{ji}=\frac{\exp\{\widetilde{\alpha}_{ji}\}}{\sum_{i'=1}^{|\bm x|}\exp\{\widetilde{\alpha}_{ji'}\}}
\end{equation}
where $\alpha_{ji} \in [0,1]$ is the weight given to source output $i$ and $\widetilde{\alpha}_{ji}$ is a pre-normalized score. In the literature of \gls{Seq2Seq} models, two methods have been proposed to calculate $\widetilde{\alpha}_{ji}$. 

\begin{figure}[!t]
\centering
  \includegraphics[width=0.7\linewidth]{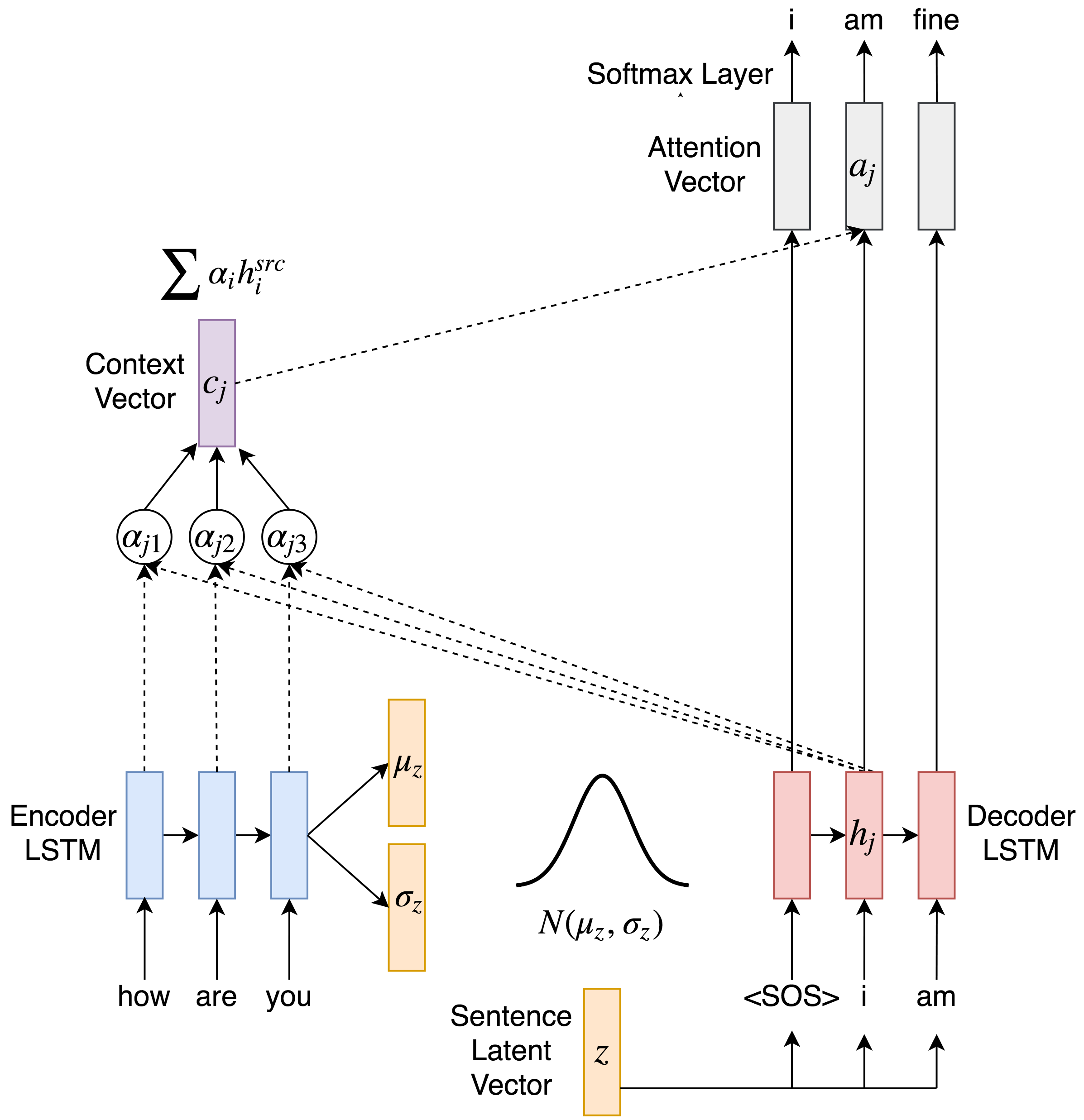}
  \caption{Illustration of VED with Deterministic Attention Mechanism}
  \label{fig:ved-det-attn}
\end{figure}

\begin{enumerate}
    \item \textbf{Multiplicative} \citep{luong2015effective} 
    \begin{equation}
    \widetilde{\alpha}_{ji}=\bm h\tar_j W^T\bm h\src_i    
    \end{equation}
    
    \item \textbf{Additive} \citep{bahdanau2014neural}
    \begin{equation}
    \widetilde{\alpha}_{ji}= v_a^T \tanh{\Big(W_1\bm h\tar_j + W_2\bm h\src_i \Big)}
    \end{equation}
    where, $W$, $W_1$, $W_2$ and $v_a^T$ are weights that are learnt via error backpropagation.
\end{enumerate}
While both methods work equally well in practice, this thesis uses the multiplicative style attention for all the encoder-decoder experiments. The next step is to take the sum of the source outputs $\{\bm h_i\src\}_{i=1}^{|\bm x|}$ weighted by $\alpha_{ji}$ to obtain the context vector.
\begin{equation}
    \bm c_j = \sum_{i=1}^{|\bm x|} \alpha_{ji}\bm h\src_i\label{eqn:context}
\end{equation}
Finally the attention vector can be computed using a $\tanh$ non-linear operation and learnt weights $W_c$ as follows 
\begin{equation}
    \bm a_j = f(\bm c_j, \bm h\tar_j) = \tanh(W_c[\bm c_j; \bm h\tar_j])
    \label{eqn:att}
\end{equation}
This attention vector can now be fed to the $\operatorname{softmax}$ layer at timestep $j$. Rewriting Equation~\ref{eqn:vocabsoftmax}, we obtain
\begin{align}
    p(y_j)=\operatorname{softmax}(W_\text{out}\bm a_j) 
\end{align}
In this manner, attention mechanisms are capable of dynamically aligning target information to the source information, during the generation process. This process has been illustrated in Figure~\ref{fig:ved-det-attn} and we refer to this as deterministic attention. 

\section{Variational Attention}
\label{sec:ved-var-attn}
One can observe from Figure~\ref{fig:ved-det-attn}, that the decoder LSTM has direct access to source information via the context vector $\bm c_j$ during attention computation. Although the latent vector $\bm z$ is fed to the decoder at every timestep, we fear that availability of source information via  $\bm c_j$ may cause the decoder \gls{lstm} to ignore $\bm z$. This may result in bypassing phenomenon due to which the \gls{ved} model does not encode any useful information into the $\bm z$ latent space.

\begin{figure}[!t]
\centering
  \includegraphics[width=0.7\linewidth]{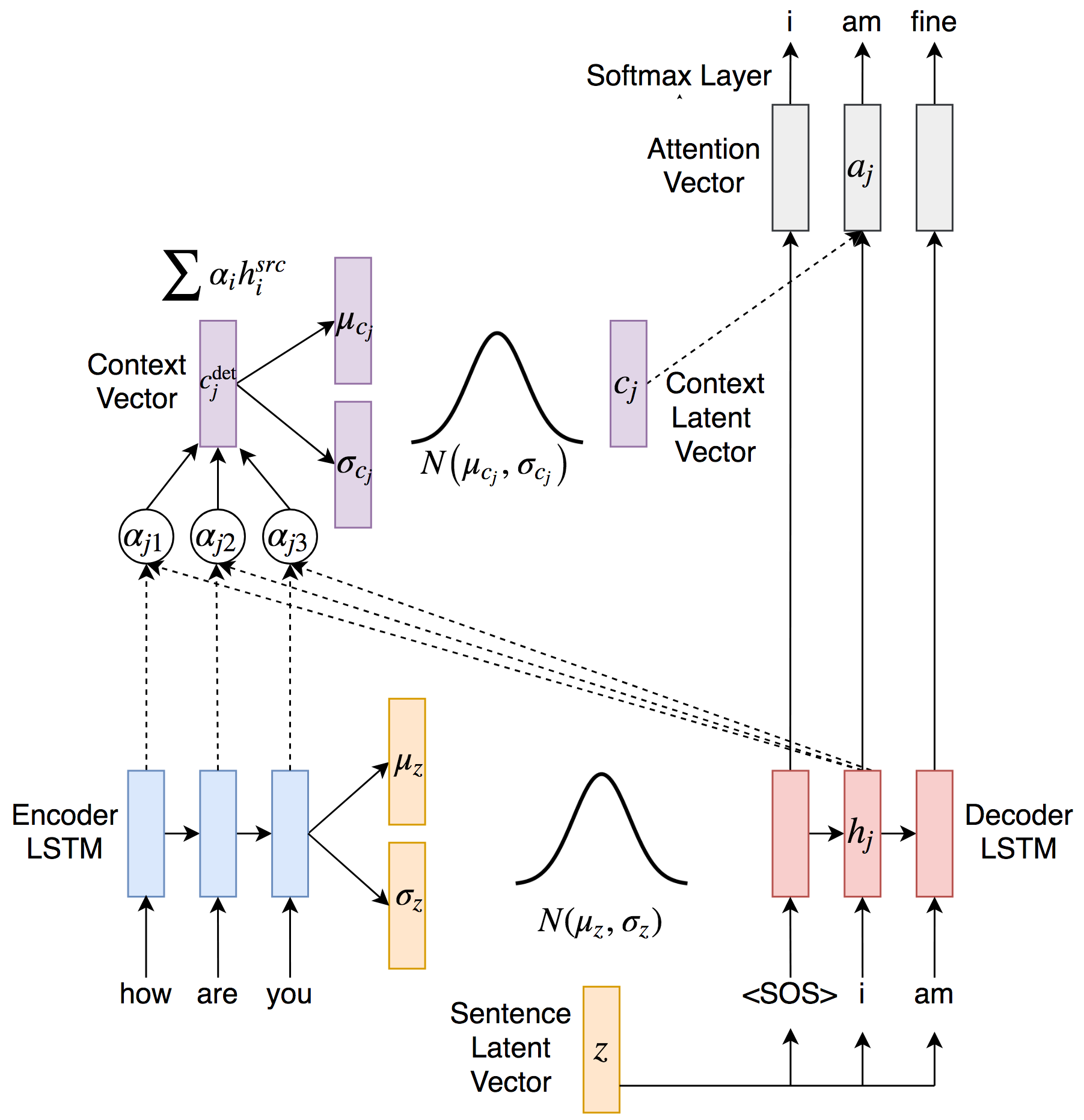}
  \caption{Illustration of VED with Variational Attention Mechanism}
  \label{fig:ved-var-attn}
\end{figure}

In this work, we propose variational attention mechanism to prevent bypassing.  Specifically, the context vector is modelled as a Gaussian random variable, which can be sampled using its mean and standard deviation. This results in a stochastic node before the $\operatorname{softmax}$ layer as shown in Figure~\ref{fig:ved-var-attn}. That is, we now have two nodes in the computational graph at which we do sampling, one for the sentence latent representation $\bm z$ and the other latent space for the context vector $\bm c_j$ at every timestep $j$. The attention is not modelled in a deterministic fashion anymore. In this way, we can prevent the bypassing issue in \gls{Seq2Seq} VED models with attention. 

\subsection{Derivation of Loss Function}
Variational attention treats both the sentence representation $\bm z$ and the context vector $\bm c_j$ as random variables. We would need to regularize the distribution of both these latent variables in the loss function. Analogous to the case of \gls{vae}, we rewrite Equation~\ref{eqn:VAElb} for evidence lower bound (\gls{elbo}) by adding the extra variable $\bm c_j$. The ELBO for the $j$th timestep of the $n$th data point, i.e., input $x\n$ and target $y\n$ , can be expressed as follows.

\begin{align}
  {\text{ELBO}}_j\n(\btheta, \bphi) &= \begin{aligned}[t]
      &\mathbb{E}_{z, c_j\sim q_\bphi( z, c_j| x\n)}\left[
\log p_\btheta(y\n|z, c_j)
\right]-\KL\left(q_\bphi( z, c_j| x\n)\|p( z, c_j)\right)
       \end{aligned}
\end{align}

\noindent where $q_\bphi$ and $p_\btheta$ correspond to the encoder and decoder neural networks with respective weights $\bphi$ and $\btheta$. Given $\bm x$, we can assume conditional independence between $\bm z$ and $\bm c_j$. Hence, the posterior factorizes as $q_\phi(\bm z, \bm c_j|\cdot)=q_\phi^{(z)}(\bm z|\cdot)\ q_\phi^{(c_j)}(\bm c_j|\cdot)$. We also assume separate priors for $\bm z$ and $\bm c_j$. In this way, the sampling procedure can be done separately and the $\KL$ loss can also be computed independently.

\begin{align}
    {\text{ELBO}}_j\n(\btheta, \bphi) &= \begin{aligned}[t]
      &\mathbb{E}_{ z\sim q^{(z)}_\bphi(z| x\n), c_j\sim q^{(c_j)}_\bphi( c_j| x\n)}\left[
\log p_\btheta( y\n| z, c_j)
\right]\\
&-\KL\left(q^{(z)}_\bphi( z| x\n)\|p( z)\right)-\KL\left(q^{(c_j)}_\bphi(c_j| x\n)\|p(c_j)\right)
       \end{aligned}\label{eqn:lb2}
\end{align}

\noindent As mentioned in Chapter~\ref{chap:vae}, we are required to maximize the evidence lower bound, which is equivalent to minimizing $-{\text{ELBO}}_j\n(\btheta, \bphi)$. The first term then becomes the negative log-likelihood, which can be computed as the standard Seq2Seq word prediction categorical cross-entropy loss, summed across all timesteps ($J_\text{rec}(\btheta,\bphi, \bm y\n)$). Next, we have two KL regularization terms - one for the sentence latent space $\bm z$ and the other for the context vector at each timestep $\bm c_j$. In both cases, we are required to minimize the KL divergence between the computed posterior and the pre-specified prior. Hence, the overall training objective of the Seq2Seq \gls{ved} with both variational sentence latent space $\bm z$  and variational attention $\bm c$ is to minimize the following loss function,
\begin{align}
J\n(\btheta,\bphi)&=J_\text{rec}(\btheta,\bphi,  y\n)+ \lambda_\KL\Big[\KL\left(q_\phi^{(z)}( z| x\n)\|p( z)\right)\nonumber\\ &\text{\quad\quad}+\gamma_a\sum_{j=1}^{| y|}
\KL\left(q_\phi^{(c_j)}( c_j| x\n)\|p( c_j)
\right)
\Big]\label{eqn:vatt.obj}
\end{align}
In the above equation, $\lambda_\KL$ acts as coefficient of both the \gls{kl} terms. We only anneal this hyperparameter in a manner similar to that of the \gls{vae} as discussed in Section~\ref{sec:vae-experiments}. $\gamma_a$ is the coefficient of the context vector's KL term which is kept constant throughout training. This is done because training \gls{Seq2Seq} variational models can be difficult and we would like to anneal just one of the coefficients which can simultaneously influence both KL terms. 

\subsubsection{Prior}
\label{sec:ved-priors}
Similar to the VAE described in Chapter~\ref{chap:vae}, the sentence latent code $\bm z$ is assumed to have a standard normal prior $\mathcal{N}(\textbf{0},\textbf{I})$. For the context vector $\bm c_j$, we propose two plausible priors:
\begin{enumerate}
    \item The simplest option would be to set the prior $p(\bm c_j)$ to be $\mathcal{N}(\textbf{0},\textbf{I})$. 
    \item With an understanding of attention mechanism in Seq2Seq models, one can observe that $\bm c_j$ is calculated as a linear combination of the source hidden states, i.e., $\bm c_j = \sum_{i=1}^{|\bm x|} \alpha_{ji}\bm h\src_i$. Geometrically, this means that the context vector is inside the convex hull of hidden representations of the source sequence, i.e., $\bm c_j\in\operatorname{conv}\{\bm h\src_i\}$. Thus, we choose $p(\bm c_j)=\mathcal{N}(\bar{\bm h}\src, \mathbf{I})$ as an alternative choice for the prior, where $\bar{\bm h}\src = \frac{1}{|\bm x|}\sum_{i=1}^{|\bm x|}\bm h_i\src$, the mean of the source hidden states. 
\end{enumerate}

\subsubsection{Posterior}
\label{sec:ved-posteriors}
Once we have pre-specified the prior distributions of both latent variables $\bm z$ and $\bm c_j$, we now discuss how the corresponding approximate posteriors are computed. In both cases, an \gls{lstm} based encoder neural network is used to parameterize the posterior distributions. More concretely, for the sentence latent space, the encoder's final hidden state $\bm h_t\src$ is linearly transformed to obtain $\bm \mu_z$ and $\bm \sigma_z$, the parameters of the posterior Gaussian distribution. Now the posterior of $\bm z$ can be defined as $q_\phi^{(z)}(\bm z| \bm x) = \mathcal{N}({\bm \mu_z},{\bm \sigma_z})$.

In an analogous manner, the posterior of the context vector at timestep $j$, denoted by $q_\bphi^{(c_j)}(\bm c_j|\bm x)$ is modeled as a Gaussian distribution $\mathcal{N}(\bm \mu_{c_j}, \bm\sigma_{c_j})$ using the same encoder neural network. We first compute the context vector in a deterministic manner ($\bm c_j^\text{det}$) as mentioned in Equation~\ref{eqn:context} and then transform it into the variational space using learnt parameters $\bm \mu_{c_j}$ and $\bm\sigma_{c_j}$. For the mean $\bm \mu_{c_j}$, we apply an identity transformation, i.e., $\bm \mu_{c_j}\equiv \bm c_j^\text{det}$. To compute $\bm \sigma_{c_j}$, we first transform $\bm c_j^\text{det}$ by a neural layer with $\tanh$ activation followed by another linear transformation. We can then sample the context vector $\bm c_j$ from its posterior $\mathcal{N}(\bm \mu_{c_j}, \bm\sigma_{c_j})$. Readers are referred to Figure~\ref{fig:ved-var-attn} for a diagrammatic overview of the method in which the parameters of the posterior distributions are computed.

\section{Experiments}
\label{sec:ved-experiments}
In this section, we provide the \gls{ved} training details, the datasets used and finally report the qualitative and quantitative results. 

\subsection{Datasets}
We evaluate and compare the proposed variational attention on two tasks - (1) Question Generation, (2) Dialog Systems. In question generation, the task is as follows - given an input sentence or paragraph, generate a question relevant to the input. Although this may seem trivial for humans, it is more difficult for computers since the generated questions need to be semantically and syntactically correct, and maintain topical relevance to the input. Apart from these, we  expect to obtain diverse but relevant questions by sampling different points from the latent space, conditioned on the same input. According to \cite{du2017learning}, attention mechanisms are especially critical in this task in order to generate relevant questions. We use the Stanford Question Answering Dataset~\citep[SQuAD]{rajpurkar2016squad} to carry out this task. The SQuAD dataset was originally curated for the task of machine comprehension, i.e., given a paragraph and a set of relevant questions, the computer is required to pick the sentences from the paragraph that are answers to the questions. The dataset has about 100k question-answer pairs, and we follow the same train-validation-test split as in \cite{du2017learning}. 

The second task is that of developing a generative conversational agent. The goal of such a dialog system is to generate replies in response to user utterances. Here again, response fluency and relevance are critical, for which attention mechanisms can play an important role. For this experiment, we use the Cornell Movie-Dialogs Corpus\footnote{\url{https://www.cs.cornell.edu/~cristian/Cornell_Movie-Dialogs_Corpus.html}}~\citep{cornell} as our dataset, which contains more than 200k conversational exchanges. While 180k sentences were used for training the model, 10k was held out for validation, and the rest for reporting test set performance. 

\subsection{Training Details}

We used \gls{rnn}s with 100d \gls{lstm} units for both the encoder and decoder; the dimension of the latent vector $\bm z$ was also 100d. Pre-trained \texttt{word2vec} \citep{mikolov2013distributed} word embeddings of 300d were used. For the question generation experiment, the vocabulary was set to the most frequent 40k words. The dataset in the dialog system experiment had fewer distinct unigrams and hence we set the model vocabulary size to 30k words. ADAM optimizer was used to train all the models. The convergence of the validation loss was set to be the stopping criterion. The question generation model is trained for 20 epochs. Dialog systems typically take more time for convergence and are trained for 250 epochs. The rest of the hyperparameters including the annealing schedule, word dropout rate, learning rate, etc., are set to those of the best performing \gls{vae} model described in Chapter~\ref{chap:vae}, namely  ADAM-$\tanh$-$3000$.

\noindent\textbf{Evaluation Metrics.} To determine the sentence generation performance, we report BLEU scores on the test set. Diversity of generated sentences is an important aspect in our study and this is evaluated using the automatic metrics described in Chapter~\ref{chap:bypass}, i.e., entropy and \emph{distinct} scores. 

\subsection{Quantitative Evaluation}
For the question generation experiment, we first replicate the neural network architecture proposed by \cite{du2017learning} for this task. As shown in Table~\ref{tab:ved-qgen}, we obtain similar BLEU scores as those of the deterministic encoder-decoder (\gls{ded}) described in \cite{du2017learning}.  Incorporating deterministic attention to this vanilla \gls{Seq2Seq} \gls{ded}, we obtain the model referred to as DED+DAttn. As expected, the attention mechanism improves the BLEU scores in comparison to the regular DED model. 

\begin{table*}[!t]
  \centering
  \resizebox{\textwidth}{!}{
    \begin{tabular}{l|c|ccccccc}
    \toprule
    \textbf{Model} & \textbf{Inference} & \textbf{BLEU-1} & \textbf{BLEU-2} & \textbf{BLEU-3} & \textbf{BLEU-4} & \textbf{Entropy} & \textbf{Dist-1} & \textbf{Dist-2} \\
    \midrule
    DED (w/o Attn) \cite{du2017learning} & MAP   & 31.34 & 13.79 & 7.36 & 4.26 & -     & -     & - \\
    \midrule
    DED (w/o Attn)& MAP   & 29.31 & 12.42 & 6.55 & 3.61  & -     & -     & - \\
    DED+DAttn & MAP   & 30.24 & 14.33 & 8.26 & 4.96 & -     & -     & - \\
    \midrule
    \multirow{2}{*}{VED+DAttn} 	& MAP   & \textbf{31.02} & 14.57 & 8.49 & 5.02 & -  & -  & - \\
						          & Sampling & 30.87 & \textbf{14.71} & \textbf{8.61} & \textbf{5.08} & 2.214  & 0.132  & 0.176 \\
	\midrule
    \multirow{2}{*}{VED+DAttn (2-stage training)} 	& MAP   & 28.88 & 13.02 & 7.33 & 4.16 & -  & -  & - \\
						        & Sampling & 29.25 & 13.21 & 7.45 & 4.25 & 2.241  & 0.140  & 0.188 \\
    \midrule
    \multirow{2}{*}{VED+VAttn-$0$} & MAP   & 29.70 & 14.17 & 8.21 & 4.92 & -  & -  & - \\
                                   & Sampling & 30.22 & 14.22 & 8.28 & 4.87 & \textbf{2.320}  & \textbf{0.165}  & \textbf{0.231} \\
    \midrule
    \multirow{2}{*}{VED+VAttn-$\bar{h}$} & MAP   & 30.23 & 14.30 & 8.28 & 4.93 & -  & -  & - \\
                                  & Sampling & 30.47 & 14.35 & 8.39 & 4.96 & 2.316 & 0.162 & 0.228 \\
    \bottomrule
    \end{tabular}
}
    \caption{BLEU, entropy, and distinct scores on the question generation task. We compare the deterministic encoder-decoder (DED) and variational encoder-decoders (VEDs). For VED, we have several variates: deterministic attention (DAttn) and the proposed variational attention (VAttn). Variational models are evaluated by both max \textit{a posteriori} (MAP) inference and sampling.}
  \label{tab:ved-qgen}%
\end{table*}

The next set of results correspond to the variational encoder-decoder models (\gls{ved}). Given an input, the output sentence generation at inference time can be done either by sampling from the latent space or by using the max \textit{a posteriori} estimate (MAP). This refers to the most probable output corresponding to the given input, which in the Gaussian case is the mean of the posterior distribution. In the \textit{sampling} setting, we draw 10 samples ($\bm z$ and/or $\bm a$) from the posterior given $\bm x$, i.e., for each data point, and report average BLEU scores. We also compute the diversity metrics on each set of 10 sampled sentences and average it across the test data. Note that these metrics can only be calculated for the \textit{sampling} setting. 

Comparing the proposed variational attention (VED+VAttn) and the deterministic attention (VED+DAttn) models, we observe that VED+VAttn significantly outperforms VED+DAttn in terms of the diversity of generated sentences. It should be noted that entropy is a logarithmic measure, and hence the difference of 0.1 in Table~\ref{tab:ved-qgen} is significant; VED+VAttn also generates more distinct unigrams and bigrams than VED+DAttn. The BLEU scores obtained by the variational attention models are only slightly lower than the deterministic attention counterparts. This means that the output sentence reconstruction performance of both models is similar, while the VED+VAttn models can generate more diverse outputs. This confirms our hypothesis that deterministic attention serves as a  bypassing connection that affects the learning of a good latent space. The lower diversity of DAttn models can be attributed to the learnt posterior distributions being more peaked, similar to the issue depicted in Figure~\ref{fig:vae-hidden-state}.

Table~\ref{tab:ved-qgen} reports results for the two priors that were proposed in Section~\ref{sec:ved-priors}. \newline VED+VAttn-0 and VED+VAttn-$\bar h$ refer to the variational attention models with priors $\mathcal{N}(\bm 0, \mathbf{I})$ and $\mathcal{N}(\barh, \mathbf{I})$ respectively. VED+VAttn-0 has slightly lower BLEU but higher diversity. The results are generally comparable, showing both priors are reasonable.

The model indicated by VED+DAttn (2-stage) refers to a heuristic based model which was implemented for the same task. In this setting, the VED is first trained without attention for 6 epochs, and then the attention mechanism is incorporated into the model for the rest of the training. The logic behind this simple heuristic is that we first allow the VED to learn a meaningful latent space during the early stages of training, and we only introduce the deterministic attention connection at a later stage. However, this still continues to influence the model performance as a result of bypassing. We conclude that such simple heuristics do not help much, and are worse than the principled variational attention mechanism in terms of all BLEU and diversity metrics.

Next we present the results on the dialog systems experiment. In general, automatic conversational systems tend to have low BLEU scores because for a given user utterance, there are multiple possible responses which are valid \citep{liu2016not}. The dataset typically only provides only one ground truth response. Hence, it is much harder to evaluate dialog systems. Nevertheless, we report BLEU scores and diversity metrics on the three main models (refer Table~\ref{tab:ved-dialog}). The deterministic encoder-decoder model (\gls{ded}) tends to have a better output reconstruction capability. This is expected since it is explicitly trained to minimize the negative log likelihood of observing the data. On the other hand, \gls{ved} models have additional KL regularization terms in their loss functions.  VED+VAttn-0 and VED+VAttn-$\bar h$ provide similar results and only the model with $\mathcal{N}(\barh, \mathbf{I})$ prior is chosen for reporting purposes.

It can be observed that both the quality and diversity of sentences generated by VED+VAttn-$\bar h$ are slightly better than VED+DAttn. However, we find the improvement is not so large as in the question generation task. We conjecture the reason for this to be that in conversational systems, there is a weaker alignment between the source and target information. As a result, the attention mechanism itself is less effective. 

\begin{table}[!t]
  \centering
    \resizebox{\textwidth}{!}{
    \begin{tabular}{l|c|ccccccc}
    \toprule
    \textbf{Model} & \textbf{Inference} & \textbf{BLEU-1} & \textbf{BLEU-2} & \textbf{BLEU-3} & \textbf{BLEU-4} & \textbf{Entropy} & \textbf{Distinct-1} & \textbf{Distinct-2} \\
    \midrule
    DED+DAttn & MAP   & 5.75  & 1.84  & 0.99  & 0.64  & -     & -     & - \\
    \midrule
    \multirow{2}[2]{*}{VED+DAttn} & MAP   & 5.33  & 1.68  & 0.88  & 0.57  & -     & -     & - \\
          & Sampling & 5.34  & 1.68  & 0.89  & 0.57  & 2.113 & 0.311 & 0.450 \\
    \midrule
    \multirow{2}[2]{*}{VED+VAttn-$\bar{h}$} & MAP   & 5.48  & 1.78  & 0.97  & 0.64  & -     & -     & - \\
          & Sampling & 5.55  & 1.79  & 0.97  & 0.64  & 2.167 & 0.324 & 0.467 \\
    \bottomrule
    \end{tabular}%
    }
    \caption{Results on the conversational systems experiment}
  \label{tab:ved-dialog}%
\end{table}%

\subsubsection{Learning Curves}
\begin{figure}[!t]
\centering
  \includegraphics[width=\linewidth]{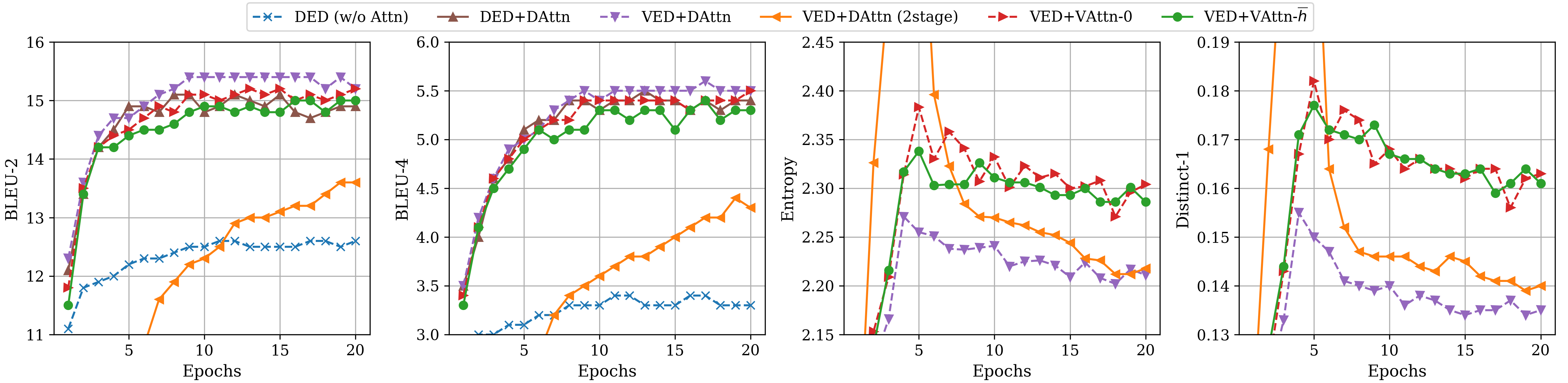}
  \caption{BLEU-2, BLEU-4, Entropy, and Distinct-1 calculated on the validation set as training progresses.}
  \label{fig:ved-model-trends}
\end{figure}

For the question generation experiment, we also studied the learning curves of the evaluation metrics (on the validation set) as training progresses. In Figure~\ref{fig:ved-model-trends}, we illustrate BLEU-2 and BLEU-4 (representative of output reconstruction quality), and entropy and Distinct-1 (representative of diversity). It can be observed that BLEU scores and diversity are conflicting objectives. In order to attain a higher BLEU score, the model needs to be more \textit{deterministic} in nature, as a result of which it may lose interesting \textit{variational} properties such as diversity. The red and green lines indicate the variational attention models, which tend to have comparable BLEU scores while maintaining high diversity. This verifies the effectiveness of our model design which circumvents the bypassing phenomenon. 

\subsubsection{Strength of Attention KL Loss}
\begin{figure}[!t]
\centering
  \includegraphics[width=\linewidth]{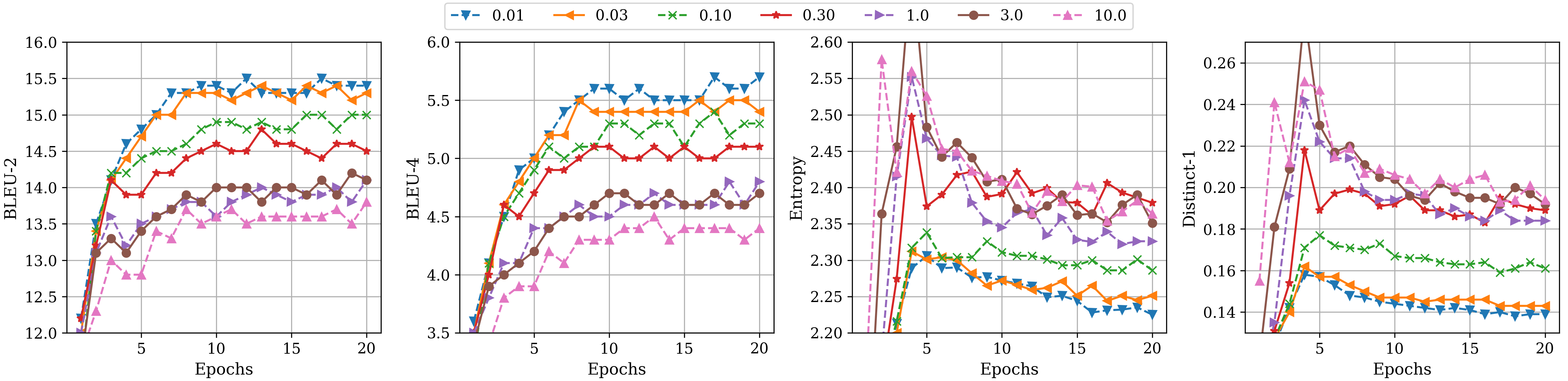}
  \caption{BLEU-2, BLEU-4, Entropy, and Distinct-1 for multiple runs of the same model (VED+VAttn-$\bar h$) with different $\gamma_a$ values.}
  \label{fig:ved-gamma-sensitivity}
\end{figure}

As mentioned in Equation~\ref{eqn:vatt.obj}, annealing is only done for the common \gls{kl} coefficient, i.e., $\lambda_{\KL}$, whereas the coefficient of the attention KL $\gamma_a$ loss is fixed for each experiment. In this section, we study the influence of $\gamma_a$, which affects the strength of attention KL loss in variational attention models, specifically the VED+VAttn-$\bar h$ variant. This is done by running multiple experiments on the question generation task with the same model, the only difference being the value of $\gamma_a$ in each run. The learning curves of the different runs are presented in Figure~\ref{fig:ved-gamma-sensitivity}. It can be seen that with lower values of $\gamma_a$, the BLEU scores are higher, while the corresponding diversity metrics are lower. This is expected because a lower $\gamma_a$ gives the model less incentive to optimize the attention's KL term, which then causes the model to behave more \textit{deterministic}. On the other hand, high $\gamma_a$ increases the diversity of the sentences generated by the model, at the cost of output reconstruction performance, i.e, lower BLEU scores. Based on this experiment, we chose a value of 0.1 for $\gamma_a$, as it yields a learning curve in the middle among the different $\gamma_a$ values, being a good balance between quality and diversity. The results reported in Table~\ref{tab:ved-qgen} and Table~\ref{tab:ved-dialog} correspond to this setting.


\subsection{Qualitative Evaluation}
\label{sec:ved-qualitative}
In this section, we compare the proposed variational attention to the deterministic attention counterpart in terms of qualitative samples. We generate multiple samples conditioned on the same input for both models. This was done on the task of question generation, where we input a sentence and  attempt to generate a relevant question. It can be observed from Table~\ref{tab:case} that VED+VAttn-$\bar h$  is capable of producing multiple diverse questions relevant to the input sentence. In contrast, the questions obtained by sampling from the latent spaces of VED+DAttn are less diverse. In many cases, the same sentence is generated multiple times, indicating that the posterior distributions corresponding to the inputs tend to be more peaked. This is caused by low standard deviations due to which the sampling always happens from just around the mean value. Thus, we are able to demonstrate the benefits of variational attention which is proposed as a method to alleviate the bypassing issue in \gls{Seq2Seq} \gls{ved} networks.

\begin{table}[!t]
	\centering
    \resizebox{0.8\textwidth}{!}{
	\begin{tabular}{|rl|}
		\toprule
		\textbf{Source}\!\!\!& \textit{when the british forces evacuated at the close of the war in 1783 ,\!\!\!} \\
		&\textit{they transported 3,000 freedmen for resettlement in nova scotia .}\\
		\textbf{Reference}\!\!\!& \textit{in what year did the american revolutionary war end ?} \\
		\midrule
		\multirow{3}[2]{*}{\textbf{VED+DAttn}}\!\!\!& \textit{how many people evacuated in newfoundland ?} \\
		& \textit{how many people evacuated in newfoundland ?} \\
		& \textit{what did the british forces seize in the war ?} \\
		\midrule
		\!\!\!\multirow{3}[2]{*}{\textbf{VED+Vattn}-$\bar{\bm h}$}\!\!\!& \textit{how many people lived in nova scotia ?} \\
		& \textit{where did the british forces retreat ?} \\
		& \textit{when did the british forces leave the war ?} \\
		\bottomrule
    
		\toprule
		\textbf{Source}\!\!\!& \textit{downstream , more than 200,000 people were evacuated from} \\
		& \textit{mianyang by june 1 in anticipation of the dam bursting .}\\
		\textbf{Reference}\!\!\!& \textit{how many people were evacuated downstream ?} \\
		\midrule
		\multirow{3}[2]{*}{\textbf{VED+DAttn}}\!\!\!& \textit{how many people evacuated from the mianyang basin ?} \\
		& \textit{how many people evacuated from the mianyang basin ?} \\
		& \textit{how many people evacuated from the mianyang basin ?} \\
		\midrule
		\!\!\!\multirow{3}[2]{*}{\textbf{VED+VAttn}-$\bar{\bm h}$}\!\!\!& \textit{how many people evacuated from the tunnel ?} \\
		& \textit{how many people evacuated from the dam ?} \\
		& \textit{how many people were evacuated from fort in the dam ?} \\
		\bottomrule
	\end{tabular}
}
	\caption{Qualitative samples of the question generation task.}
	\label{tab:case}%
\end{table}%

\subsubsection{Human Evaluation}
Although we quantitatively evaluate our model in the previous section, none of the metrics capture the language fluency of the model. It is difficult to come up with an automatic evaluation metric that can compare the level of fluency of the text generated by each model. However, this is an important aspect considering the recent developments in artificial intelligence where machines are trained to be more \textit{human-like}. 

Thus, in order to assess the quality of the generated text in terms of language fluency, a human evaluation study was carried out with the text generated from the question generation task. For the two main models under comparison, VED+DAttn and VED+VAttn-$\bar h$, a randomly shuffled subset of 100 generated questions was selected. Six human evaluators were asked to rate the fluency of these 200 questions on a 5-point scale: 5-Flawless, 4-Good, 3-Adequate, 2-Poor, 1-Incomprehensible, following the annotation scheme described in \citep{stent2005evaluating}. The human evaluators were fellow researchers proficient in English. No additional instructions apart from the 5-point annotation scheme were provided to the evaluators. Essentially, this gave them the freedom to decide what they thought to be a fluent question versus a poor question. 

The average rating obtained for VED+DAttn was 3.99 and for VED+VAttn-$\bar h$ was 4.01, which shows that, on an average, the sentences generated by both models are of \textit{good} fluency. It is to be further noted that the difference between the scores of VED+DAttn and VED+VAttn-$\bar h$ is not statistically significant (based on hypothesis test comparing two means, at the $5\%$ level). The human annotations achieved 0.61 average Spearman correlation coefficient (measuring order correlation) between any two annotators. According to \cite{swinscow1976statistics}, this indicates \emph{moderate} to \emph{strong} correlation among different annotators. Hence, we can safely conclude that variational attention does not negatively affect the fluency of sentences.

\chapter{Summary and Conclusions}
\section{Summary of Research Work}
In this research, we present deep learning approaches to probabilistic natural language generation. In particular, we design and implement variational neural network models for text generation. Variational autoencoders are capable of mapping sentences into a continuous latent space from which it is possible to sample and generate new sentences. However, VAEs are notoriously difficult to train due to issues associated with the KL regularization term vanishing to zero, resulting in model collapse. The first part of the thesis addresses this problem with the help of various optimization strategies.

Next, we further explore VAE architectures and discover the \textit{bypassing} phenomenon. Hidden state initialization of the decoder results in a bypassing connection which degrades the VAE into a deterministic model. We describe how this can negatively impact the learning of a meaningful latent space.

Finally, we move on to variational encoder-decoder models wherein we require to transform a given source sequence into a desired target sequence. We realize that traditional sequence-to-sequence attention mechanisms act as bypassing connections. To circumvent this problem, we propose the variational attention mechanism. We show that by treating the contect vector as a random variable, it is possible to overcome the \textit{bypassing} issue.

\section{Conclusions}
Being able to generate meaningful textual data is an important characteristic of intelligent machines. For probabilistic generation of natural language sentences, we developed a sequence-to-sequence variational autoencoder. The VAE was successfully trained by implementing optimization heuristics, namely KL cost annealing and word dropout. By carefully engineering the annealing rate, schedule and threshold, the VAE that we trained was able to learn a meaningful continuous latent space. We demonstrate interesting properties of the latent space such as random sampling, linear interpolation and sampling from the neighbourhood, and compare it to a baseline deterministic autoencoder. We show that VAEs can learn meaningful sentence representations and also generate previously unseen sentences which are semantically and syntactically correct. 

We studied the problem of bypassing phenomenon in VAEs wherein the decoder has a deterministic access to source information. We illustrate with the example of decoder hidden state initialization that such bypassing connections cause the VAE model to ignore the latent space during training. We show both quantitatively and qualitatively that bypassing results in the loss of interesting properties of the variational latent space and degrades the model into a deterministic autoencoder. 

In variational encoder decoder models, we observe similar bypassing issues when traditional sequence to sequence attention is used. To prevent the decoder from having direct access to the encoder, we proposed a variational attention mechanism for VED frameworks. This technique introduces an additional stochastic node in the computational graph by modelling the context vector as a random variable with a pre-specified probability distribution. In practice, we sample the context vector at each timestep of the decoding process. An additional term, that computes the KL divergence between the context vector's posterior and prior distributions, is added to the loss function. Two plausible priors were proposed, which work equally well. With empirical results on two tasks - question generation and dialog systems, we show that variational attention yields more diversified samples while retaining high quality. 

\section{Future Work}
For the VED model with variational attention, we impose a probabilistic distribution on the context vector. However, the context vector itself is computed using attention weights, $\alpha_{ji}$ (see Figure~\ref{fig:ved-var-attn}). Instead of the context vector $\bm c_j$ being modeled as a random variable, we could assume the attention weights $\bm \alpha_j =  \{\alpha_{ji}\}_{i=1}^{|\bm x|}$ as random variables which are sampled from a distribution. Note that $ \alpha_j $ represent probability values that sum up to 1. Hence, they can be thought of as parameters of a categorical distribution which has Dirichlet distribution as its conjugate prior. To incorporate this scenario into the VED framework, we are required to sample from a Dirichlet distribution instead of sampling from a Gaussian distribution. However, this relies on a reparametrization trick to propagate error gradient back to the recognition neural network (refer Figure~\ref{fig:reparam}). In other words, we should be able to sample from a fixed distribution (i.e., without learnt parameters) and then obtain the variable in the desired latent space by doing necessary variable transformation. However, doing this is non-trivial for  Dirichlet distribution.
In future work, it would be interesting to investigate VEDs that model the attention weights with Dirichlet distributions. 

Since the latent space learnt by VAEs is continuous and meaningful, future work can also explore how VAEs can be used for disentangling sentence level attributes. For example, if we are able to represent different attributes such as  sentiment, tense, topic, etc., into different dimensions of the latent vector, this would be useful for supervised tasks later in the machine learning pipeline. 

\nocite{bahuleyan2018VarAtt, bahuleyan2018wae}



\bibliographystyle{plainnat}
\cleardoublepage 
\phantomsection  
\renewcommand*{\bibname}{References}

\addcontentsline{toc}{chapter}{\textbf{References}}

\bibliography{uw-ethesis}


\appendix
\chapter*{APPENDICES}
\addcontentsline{toc}{chapter}{APPENDICES}
\chapter*{Appendix A: Python Code for Encoder-Decoder Models}
\label{AppendixA}
Code is available at \url{https://github.com/HareeshBahuleyan/tf-var-attention}

\end{document}